\documentclass[conference]{IEEEtran}
\usepackage{times}

\usepackage[numbers]{natbib}
\usepackage{multicol}
\usepackage[bookmarks=true]{hyperref}
\usepackage{graphicx}
\usepackage{multirow}
\usepackage{booktabs}
\usepackage{pifont}
\usepackage{natbib}
\usepackage{listings}
\usepackage{xcolor}
\usepackage{colortbl}
\usepackage{subcaption} 
\definecolor{gold}{rgb}{1.0, 0.84, 0.0} 
\definecolor{Gray}{gray}{0.5}
\definecolor{LGray}{gray}{0.9}
\definecolor{darkblue}{RGB}{94,110,186}
\definecolor{darkGreen}{RGB}{92, 148, 110}
\definecolor{myblue}{RGB}{14, 121, 178}
\definecolor{myred}{RGB}{192, 0, 0}
\definecolor{darkgreen}{HTML}{04bf29}
\definecolor{darkred}{HTML}{D1191F}
\definecolor{crimson}{RGB}{153, 0, 0}
\usepackage[table,xcdraw]{xcolor}
\usepackage{hhline}

\newcommand{\darkblue}[1]{\textcolor{darkblue}{#1}}

\begin{document}

\title{LIBERO-X: Robustness Litmus for Vision-Language-Action Models}


\author{
  Guodong Wang$^{\ast1}$, Chenkai Zhang$^{\ast2}$, Qingjie Liu$^2$, Jinjin Zhang$^2$,
  Jiancheng Cai$^1$, Junjie Liu$^{\dagger1}$,  Xinmin Liu$^{\ddagger1}$ \\ 
  $^1$Meituan,
  $^2$Beihang University \\
  $\ast$Equal Contribution $\dagger$Corresponding author ${\ddagger}$Project Leader \\
  Project page: \url{https://zackhxn.github.io/LIBERO-X/} 
}



%

\maketitle

\begin{abstract}

Reliable benchmarking is critical for advancing Vision–Language–Action (VLA) models, as it reveals their generalization, robustness, and alignment of perception with language-driven manipulation tasks. However, existing benchmarks often provide limited or misleading assessments due to insufficient evaluation protocols that inadequately capture real-world distribution shifts. This work systematically rethinks VLA benchmarking from both evaluation and data perspectives, introducing LIBERO-X, a benchmark featuring: 1) A hierarchical evaluation protocol with progressive difficulty levels targeting three core capabilities: spatial generalization, object recognition, and task instruction understanding. This design enables fine-grained analysis of performance degradation under increasing environmental and task complexity; 2) A high-diversity training dataset collected via human teleoperation, where each scene supports multiple fine-grained manipulation objectives to bridge the train-evaluation distribution gap. Experiments with representative VLA models reveal significant performance drops under cumulative perturbations, exposing persistent limitations in scene comprehension and instruction grounding. By integrating hierarchical evaluation with diverse training data, LIBERO-X offers a more reliable foundation for assessing and advancing VLA development.

\end{abstract}

\IEEEpeerreviewmaketitle

\section{Introduction}

Vision–Language–Action (VLA) models~\cite{zitkovich2023rt,kim2024openvla,black2410pi0,wen2025tinyvla,bjorck2025gr00t,team2024octo,kim2025fine,li2025unified,pertsch2025fast} have recently made significant strides in robotic manipulation, emerging as a key paradigm for integrating perception, language understanding, and control. By processing visual observations and natural language instructions in an end-to-end manner, these models directly map multimodal inputs to low-level control signals and have achieved strong performance across a wide range of complex manipulation tasks. With rapid advances in VLA architectures and training algorithms, a central challenge for the field is now the systematic evaluation of their actual capabilities~\cite{wang2025vlatest,zhang2025vlabench,li2024evaluating}.

In response, the research community has introduced a series of standardized benchmarks, such as LIBERO~\cite{liu2023libero} and SimpleEnv~\cite{li2024evaluating}, which provide controlled, reproducible environments covering diverse task configurations, and are widely used to assess multimodal understanding, long-horizon reasoning, and knowledge transfer. Despite impressive results on these existing benchmarks, current VLA evaluations are often confounded by limitations in both \textbf{evaluation protocols} and \textbf{data distribution}, leading to overly optimistic or even misleading conclusions about model capability. 

\begin{figure*}[htbp]
    \centering
    \includegraphics[width=\textwidth]{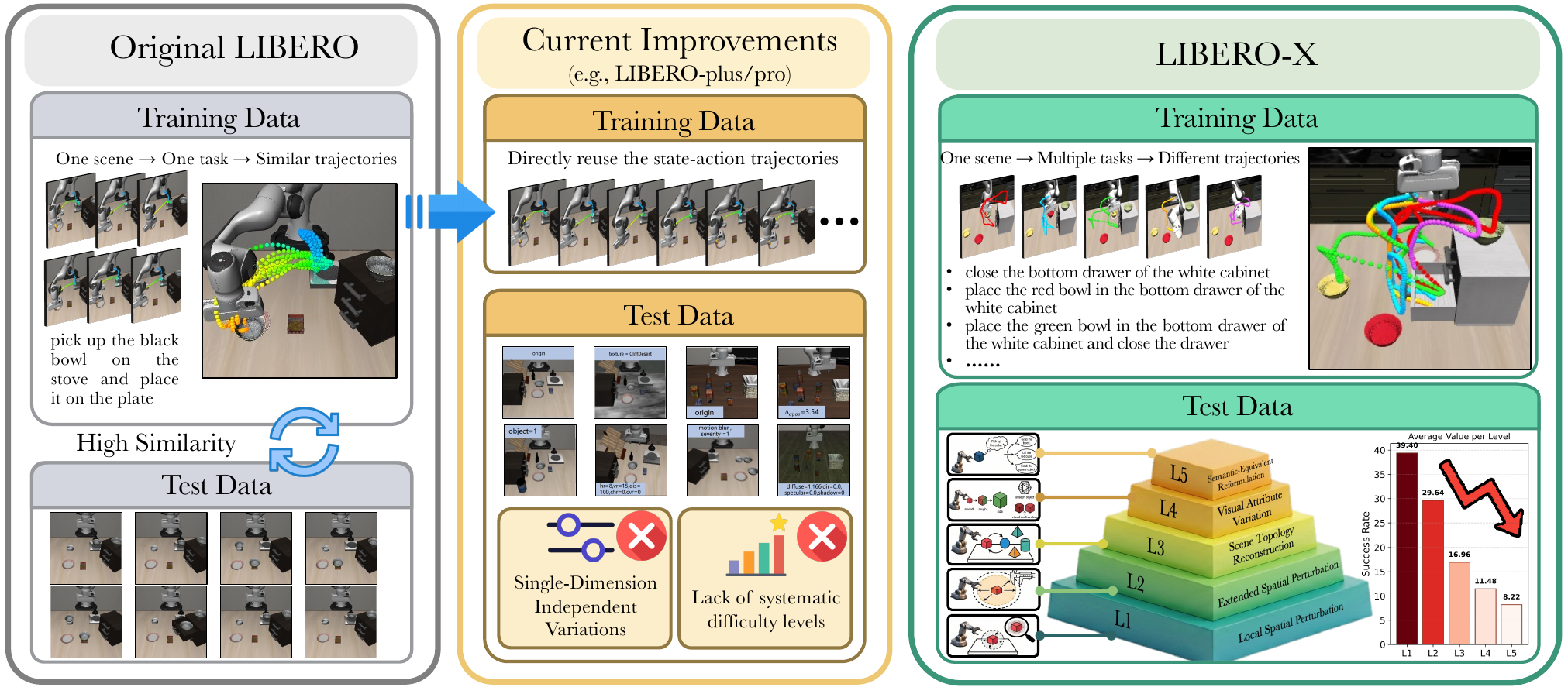}
    \raggedright
    
    \hspace*{30pt} \textbf{(a)} Original LIBERO \hspace{40pt} \textbf{(b)} Current Improvements \hspace{90pt} \textbf{(c)} LIBERO-X
    \caption{\textbf{Comparison with related simulation benchmarks.}  \textbf{(a)}: LIBERO lacks diversity by coupling scenes with homogeneous trajectories, inducing action overfitting, while its test set closely mirrors training data. \textbf{(b)}: Recent extensions reuse original training trajectories, introducing perturbations but model them independently, failing to capture complex distribution shifts. \textbf{(c)}: LIBERO-X enhances training via multi-task scenes with diverse trajectories. Its multi-level evaluation protocol progressively escalates complexity, revealing significant performance degradation across VLA models as difficulty increases, thus enabling rigorous robustness assessment.}
    \vspace{-0.5cm}
    \label{fig:pipeline}
\end{figure*}

First, existing benchmarks typically focus on individual perturbation types, primarily spatial ones. Extensions such as SafeLIBERO~\cite{hu2025vlsa}, LIBERO-PRO~\cite{zhou2025liberopro}, and LIBERO-Plus~\cite{fei25libero-plus} increase the diversity of test conditions by introducing additional perturbation factors (e.g., random obstacles for safety studies, or initial states, task instructions, and environment configurations). However, as shown in Figure \ref{fig:pipeline} \textbf{(b)}, they still suffer from two major issues: (i) perturbations along different dimensions are mostly modeled independently, failing to capture the multi-source distribution shifts that arise in realistic deployment; and (ii) there is no systematic progression of difficulty from easy to hard, making it difficult to characterize performance degradation as environmental and task complexity increase. Consequently, these limitations hinder a comprehensive assessment of model robustness under realistic, compounded distribution shifts.

Second, evaluation is frequently unfaithful due to shortcomings in both the training and testing datasets. Specifically, training data lack sufficient diversity, and there is limited divergence between training and testing scenarios and tasks. Benchmarks such as LIBERO typically adopt test configurations that closely resemble the training settings, introducing only minor perturbations, such as small changes in initial object positions, as shown in Figure \ref{fig:pipeline} \textbf{(a)}. Under this narrow train–test gap, models often attain near-saturated success rates, illustrating that superficially varied test scenarios do not necessarily reveal the model's true generalization ability. On the training data side, existing datasets exhibit limited diversity in tasks, scenes, and trajectories, which fundamentally constrains the attainable capability of learned models. A single scene is usually associated with only one or a few tasks, and demonstrations for the same task follow highly homogeneous action sequences and intermediate states, effectively collapsing into a small set of template-like strategies. Overfitting is further exacerbated under such narrow data distributions, causing models to memorize a limited number of stereotyped behaviors and to mechanically replay training actions~\cite{fei25libero-plus, zhou2025liberopro}. 

In this work, we improve benchmarking by jointly addressing evaluation hierarchy and training data diversity, as shown in Figure \ref{fig:pipeline} \textbf{(c)}. Regarding the evaluation, we focus on three core capabilities of VLA models for robotic manipulation: spatial generalization, object recognition, and instruction understanding. Centered on these capabilities, we introduce LIBERO-X, an evolved version of the original LIBERO benchmark, where “X” denotes \textit{eXtreme} to reflect its goal of pushing evaluation limits. LIBERO-X first defines a new evaluation protocol that delivers a progressively challenging, multi-level test suite for assessing the generalization of VLA models in complex environments. Specifically, we adopt a \textit{reuse–adjust–reconstruct} strategy over training scenes to construct a five-level hierarchy of fine-grained manipulation tasks. The first two levels introduce increasing magnitudes of spatial perturbations, progressing from minor positional jitter to broader spatial randomization. The third level alters scene topology (e.g., swapping target and placement locations), breaking fixed spatial associations learned during training. The fourth level incorporates unseen and confounding objects, and varies fine-grained attributes such as color and texture. The fifth level further incorporates semantic variations in task instructions to evaluate language–vision–action alignment. Higher levels are formed by adding new perturbation dimensions on top of lower levels, thereby progressively escalating difficulty to enable precise analysis of performance degradation as environmental and task complexity increase. Complementing this hierarchy, we implement a multi-label evaluation strategy where each task is annotated with fine-grained attributes: interaction type, subtask count, spatial relation, and object attribute. This granular labeling facilitates in-depth diagnostics, allowing researchers to pinpoint specific failure modes across distinct manipulation dimensions.

To mitigate the overfitting risks associated with the relatively homogeneous scenes and trajectories in the original LIBERO training set, we further construct a high-diversity training dataset via human teleoperation. This dataset fundamentally increases scene and task diversity by incorporating variations in object attributes (e.g., size, color, texture) and spatial relations (e.g., left/right, near/far, front/behind), together with language instructions. Importantly, a single scene is associated with multiple manipulation tasks, avoiding template-like scene–task–trajectory mappings. This design provides a data foundation by exposing models to broader task–scene distributions during training. Consequently, performance on LIBERO-X more faithfully reflects models’ transferability and robustness in complex, dynamic environments. 

Overall, by combining richer training diversity and multi-dimensional hierarchical perturbations at test time, LIBERO-X represents a systematic rethinking of VLA model evaluation for robotic manipulation. We expect it to serve as a comprehensive benchmark that more effectively guides VLA development toward reliable deployment in robotics.

The main contributions of this work are three-fold. 
\begin{itemize}
    \item We propose LIBERO‑X, a comprehensive benchmark for manipulation that introduces a progressively challenging evaluation framework with multi-label annotations and jointly perturbs spatial layouts, object properties, and instruction semantics to systematically characterize model performance under multi-dimensional distribution shifts.
    \item We construct a high-diversity training dataset comprising 2,520 demonstrations, 600 tasks, and 100 scenes collected via human teleoperation, substantially increasing scene diversity and task granularity.
    \item Extensive experiments on fine-tuned representative VLA models reveal notable performance degradation as task and scene complexity increase, highlighting critical limitations in scene comprehension and instruction grounding and offering empirical insights for future model design.

\end{itemize} 

\section{Related Work}

\textbf{Generalist Robot Policies.} Early imitation learning methods, such as ACT~\cite{zhao2023learning} and Diffusion Policy (DP)~\cite{chi2025diffusion}, addressed sequential decision-making via action chunking and conditional denoising, respectively. Building on these, Octo~\cite{ghosh2024octo} and RDT-1B~\cite{liu2025rdtb} scaled these architectures on large-scale robot dataset (e.g. the Open X-Embodiment dataset~\cite{oxe2024}), demonstrating that increasing model capacity significantly improves multi-task manipulation. 

Inspired by the generalization and reasoning capabilities of Large Language Models (LLMs), a new paradigm of VLA models has emerged. These methods aim to leverage the rich semantic representations of pretrained VLM backbones for robotic control, diverging primarily in their action modeling strategies. Discrete approaches like RT-2~\cite{zitkovich2023rt} and OpenVLA~\cite{kim2024openvla} discretize continuous actions into text tokens, treating manipulation as a next token prediction task. Recent work like FAST~\cite{karl2025fast} further optimizes this tokenization for higher efficiency. Conversely, continuous approaches retain the precision of the action space but vary in their formulation. Some methods, such as RoboFlamingo~\cite{li2024visionlanguage} and OpenVLA-OFT~\cite{kim2025fine}, adopt direct regression heads optimized via mean squared error (MSE) loss. Others integrate more expressive generative policies: CogACT~\cite{li2024cogact} separates high-level understanding from low-level execution by attaching a diffusion head to the VLM, while $\pi_0$~\cite{intelligence2025pi05}, $\pi_{0.5}$~\cite{intelligence2025pi05} and GR00T~\cite{bjorck2025gr00t} achieve state-of-the-art performance by incorporating flow matching directly with VLM backbones. To enhance the grounding of VLA models, recent approaches also focus on explicitly or implicitly incorporating additional information, such as spatial structure and geometry~\cite{quspatialvla, zhen20243d} and reasoning and planning signals~\cite{wen2025diffusionvla, intelligence2025pi05}. 

Despite these advancements, evaluating VLA models remains challenging due to the high cost and variability of real-robot experiments. To address this, we develop a comprehensive simulation benchmark to enable the systematic and reproducible assessment of VLA methods.

\textbf{Robotic Manipulation Benchmarks.} Standardized benchmarks are crucial for evaluating the generalization and robustness of robotic manipulation policies. While numerous benchmarks have been developed for real‑world systems, this review focuses primarily on simulation‑based benchmarks due to their experimental convenience and reproducibility. Early efforts such as RLBench~\cite{james2020rlbench} and LIBERO~\cite{liu2023libero} have significantly advanced the field by establishing unified evaluation protocols. Recent benchmarks~\cite{fei25libero-plus, pumacay2024colosseum, zhou2025liberopro, zhou2025exploring, li2025task} have explored specific facets of generalization. For example, AGNOSTOS~\cite{zhou2025exploring} and CALVIN~\cite{mees2022calvin} investigate transfer across tasks, while RoboCerebra~\cite{han2025robocerebra} and VLABench~\cite{zhang2025vlabench} focus on long-horizon manipulation and complex scene-task understanding. Meanwhile, works such as LIBERO-PRO~\cite{zhou2025liberopro} and LIBERO-Plus~\cite{fei25libero-plus} expand the evaluation by introducing controlled perturbations across dimensions like target placement, object attributes, and robot pose. These studies consistently find that high nominal success rates often decline sharply under minor perturbations, suggesting a reliance on memorization of training data rather than generalized skill acquisition.

However, such evaluations usually test isolated distribution shifts, like visual appearance or object position, rather than the coupled, multi-factor shifts encountered in real environments. Other works organize test tasks into progressively harder levels to probe model limits. For example, GemBench~\cite{ricardo2025towards} uses a multi-level protocol from novel placements to long-horizon tasks, observing performance degradation across levels. Yet its limited training scale and task coverage restrict its use for large-scale VLA assessment. Additionally, LIBERO-Plus~\cite{fei25libero-plus} stratifies difficulty via multi-model averaging, which can introduce bias and lacks interpretability.

Another limitation of many benchmarks~\cite{james2020rlbench, liu2023libero} is that training diversity is often secondary to test design, with assessments frequently conducted using off-the-shelf checkpoints~\cite{fei25libero-plus, zhou2025liberopro}. However, the fine-tuning datasets for these VLA models generally lack sufficient variation in scenes, tasks, and trajectories, thereby fundamentally constraining the generalization capacity of the resulting policies~\cite{zhang2025vlabench}. While LIBERO-Plus~\cite{fei25libero-plus} attempts to increase variation by re-rendering scenes using state-action pairs from the original LIBERO~\cite{liu2023libero}, it still inherits the limited diversity of trajectories.

In contrast, our benchmark integrates a highly diverse, teleoperation-collected training set with a multi-level evaluation protocol that systematically increases difficulty through spatial perturbations, object substitutions, and instruction rewrites. This approach enables a more faithful and systematic assessment of VLA models under complex, multi-dimensional distribution shifts, closely reflecting real-world conditions.

\section{LIBERO-X benchmark}
We introduce LIBERO-X, a comprehensive benchmark designed to advance the evaluation and development of VLA models for robotic manipulation. LIBERO-X provides a highly diverse training dataset and a multi-dimensional evaluation framework tailored to real-world complexity. Specifically, we first describe the training dataset collected via human teleoperation, which covers extensive task and scene variations. We then present the multi-level evaluation framework with fine-grained multi-label metrics, progressively incorporating spatial, object, and semantic perturbations. These dimensions capture the core factors governing perception, reasoning, and action coupling in VLA-based robotic manipulation, enabling a systematic assessment of model generalization and robustness across diverse environmental and task conditions. Following LIBERO~\cite{liu2023libero}, we employ the Planning Domain Definition Language (PDDL)~\cite{aeronautiques1998pddl} to describe scenes and tasks.

\begin{figure*}[htbp]
    \centering
    \includegraphics[width=\textwidth]{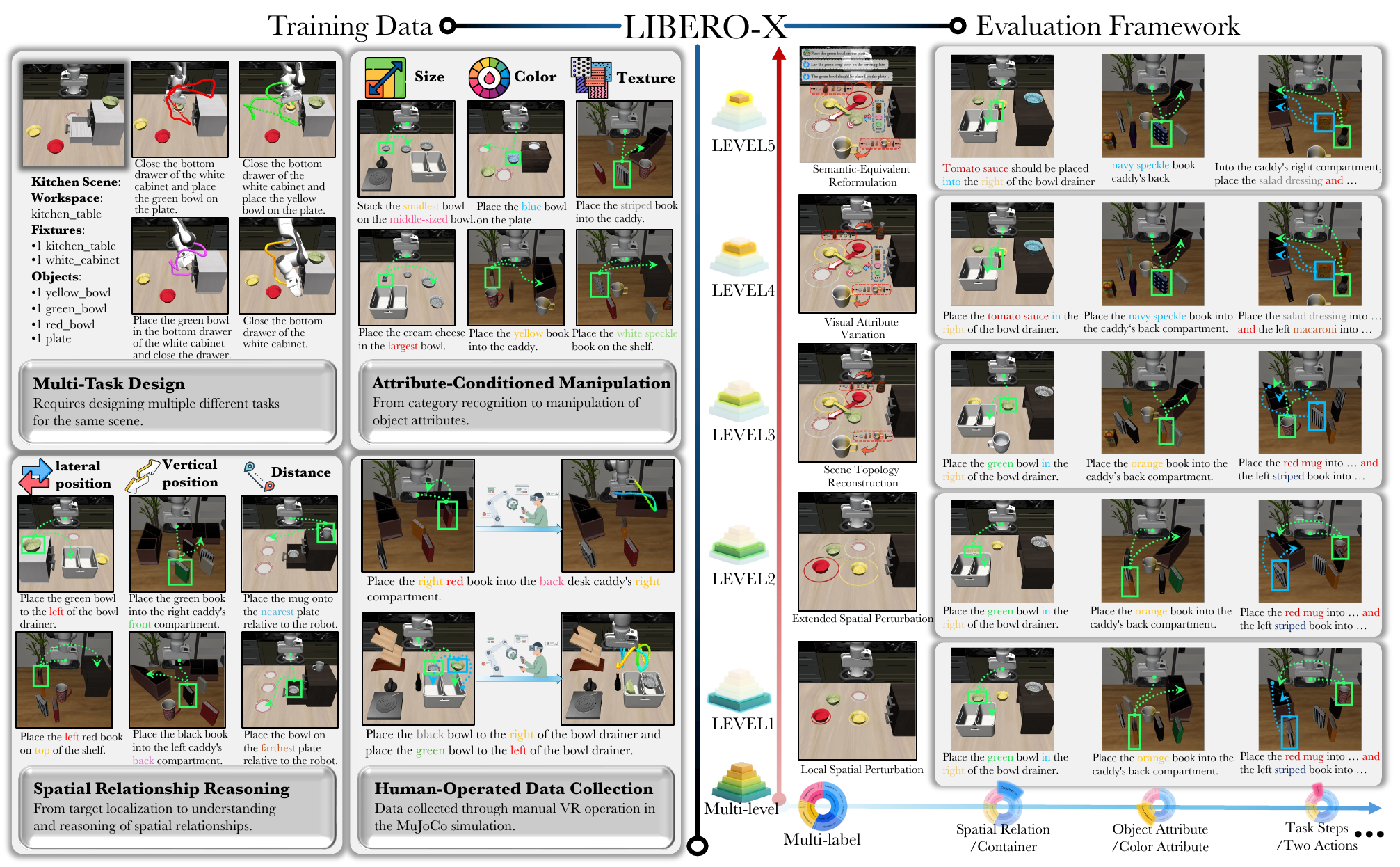}
    \caption{\textbf{Overview of LIBERO-X.} LIBERO-X provides a high-diversity training dataset constructed through human teleoperation, along with multi-level and multi-label evaluation data.}
    \vspace{-0.5cm}
    \label{fig:benchmark}
\end{figure*}
\subsection{Training Dataset}

\begin{figure}[htbp]
    \centering
    \includegraphics[width=\columnwidth]{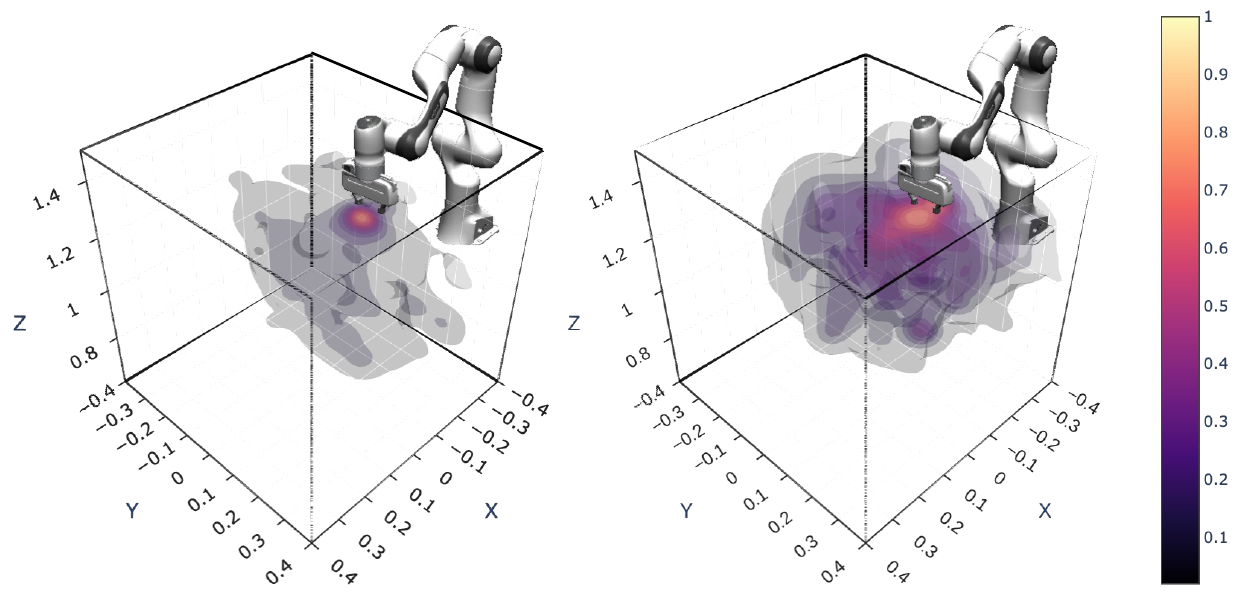}
    \raggedright
    \hspace*{25pt} \textbf{(a)} LIBERO \hspace{60pt} \textbf{(b)} LIBERO-X
    \caption{\textbf{Distribution visualization of training trajectories.}}
    \label{fig:trajectory}
\end{figure}

\begin{figure*}[htbp]
    \centering
    \begin{subfigure}{\textwidth}
        \centering
        \includegraphics[width=\textwidth]{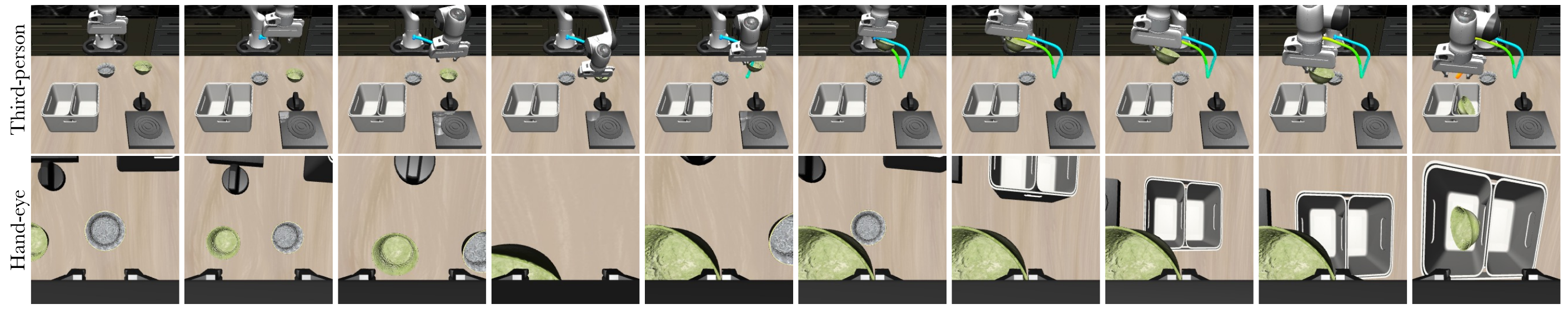}
        \caption{\textbf{Kitchen scene.} Task instruction: \textit{Place the green bowl to the right of the bowl drainer.}} 
        \label{fig:image_a}
    \end{subfigure}
    \vspace{0.5cm}
    \begin{subfigure}{\textwidth}
        \centering
        \includegraphics[width=\textwidth]{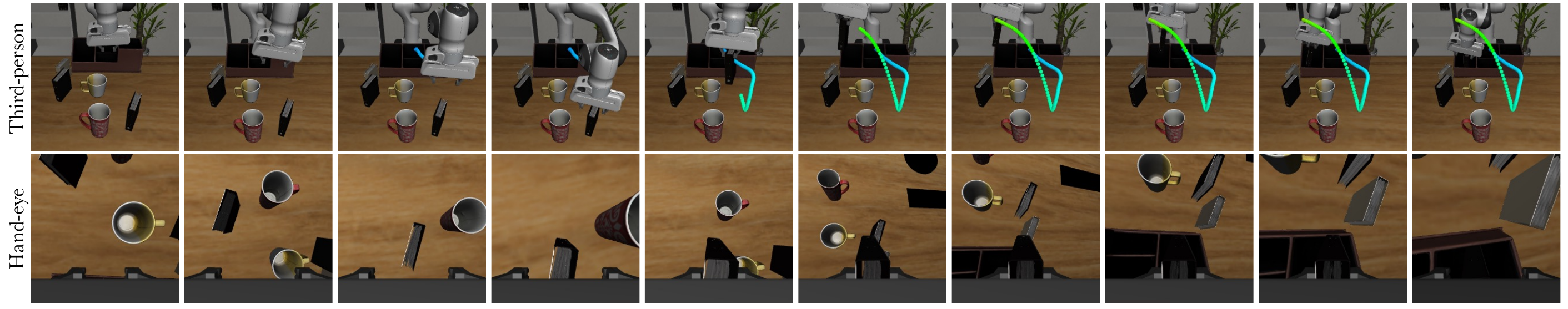}
        \caption{\textbf{Study scene.} Task instruction: \textit{Place the right black book into the desk caddy's left compartment.}}
        \label{fig:image_b}
    \end{subfigure}
    \vspace{-25pt}
    \caption{\textbf{Practical examples from the LIBERO-X training dataset.}}
    \vspace{-0.5cm}
    \label{fig:train_example}
\end{figure*}

To mitigate the overfitting risk arising from the limited scene diversity in the original LIBERO, where each scene is typically associated with a single task and the collected trajectories exhibit high similarity, LIBERO-X constructs a training dataset that better reflects real-world variability. Specifically, the dataset consists of 2,520 demonstration trajectories spanning 600 tasks and 100 scenes, as shown on the left side of Figure \ref{fig:benchmark}. Unlike the original LIBERO~\cite{liu2023libero}, which emphasizes manipulating targets by category thus encourages learning category-specific action patterns, LIBERO-X introduces finer-grained task-level extensions to expose models to diverse task formulations and workspace configurations:

\begin{itemize}
    \item \textbf{Multi-Task Scene Design}: In contrast to LIBERO~\cite{liu2023libero}, which is predominantly composed of single-task scenes, resulting in an average of 2.6 tasks per scene, our dataset significantly increases task density to an average of 6 tasks per scene. This design entails multi-object interactions and versatile manipulation skills, requiring models to adapt their manipulation strategies under identical visual and physical conditions.
    \item \textbf{Attribute-Conditioned Manipulation}: Actions are explicitly conditioned on fine-grained object properties (e.g., size, color, texture) rather than broad categories. To rigorously test perceptual grounding, scenes include distractor objects with similar attributes, requiring the model to distinguish the target amidst visual ambiguity.    
    \item \textbf{Spatial Relationship Reasoning}: Tasks extend beyond target localization to require understanding and reasoning about spatial relationships among objects, including left/right, front/back, and near/far.
\end{itemize}

With this training setup, the model is encouraged to learn the composable semantic correspondence between target objects and their language descriptions, rather than relying on template instructions or fixed scene structures to make “memorized” decisions. By increasing the diversity of tasks and scenes at the data source, LIBERO-X raises the upper bound of model performance, allowing models to truly acquire generalizable manipulation strategies. As a result, performance variations observed across subsequent LIBERO-X test levels more faithfully reflect the model’s real-world transferability and robustness in dynamic environments.

\subsubsection{Training Trajectory Collection}
The trajectories were collected via VR teleoperation using a Meta Quest 3 in MuJoCo~\cite{todorov2012mujoco}, facilitated by LeRobot~\cite{cadene2024lerobot}. During data collection, strict protocols were followed: state-action sequences were recorded at 20 Hz, where the state included the 6D Cartesian pose of the end effector and the gripper open/close status, and actions consisted of Cartesian increments and gripper commands. Multi-view images (third-person and hand-eye), were rendered by rolling out these recorded actions. To ensure quality, trajectories with noticeable pauses or non-optimal paths were discarded, retaining only smooth and successful demonstrations. As shown in Figure \ref{fig:trajectory}, heatmaps visualize the trajectory distributions, where brighter regions correspond to areas with higher trajectory density. Comparing with LIBERO, LIBERO-X exhibits a broader spread and higher trajectory density, demonstrating its greater diversity.

\subsubsection{Example Training Demonstrations}
The training dataset is built upon \textbf{Kitchen} and \textbf{Study} base scenes. By integrating diverse object categories, we created 100 scenarios encompassing 600 distinct tasks. These tasks cover object types, ranging from rigid to articulated objects, and include multi-object and attribute-conditioned interactions. Representative examples are illustrated in Figure \ref{fig:train_example}.

\subsection{Evaluation Framework}
The LIBERO-X test set employs a multi-level evaluation framework combined with a fine-grained multi-label system, forming a two-dimensional assessment design (Figure~\ref{fig:benchmark}). Along the vertical axis, the multi-level framework evaluates models on tasks of progressively increasing difficulty, creating a curriculum-style, incremental assessment. Along the horizontal axis, the multi-label system assigns detailed labels to tasks, enabling fine-grained quantitative analysis. Together, these two components allow systematic measurement of performance under various perturbations and across diverse tasks and environments. Table \ref{tab:compare} compares LIBERO-X with other simulation benchmarks, highlighting its multi-level framework and fine-grained labeling scheme.

\begin{table*}[!t]
\centering
\caption{
    \textnormal{\textbf{Comparison with other simulation benchmarks.} \textbf{Name:} Name of each benchmark. \textbf{Simulator:} ManiSkill2~\cite{gumaniskill2}, MuJoCo~\cite{todorov2012mujoco}, RLBench~\cite{james2020rlbench} and NVIDIA Isaac Sim~\cite{nvidia2023isaac}. \textbf{Date:} Release date. \textbf{Training Data:} Availability of training set. If provided, \includegraphics[width=0.017\textwidth]{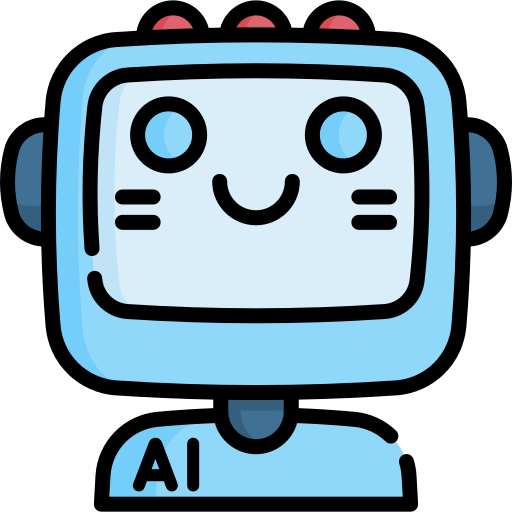} indicates automation, \includegraphics[width=0.017\textwidth]{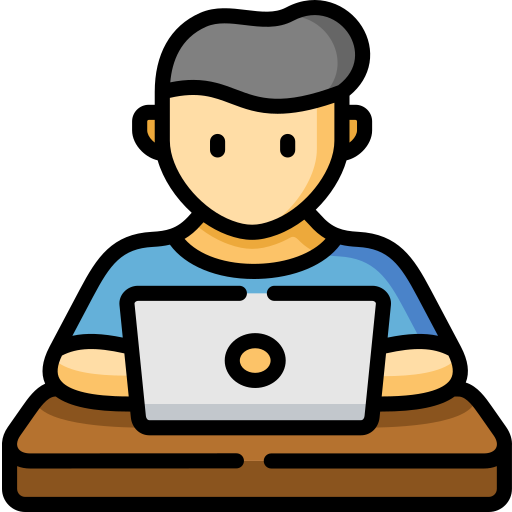} indicates manual annotation. \textbf{Testing Data:} Availability of independent testing set. \textcolor{darkgreen}{\ding{51}} represents full compliance, \textcolor{darkred}{\ding{55}} represents non-compliance, and \raisebox{-2pt}{\includegraphics[width=0.017\textwidth]{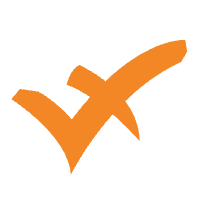}} indicates partial compliance. \textbf{Multi-level Evaluation:} Adoption of multi-level framework. L1: Local Spatial Perturbation, L2: Extended Spatial Perturbation, L3: Scene Topology Reconstruction, L4: Visual Attribute Variation, and L5: Semantic-Equivalent Reformulation. \textbf{Multi-label Evaluation:} Task label across multiple dimensions. \textbf{IT:} Interaction Type. \textbf{SC:} Subtask Count. \textbf{SR:} Spatial Relation. \textbf{OA:} Object Attribute.}
}
\setlength\tabcolsep{10pt}
\renewcommand{\arraystretch}{1.2} 
\resizebox{1.0\linewidth}{!}{
\begin{tabular}{lccccccccccccc}
\toprule
\multirow{2}{*}{\textbf{Name}} & \multirow{2}{*}{\textbf{Simulator}} & \multirow{2}{*}{\textbf{Date}} & \multirow{2}{*}{\textbf{Training Data}} & \multirow{2}{*}{\textbf{Testing Data}} & \multicolumn{5}{c}{\textbf{Multi-level Evaluation}} & \multicolumn{4}{c}{\textbf{Multi-label Evaluation}} \\
\cmidrule(lr){6-10} \cmidrule(lr){11-14}
 & & & & & \textbf{L1} & \textbf{L2} & \textbf{L3} & \textbf{L4} & \textbf{L5} & \textbf{IT} & \textbf{SC} & \textbf{SR} & \textbf{OA} \\
\toprule

RoboMIND~\cite{wu2024robomind} & Isaac Sim & 2024.12 & \includegraphics[width=0.017\textwidth]{image/human.png} & \textcolor{darkgreen}{\ding{51}} & \textcolor{darkred}{\ding{55}} & \textcolor{darkred}{\ding{55}} & \textcolor{darkred}{\ding{55}} & \raisebox{-2pt}{\includegraphics[width=0.017\textwidth]{image/half.png}} & \textcolor{darkred}{\ding{55}} & \textcolor{darkred}{\ding{55}} & \textcolor{darkred}{\ding{55}} & \textcolor{darkred}{\ding{55}} & \textcolor{darkred}{\ding{55}} \\

COLOSSEUM~\cite{pumacay2024colosseum} & RLBench & 2024.05 & \textcolor{darkred}{\ding{55}} & \textcolor{darkgreen}{\ding{51}} & \textcolor{darkred}{\ding{55}} & \textcolor{darkred}{\ding{55}} & \raisebox{-2pt}{\includegraphics[width=0.017\textwidth]{image/half.png}} & \raisebox{-2pt}{\includegraphics[width=0.017\textwidth]{image/half.png}} & \textcolor{darkred}{\ding{55}} & \textcolor{darkred}{\ding{55}} & \textcolor{darkred}{\ding{55}} &\textcolor{darkred}{\ding{55}} & \textcolor{darkgreen}{\ding{51}}  \\

AGNOSTOS~\cite{zhou2025exploring} & RLBench & 2025.10 & \textcolor{darkred}{\ding{55}} & \textcolor{darkgreen}{\ding{51}} & \textcolor{darkred}{\ding{55}} & \textcolor{darkred}{\ding{55}} & \textcolor{darkred}{\ding{55}} & \textcolor{darkred}{\ding{55}} & \textcolor{darkred}{\ding{55}} &\textcolor{darkred}{\ding{55}} & \raisebox{-2pt}{\includegraphics[width=0.017\textwidth]{image/half.png}} & \textcolor{darkred}{\ding{55}} & \textcolor{darkred}{\ding{55}} \\

GemBench~\cite{ricardo2025towards} & RLBench & 2025.03 & \includegraphics[width=0.017\textwidth]{image/robot.png} & \textcolor{darkgreen}{\ding{51}} & \raisebox{-2pt}{\includegraphics[width=0.017\textwidth]{image/half.png}}  & \textcolor{darkred}{\ding{55}} & \textcolor{darkred}{\ding{55}} & \raisebox{-2pt}{\includegraphics[width=0.017\textwidth]{image/half.png}}  & \textcolor{darkred}{\ding{55}} & \textcolor{darkred}{\ding{55}} & \raisebox{-2pt}{\includegraphics[width=0.017\textwidth]{image/half.png}} & \textcolor{darkred}{\ding{55}} & \raisebox{-2pt}{\includegraphics[width=0.017\textwidth]{image/half.png}} \\

VLATest~\cite{wang2025vlatest} & Maniskill2 & 2025.05 &  \textcolor{darkred}{\ding{55}} & \textcolor{darkgreen}{\ding{51}}  & \textcolor{darkgreen}{\ding{51}} & \textcolor{darkred}{\ding{55}}  &  \raisebox{-2pt}{\includegraphics[width=0.017\textwidth]{image/half.png}}  & \raisebox{-2pt}{\includegraphics[width=0.017\textwidth]{image/half.png}}& \raisebox{-2pt}{\includegraphics[width=0.017\textwidth]{image/half.png}} & \raisebox{-2pt}{\includegraphics[width=0.017\textwidth]{image/half.png}}& \raisebox{-2pt}{\includegraphics[width=0.017\textwidth]{image/half.png}}& \raisebox{-2pt}{\includegraphics[width=0.017\textwidth]{image/half.png}}&  \textcolor{darkred}{\ding{55}} \\

INT-ACT~\cite{fang2025intention} & Maniskill2 & 2025.06 & \textcolor{darkred}{\ding{55}} & \textcolor{darkgreen}{\ding{51}} & \raisebox{-2pt}{\includegraphics[width=0.017\textwidth]{image/half.png}}  & \textcolor{darkred}{\ding{55}} & \textcolor{darkred}{\ding{55}} & \textcolor{darkgreen}{\ding{51}} & \raisebox{-2pt}{\includegraphics[width=0.017\textwidth]{image/half.png}}  & \textcolor{darkred}{\ding{55}} & \textcolor{darkred}{\ding{55}} & \textcolor{darkred}{\ding{55}} & \textcolor{darkred}{\ding{55}} \\

RL4VLA~\cite{liu2025can} & Maniskill2 & 2026.01 & \includegraphics[width=0.017\textwidth]{image/robot.png} & \textcolor{darkgreen}{\ding{51}}  & \textcolor{darkgreen}{\ding{51}}  & \textcolor{darkgreen}{\ding{51}}  & \textcolor{darkgreen}{\ding{51}} & \textcolor{darkred}{\ding{55}} & \raisebox{-2pt}{\includegraphics[width=0.017\textwidth]{image/half.png}}  & \raisebox{-2pt}{\includegraphics[width=0.017\textwidth]{image/half.png}}& \raisebox{-2pt}{\includegraphics[width=0.017\textwidth]{image/half.png}}& \raisebox{-2pt}{\includegraphics[width=0.017\textwidth]{image/half.png}}& \raisebox{-2pt}{\includegraphics[width=0.017\textwidth]{image/half.png}} \\

LIBERO~\cite{liu2023libero} & MuJoCo & 2023.06 & \includegraphics[width=0.017\textwidth]{image/human.png} & \textcolor{darkgreen}{\ding{51}} & \textcolor{darkgreen}{\ding{51}} & \textcolor{darkred}{\ding{55}} & \textcolor{darkred}{\ding{55}} & \textcolor{darkred}{\ding{55}} & \textcolor{darkred}{\ding{55}} & \textcolor{darkred}{\ding{55}} &\textcolor{darkred}{\ding{55}} &\textcolor{darkred}{\ding{55}} &\textcolor{darkred}{\ding{55}} \\

LIBERO-OOD~\cite{li2025task} & MuJoCo & 2025.05 & \textcolor{darkred}{\ding{55}} & \textcolor{darkgreen}{\ding{51}} &\textcolor{darkred}{\ding{55}} & \textcolor{darkred}{\ding{55}} & \raisebox{-2pt}{\includegraphics[width=0.017\textwidth]{image/half.png}} &\raisebox{-2pt}{\includegraphics[width=0.017\textwidth]{image/half.png}} & \textcolor{darkred}{\ding{55}} & \textcolor{darkred}{\ding{55}} & \textcolor{darkred}{\ding{55}} &  \textcolor{darkred}{\ding{55}} & \textcolor{darkred}{\ding{55}} \\

RoboCerebra~\cite{han2025robocerebra} & MuJoCo & 2025.06 & \includegraphics[width=0.017\textwidth]{image/human.png} & \textcolor{darkgreen}{\ding{51}} & \textcolor{darkred}{\ding{55}} &\raisebox{-2pt}{\includegraphics[width=0.017\textwidth]{image/half.png}} &\raisebox{-2pt}{\includegraphics[width=0.017\textwidth]{image/half.png}} & \textcolor{darkred}{\ding{55}} & \textcolor{darkred}{\ding{55}} & \textcolor{darkred}{\ding{55}} & \textcolor{darkred}{\ding{55}} & \textcolor{darkred}{\ding{55}} & \textcolor{darkred}{\ding{55}}  \\

LIBERO-PRO~\cite{zhou2025liberopro} & MuJoCo & 2025.10 & \textcolor{darkred}{\ding{55}} & \textcolor{darkgreen}{\ding{51}} & \textcolor{darkred}{\ding{55}} & \textcolor{darkred}{\ding{55}} & \raisebox{-2pt}{\includegraphics[width=0.017\textwidth]{image/half.png}} & \raisebox{-2pt}{\includegraphics[width=0.017\textwidth]{image/half.png}} &\raisebox{-2pt}{\includegraphics[width=0.017\textwidth]{image/half.png}} &\raisebox{-2pt}{\includegraphics[width=0.017\textwidth]{image/half.png}} &\textcolor{darkgreen}{\ding{51}}  & \raisebox{-2pt}{\includegraphics[width=0.017\textwidth]{image/half.png}} &\raisebox{-2pt}{\includegraphics[width=0.017\textwidth]{image/half.png}}\\

LIBERO-Plus~\cite{fei25libero-plus} & MuJoCo & 2025.10 & \includegraphics[width=0.017\textwidth]{image/robot.png} & \textcolor{darkgreen}{\ding{51}} & \textcolor{darkred}{\ding{55}} & \textcolor{darkgreen}{\ding{51}} & \textcolor{darkred}{\ding{55}}  & \textcolor{darkred}{\ding{55}} & \raisebox{-2pt}{\includegraphics[width=0.017\textwidth]{image/half.png}} & \raisebox{-2pt}{\includegraphics[width=0.017\textwidth]{image/half.png}} & \raisebox{-2pt}{\includegraphics[width=0.017\textwidth]{image/half.png}} & \textcolor{darkred}{\ding{55}} & \textcolor{darkred}{\ding{55}}\\

SafeLIBERO~\cite{hu2025vlsa} & MuJoCo & 2025.12 & \textcolor{darkred}{\ding{55}} & \textcolor{darkgreen}{\ding{51}} & \textcolor{darkred}{\ding{55}} & \raisebox{-2pt}{\includegraphics[width=0.017\textwidth]{image/half.png}} & \textcolor{darkred}{\ding{55}} & \raisebox{-2pt}{\includegraphics[width=0.017\textwidth]{image/half.png}} & \textcolor{darkred}{\ding{55}} & \textcolor{darkred}{\ding{55}} & \textcolor{darkred}{\ding{55}} & \textcolor{darkred}{\ding{55}} & \textcolor{darkred}{\ding{55}}\\

\midrule
LIBERO-X (Ours) & MuJoCo & 2026.01 & \includegraphics[width=0.017\textwidth]{image/human.png} & \textcolor{darkgreen}{\ding{51}} & \textcolor{darkgreen}{\ding{51}} &\textcolor{darkgreen}{\ding{51}} &\textcolor{darkgreen}{\ding{51}} & \textcolor{darkgreen}{\ding{51}} & \textcolor{darkgreen}{\ding{51}} & \textcolor{darkgreen}{\ding{51}} & \textcolor{darkgreen}{\ding{51}} & \textcolor{darkgreen}{\ding{51}} & \textcolor{darkgreen}{\ding{51}} \\

\bottomrule
\end{tabular}}

\vspace{-0.5cm}
\label{tab:compare}
\end{table*}

\subsubsection{Multi-level Evaluation}
The vertical axis of the framework implements a multi-level evaluation following a progressive scene design that incrementally increases task and environmental complexity across spatial reasoning, object recognition, and instruction understanding. Each level builds upon the preceding one by superimposing additional perturbations, ensuring a cumulative assessment of the model’s robustness:

\textbf{Level 1 - Local Spatial Perturbation:} Test scenes closely resemble the training distribution, with minor variations in object positions, designed to measure model stability under small-scale spatial changes.  

\textbf{Level 2 - Extended Spatial Perturbation:} Building upon Level 1, spatial variations are amplified by dynamically adjusting the sampling radius according to object proximity, with a safety margin to expand layout space. This level assesses the model’s ability to cope with large-scale positional changes.  

\textbf{Level 3 - Scene Topology Reconstruction:} The relative positions of objects are altered, including swapping targets with distractors and adding irrelevant objects, to increase environmental complexity. This stage evaluates task execution under restructured object relationships and modified scene configurations.  

\textbf{Level 4 - Visual Attribute Variation:} Objects vary in texture, color, and size; target objects may belong to unseen categories, and visually similar confounders are introduced. This level tests the model’s discriminative ability, robustness to attribute changes, and generalization to unseen objects. 

\textbf{Level 5 - Semantic-Equivalent Reformulation:} Instructions are rephrased while preserving task semantics, including synonym replacement (L5-1), word compression (L5-2), word order adjustment (L5-3), voice conversion (L5-4), and redundant descriptions (L5-5). This level measures the model’s ability to generalize across diverse linguistic formulations of the same task.



\subsubsection{Multi-label Evaluation}

The horizontal axis of the framework incorporates a fine-grained multi-label evaluation system. This system enables a detailed, multi-dimensional analysis of model performance, particularly regarding its ability to manage fine-grained aspects of task execution. The multi-label evaluation is organized hierarchically, with each top-level category further divided into sub-categories (detailed definitions are provided in the appendix).

\textbf{Interaction Type:} This dimension categorizes object interactions into three types: (i) Pick-and-Place, where objects are grasped and moved to targets; (ii) Switch-based, such as opening/closing drawers or turning knobs, requiring precise gripper control; and (iii) Combined, integrating both types and requiring flexible transitions. It evaluates the model’s ability to execute diverse manipulation strategies.

\textbf{Subtask Count:} This dimension captures the number of subtasks required to complete a task, ranging from single-step to three-step sequences. It assesses the model’s planning, sequential reasoning, and ability to maintain task coherence over long-horizon tasks.

\textbf{Spatial Relation:} This dimension characterizes the relative positions and spatial arrangements of objects within a task. It evaluates the model’s spatial reasoning and environmental perception by measuring how effectively it interprets spatial relationships and their impact on task execution.

\textbf{Object Attribute:} This dimension describes the visual object properties, including color, size, texture, and other fine-grained features. It evaluates the model’s ability to recognize and leverage object attributes to accomplish tasks, emphasizing the role of precise visual perception in task success.

\section{Experiments}

\subsection{Experimental setup}

We selected five representative models for evaluation: OpenVLA-OFT~\cite{kim2025fine}, X-VLA~\cite{zheng2025xvla}, GR00T-N1.5~\cite{bjorck2025gr00t}, $\pi_0$~\cite{black2410pi0} and $\pi_{0.5}$~\cite{intelligence2025pi05}. Architecturally, these methods share a unified paradigm where a VLM backbone is utilized to extract latent representations from visual observations and language instructions, followed by an action head to generate the control sequences. All models were evaluated using their official configurations. Specifically, we employed a behavior cloning approach by performing supervised fine-tuning (SFT) on our self-collected training dataset. Detailed configurations for each model are listed in the appendix. To ensure reproducibility, we recorded the initial state configurations, including the placement and status of all objects, to guarantee consistent starting conditions across runs. Each task was evaluated over 10 trials with slight spatial perturbations applied to object positions during rollouts, following the LIBERO benchmark protocol~\cite{liu2023libero}. Furthermore, the task time limit was dynamically set to 1.1 times the average human completion time, scaling with task difficulty. This temporal buffer accounts for minor deviations in execution speed, such as a model approaching a target more slowly than a human demonstrator, thereby preventing the unfair penalization of near-complete trajectories while maintaining realistic evaluation standards.

\begin{table}[!ht] 
     \caption{
        \textbf{Accuracy under Multi-level Evaluation.}
    }
    \centering
    \small
    \renewcommand\tabcolsep{6pt} 
    \renewcommand\arraystretch{1.0} 
    \resizebox{1.0\linewidth}{!}{ 
        \begin{tabular}{c|c|c|c|c|c}
            \toprule
            \rowcolor{gray!25} 
            \textbf{Model}
            & \textbf{LEVEL 1}
            & \textbf{LEVEL 2}
            & \textbf{LEVEL 3}
            & \textbf{LEVEL 4}
            & \textbf{LEVEL 5} \\
            \midrule
            \multirow{1}{*}{OpenVLA-OFT} & 29.0 & 17.6 & 8.8 & 6.4 & 4.2 \\
            \multirow{1}{*}{$\pi_{0}$} & 29.4 & 21.9 & 11.0 & 7.6 & 5.1 \\
            \multirow{1}{*}{X-VLA} & 30.1 & 22.6 & 10.3 & 6.0 & 4.1 \\
            \multirow{1}{*}{GR00T1.5} & 43.3 & 32.9 & 18.7 & 13.3 & 9.7 \\
            \multirow{1}{*}{$\pi_{0.5}$} & 65.2 & 53.2 & 36.0 & 24.1 & 18.0 \\

            \bottomrule
        \end{tabular}
    }
    \vspace{-6pt} 
   
    \label{tab:level_results}
\end{table}

\begin{figure}[htbp]
    \centering
    \includegraphics[width=\columnwidth]{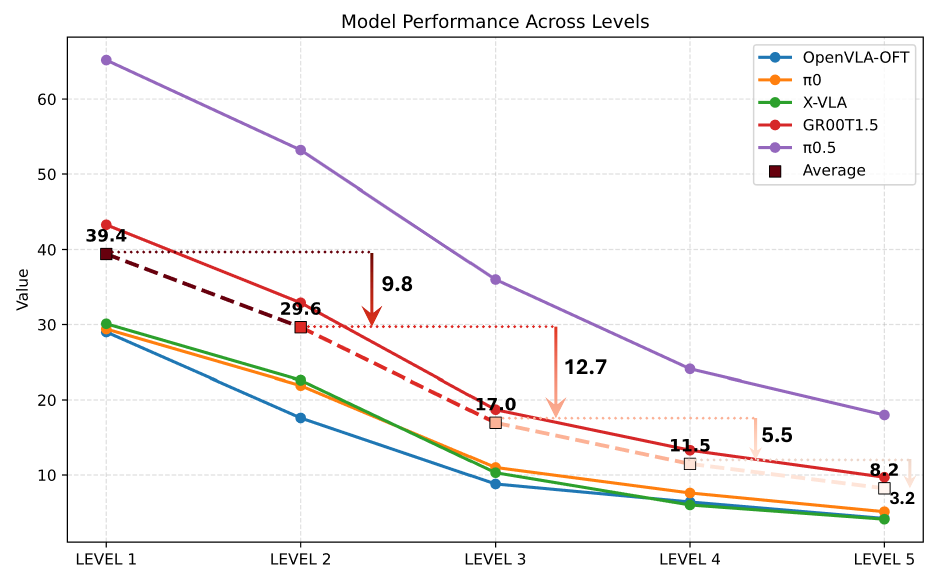}
    \caption{\textbf{Success Rate Decline across Multi-level Evaluation.}}
    \label{fig:level_results_svg}
\end{figure}

\subsection{Multi-level Evaluation}

\begin{figure*}[htbp]
    \centering
    \includegraphics[width=\textwidth]{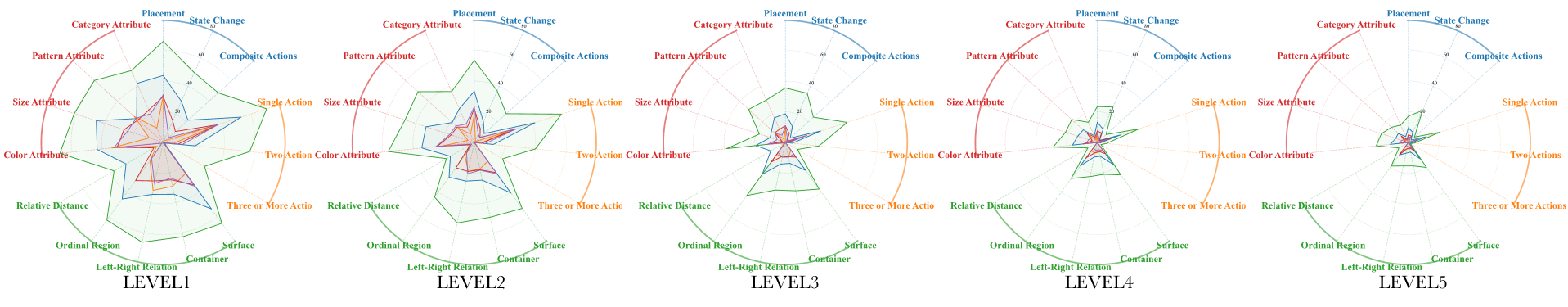}
    \vspace{-10pt}
    \caption{\textbf{Multi-label Evaluation.} 
    \textcolor[RGB]{44,160,44}{\rule{5px}{5px}} denotes $\pi_{0.5}$,
    \textcolor[RGB]{31,119,180}{\rule{5px}{5px}} GR00T1.5,
    \textcolor[RGB]{214,39,40}{\rule{5px}{5px}} $\pi_0$,
    \textcolor[RGB]{148,103,189}{\rule{5px}{5px}} X-VLA,
    and \textcolor[RGB]{255,127,14}{\rule{5px}{5px}} OpenVLA-OFT.
}
    \label{fig:radar}
\end{figure*}

\begin{figure*}[htbp]
    \centering
    \includegraphics[width=\textwidth]{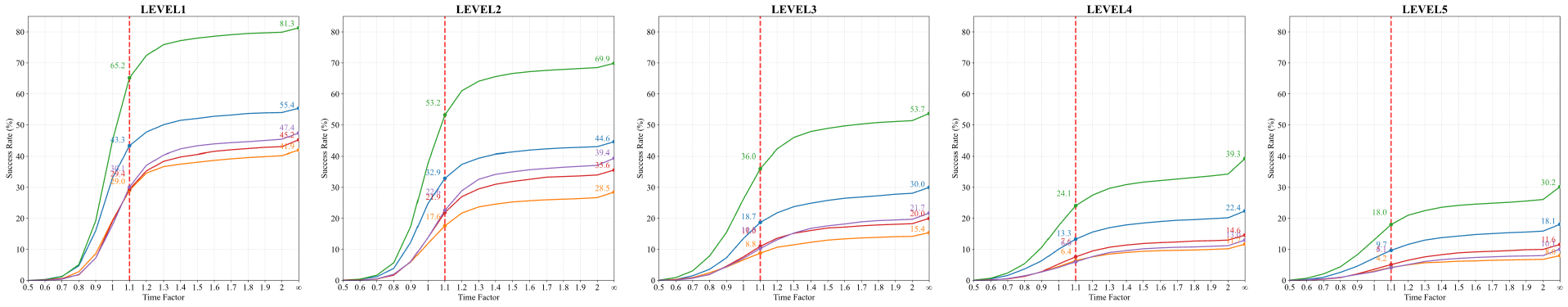}
    \caption{\textbf{Impact of Time Factor Thresholds on Success Rate.} The x-axis displays time factor multipliers relative to the manual standard, where \(\infty\) indicates denotes an unbounded threshold.}
    \label{fig:time_factor}
\end{figure*}
Table \ref{tab:level_results} summarizes the evaluation results of the models under the multi-level framework. Notably, in contrast to the near-saturated performance (often $\sim$90\%)~\cite{fei25libero-plus, zhou2025liberopro} in LIBERO~\cite{liu2023libero}, the models exhibit significant struggles in our setting. Even at Level 1, which involves only minor spatial perturbations relative to the training set, the average success rate across the five models is merely 39.4\%, with even the top-performing $\pi_{0.5}$ capped at 65.2\%. This sharp contrast not only underscores the challenging nature of our benchmark but also exposes the limitations of existing evaluation systems, where performance saturation often masks true model differences. By leveraging diverse training data, our benchmark effectively reveals the performance disparities among different methods.

Despite the overall difficulty, $\pi_{0.5}$ maintains the most robust generalization across all levels. We attribute this robustness to its joint training strategy, which effectively incorporates pre-training on the Internet-scale data and sub-task planning. However, for all models, a significant downward trend is observed as task complexity increases. With the introduction of spatial, object-level, and semantic variations from Level 1 to Level 5, the models exhibit an average performance decline of 31.2\%. These results highlight the increasing difficulty of the tasks and provide insights into the models' sensitivity to high-level perturbations. Based on these quantitative evaluations, we summarize our key findings as follows:

\textbf{Finding 1: VLA models struggle with spatial extrapolation beyond training distributions.} While models maintain reasonable performance under minor jitters (Level 1), the transition to Level 2, characterized by larger spatial shifts, triggers a notable 9.8\% drop in average success rate (Figure~\ref{fig:level_results_svg}). This decline suggests that current VLM-based policies tend to overfit to the specific spatial configurations in the training data, relying on memorized pixel-level patterns rather than acquiring a robust spatial understanding. Consequently, performance degrades significantly when targets are extrapolated to out-of-distribution positions.

\textbf{Finding 2: Structural invariance is lacking when facing topological perturbations.} Beyond mere coordinate shifts, Level 3 challenges the models' understanding of scene topology by altering the relative arrangement of objects (e.g., swapping positions). Crucially, as shown in Figure~\ref{fig:level_results_svg}, this level witnesses the most substantial performance degradation of 12.7\%, indicating that models lack structural invariance. They struggle to adapt to new scene layouts where the semantic relationships between objects remain valid but their topological organization has been reconstructed, highlighting a failure in reasoning about relative spatial relations.
\begin{table}[!ht] 
    \caption{
        \textbf{Level 4 Accuracy - Overall, Unseen Objects (UO), and Confounding Objects (CO).}
    }
    \centering
    \small
    \renewcommand\tabcolsep{6pt} 
    \renewcommand\arraystretch{1.0} 
    \resizebox{1.0\linewidth}{!}{ 
        \begin{tabular}{c|c|c|c|c|c}
            \toprule
            \rowcolor{gray!25} 
            \textbf{Category}
            & OpenVLA-OFT
            & $\pi_{0}$
            & X-VLA
            & GR00T1.5
            & $\pi_{0.5}$ \\
            \midrule
            \multirow{1}{*}{\textbf{Overall}} & 6.4 & 7.6 & 6.0 & 13.3 & 24.1 \\
            \multirow{1}{*}{\textbf{UO}} & 0.8 & 1.1 & 0.8 & 2.1 & 7.2 \\
            \multirow{1}{*}{\textbf{CO}} & 8.6 & 9.6 & 7.5 & 16.9 & 28.9 \\
            \bottomrule
        \end{tabular}
    }
    \vspace{-6pt} 
    \label{tab:level4}
\end{table}

\textbf{Finding 3: Data diversity enables emergent generalization, though semantic grounding for novel objects remains a bottleneck.} While models trained on standard datasets typically fail on out-of-distribution tasks~\cite{liu2023libero}, our diversified LIBERO-X dataset enables the model to maintain non-zero success rates, suggesting that diversity helps mitigate overfitting to specific training scenes. However, a closer inspection of Table \ref{tab:level4} reveals a significant performance gap: success rates on confounding objects are notably higher than those on unseen objects. This disparity indicates that while the model gains robustness against visual distractions, it still struggles to ground instructions to novel visual concepts. The lower performance on unseen objects suggests that despite the VLM's capabilities, the alignment between pre-trained representations and the control policy remains imperfect for handling unfamiliar visual features.
\begin{table}[!ht] 
    \caption{
        \textbf{Level 5 Accuracy - Overall and by Semantic Equivalence Reconstruction Method.}
    }
    \centering
    \small
    \renewcommand\tabcolsep{6pt} 
    \renewcommand\arraystretch{1.0} 
    \resizebox{1.0\linewidth}{!}{ 
        \begin{tabular}{l|c|c|c|c|c}
            \toprule
            \rowcolor{gray!25} 
            \textbf{Category}
            & OpenVLA-OFT
            & $\pi_{0}$
            & X-VLA
            & GR00T1.5
            & $\pi_{0.5}$ \\
            \midrule
            \multicolumn{1}{c|}{\textbf{Overall}} & 4.2 & 5.1 & 4.1 & 9.7 & 18.0 \\
            \textbf{5-1 (Synonym)} & 5.1 & 3.9 & 4.3 & 8.7 & 18.0 \\
            \textbf{5-2 (Concise)} & 5.0 & 5.0 & 4.2 & 8.3 & 14.6 \\
            \textbf{5-3 (Reorder)} & 4.4 & 5.1 & 3.0 & 8.4 & 18.0 \\
            \textbf{5-4 (Voice)} & 5.2 & 7.5 & 5.1 & 10.8 & 19.6 \\
            \textbf{5-5 (Verbose)} & 1.2 & 4.2 & 3.9 & 12.5 & 19.6 \\
            \bottomrule
        \end{tabular}
    }
    \vspace{-6pt} 
    \label{tab:level5}
\end{table}

\textbf{Finding 4: Language instruction variations can influence model performance.} Table \ref{tab:level5} reveals a modest 3.26\% drop in success rate from Level 4 to Level 5, indicating that phrasing changes affect execution. Crucially, voice conversion proved to be the least disruptive variation, achieving the highest success rate among all rephrasing methods. This implies that the model maintains better alignment with instructions under syntactic voice shifts compared to other forms of linguistic conversion.

\subsection{Multi-label Evaluation}
Figure \ref{fig:radar} visualizes success rates for task labels across five levels. Macroscopically, the progressive shrinking of the enclosed area from Level 1 to 5 illustrates general performance degradation due to escalating complexity. Crucially, we observe a consistent alignment in performance patterns. Within the same level, methods exhibit synchronized variations; dimensions challenging for one model tend to be universally difficult, reflecting inherent task difficulty. Furthermore, across levels, the relative performance hierarchy remains stable. The degradation manifests as a proportional contraction, where models maintain comparative strengths and weaknesses even as absolute success rates decline.

\textbf{Finding 5: Task horizon length critically limits performance.} As illustrated by the “Subtask Count” category (orange axes \textcolor[RGB]{255,127,14}{\rule{5px}{5px}}), models perform best on single-step tasks but degrade sharply as sequence length increases, dropping to near zero for tasks with three or more steps. Notably, for tasks requiring three or more steps, success rates for nearly all models precipitate towards zero. This non-linear performance drop-off suggests that current VLA models struggle with the compounding errors inherent in sequential manipulation. It reveals a fundamental deficiency in long-horizon reasoning and temporal consistency, where models fail to maintain precise execution over extended periods. Consequently, future research must prioritize error mitigation and long-term planning to enable robust multi-step manipulation.

\subsection{Time Limit Threshold Differences Evaluation}
Figure \ref{fig:time_factor} illustrates the impact of time limits set relative to the manual standard, which is defined as the average duration of human operation. For instance, by applying multipliers such as 0.8 (stringent) and 1.5 (relaxed) to this baseline, we evaluate the model's adaptability to varying temporal pressures ranging from 0.8 to 1.5 times the human average.

\textbf{Finding 6: Relaxed time limits buffer execution imperfections, though benefits diminish beyond a saturation point.} As shown in Figure \ref{fig:time_factor}, stringent constraints (0.8$\times$) severely penalize performance. We attribute this to inevitable inference or control errors that result in deviations from the optimal trajectory, rendering most task completion impossible under tight deadlines. Conversely, extending limits yields substantial gains, particularly between the 1.0$\times$ and 1.1$\times$, justifying our default selection of 1.1$\times$. We avoid excessive limits (e.g., 1.5$\times$ or beyond) to prevent success via brute-force trial-and-error. Notably, performance plateaus beyond 1.3$\times$, suggesting that remaining failures arise from fundamental capability deficits like perceptual hallucinations or logic errors rather than temporal insufficiency, thus further time extension cannot compensate for these intrinsic limitations.

\section{Conclusion} 
\label{sec:conclusion}

This work introduces LIBERO-X, a comprehensive benchmark for evaluating vision–language–action models under realistic, multi-dimensional distribution shifts. By combining a hierarchically structured evaluation protocol with a high-diversity teleoperation-based training dataset, LIBERO-X enables systematic analysis of model generalization across spatial layouts, object attributes, and task semantics. Experimental results reveal significant performance degradation as complexity increases, exposing limitations in scene understanding and instruction grounding. Overall, LIBERO-X provides a more faithful and rigorous framework for assessing VLA models and guiding future research toward robust robotic deployment.



\bibliographystyle{plainnat}
\bibliography{references}

\clearpage
\appendix
The supplementary materials contain the following sections:

\begin{itemize}
    \item \textbf{Detailed Results of Multi-label Evaluation} \ref{sec:multi_label_evaluation}.
    \item \textbf{Further Analysis of the Impact of Time Limits} \ref{sec:time_limits_analysis}.
    \item \textbf{Model Details} \ref{sec:model_details}.
    \item \textbf{Scene Design Details} \ref{sec:scene_design}.
    \item \textbf{Visual Attribute Variation Cases Analysis} \ref{sec:visual_attribute_variation}.
    \item \textbf{Semantic-equivalent Reformulation Cases} \ref{sec:semantic_reformulation}.
\end{itemize}

\subsection{Detailed Results of Multi-label Evaluation} 
\label{sec:multi_label_evaluation}
\begin{figure}[htbp]
    \centering
    \includegraphics[width=0.95\columnwidth]{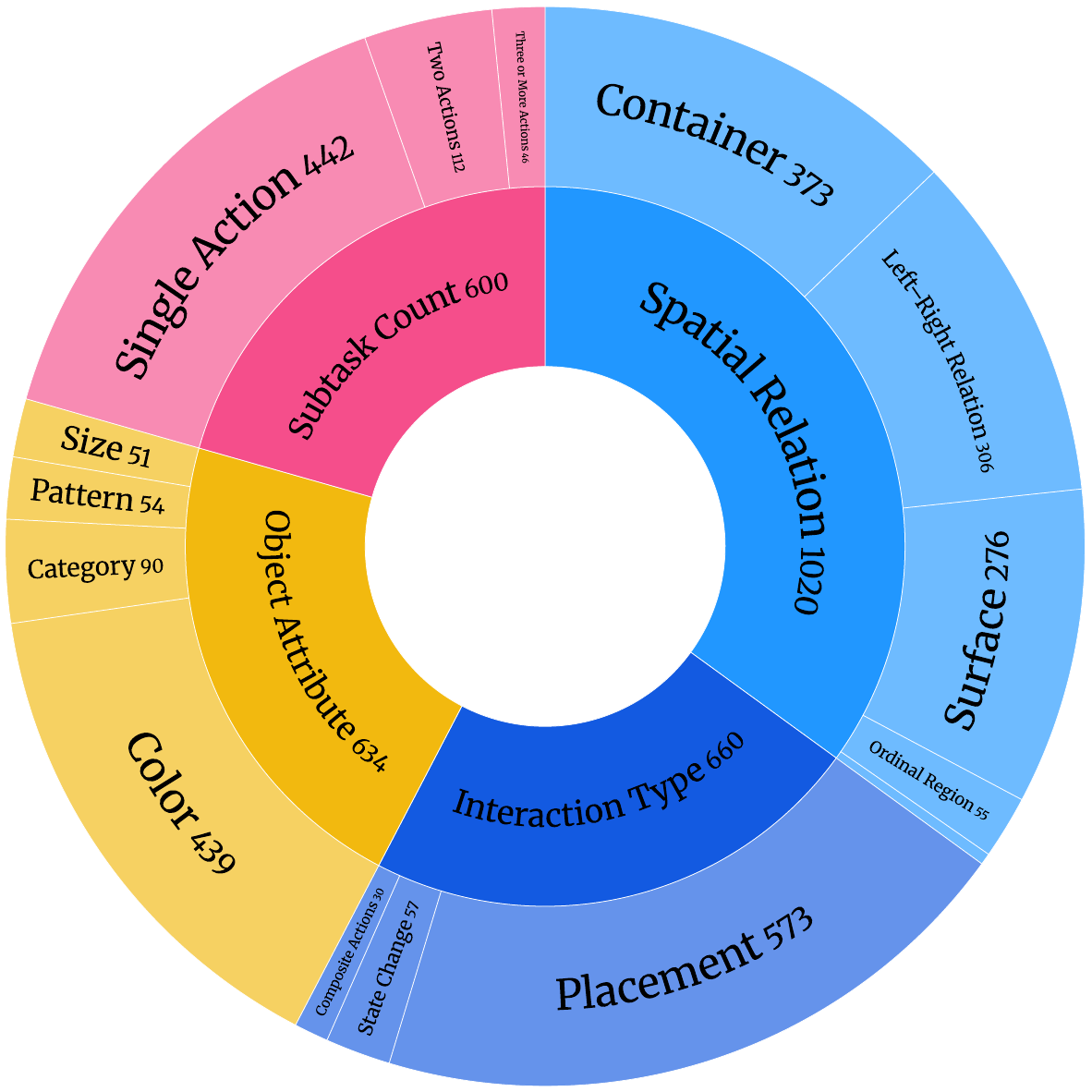} 
    \caption{\textbf{Fine-grained Multi-label Evaluation in LIBERO-X.}}
    \label{fig:label}
\end{figure}

To comprehensively assess the capabilities of VLA models, LIBERO-X introduces a fine-grained evaluation system that assigns specific labels to all evaluation tasks. These labels allow for a detailed analysis of the model's performance across multiple dimensions, particularly focusing on task execution details. The evaluation covers various aspects, including \textbf{Interaction Type}, \textbf{Subtask Count}, \textbf{Spatial Relation Type}, and \textbf{Object Attribute Recognition}, which provide insights into how the model handles semantic and physical changes in different scenarios.

As shown in Figure \ref{fig:label}, the fine-grained multi-label evaluation framework adopts a dual-layer design, divided into four major categories. Each category is further subdivided into multiple subcategories, with the number of subcategories displayed alongside. This structure enables a more comprehensive evaluation, allowing LIBERO-X to assess the model's ability to handle different task dimensions and providing deeper insights into its performance under various conditions.

\textbf{(1) Interaction Type} aims to describe the core actions involved in the task, reflecting the basic behaviors and complexity of task execution. Each interaction type represents a key operation the model performs when handling a task, helping evaluate its performance under different action modes.

\begin{itemize}
    \item \textbf{Placement}: Entails positioning an object onto a target surface or within a container. This category evaluates the model’s precision and stability in executing basic manipulation tasks.
    \item \textbf{State Change}: Involves altering the environment's state, such as manipulating articulated objects (e.g., opening drawers or turn the knob). This assesses the model's capability to perform state transitions effectively.
    \item \textbf{Composite Actions}: These tasks involve multiple action types simultaneously, such as placing followed by closing, or grabbing followed by placing. These tasks test the model's ability to coordinate multi-step operations and handle complex planning.
\end{itemize}

\textbf{(2) Subtask Count} dimension categorizes tasks by the number of required action steps, assessing the model's capacity for long-horizon planning. Varying the sequence length allows for a granular evaluation of decision-making capabilities across different complexity levels.

\begin{itemize}
    \item \textbf{Single Action}: Involves a single atomic operation without intermediate steps. This serves as a baseline benchmark for the model’s fundamental execution capability.
    \item \textbf{Two Actions}: Requires performing two sequential actions (e.g., placing an object and then closing a drawer). This evaluates the model's ability to handle short-term transitions and maintain logical consistency between consecutive steps.
    \item \textbf{Three or More Actions}: Involves complex sequences consisting of multiple operations (e.g., grasping, transporting, and placing). These tasks evaluate the model's robustness in long-horizon reasoning and its ability to mitigate error accumulation over extended task durations.
\end{itemize}

\textbf{(3) Spatial Relation Type} label describes the relative position and spatial layout of objects in the task, assessing the model’s performance in spatial reasoning and environmental perception. The model must understand the spatial relationships between objects and how these relationships impact task execution.

\begin{itemize}
    \item \textbf{Surface}: Involves placing objects onto open planar surfaces, such as tables or countertops. This evaluates the model's precision in identifying and utilizing flat substrates for placement.
    \item \textbf{Container}: Involves placing objects into enclosed volumes, such as drawers or cabinets. This assesses the model’s ability to navigate constrained spaces and execute insertion tasks accurately.
    \item \textbf{Left–Right Relation}: Specifies positioning relative to a lateral reference frame (e.g., the left side of a desk caddy). This tests the model’s lateral spatial reasoning and its ability to interpret directional instructions.
    \item \textbf{Ordinal Region}: Requires selecting a target based on sequential order or rank, such as the top, middle, or bottom drawer. This evaluates the model's understanding of hierarchical spatial relationships.
    \item \textbf{Relative Distance}: Involves selecting objects based on proximity to a reference point (e.g., the closest or farthest object). This assesses the model’s depth perception and ability to estimate relative distances.
    
\end{itemize}

\textbf{(4) Object Attribute Recognition} label assesses the model's visual perception accuracy regarding object properties. It focuses on the model's ability to ground textual instructions to specific visual attributes, such as color, size, pattern, and category, to correctly identify targets in the environment.

\begin{itemize}
    \item \textbf{Color}: Evaluates the ability to distinguish objects based on chromatic features. The model must accurately identify target objects by hue, even in scenes containing distractors with similar or contrasting color.
    \item \textbf{Size}: Assesses spatial reasoning regarding object scale. The model is tested on its ability to differentiate relative sizes (e.g., small, medium, large) and select objects that satisfy specific spatial constraints.
    \item \textbf{Pattern}: Focuses on the recognition of surface details and textures (e.g., stripes, spots). This tests the model's fine-grained visual discrimination capabilities, particularly when identifying objects in visually complex or cluttered scenes.
    \item \textbf{Category}: Involves semantic classification of objects. The evaluation focuses on the model's ability to correctly identify and select objects based on their class identity, distinguishing the target from multiple distractors of various categories.
\end{itemize}

As shown in Tables \ref{tab:app_multi_label_level1}, \ref{tab:app_multi_label_level2}, \ref{tab:app_multi_label_level3}, \ref{tab:app_multi_label_level4}, and \ref{tab:app_multi_label_level5}, the detailed results of Fine-grained Multi-label Evaluation across Levels 1 to 5 demonstrate the model's performance under varying task complexities and evaluation dimensions, highlighting its execution capabilities and adaptability when faced with diverse tasks.

\begin{table*}[!t]
\centering
\vspace{5pt}
\caption{
    \textnormal{\textbf{Results of Fine-grained Multi-label Evaluation and Different Time Limit Thresholds in \textcolor{red}{LEVEL 1}.}}
}
\setlength\tabcolsep{3pt}  
\renewcommand{\arraystretch}{1.2} 
\resizebox{1.0\linewidth}{!}{
\begin{tabular}{lccccccccccccccc}
\toprule
\multirow{2}{*}{\textbf{Model}} & \multicolumn{3}{c}{\textbf{Interaction Type}} & \multicolumn{3}{c}{\textbf{Subtask Count}} & \multicolumn{5}{c}{\textbf{Spatial Relation}} & \multicolumn{4}{c}{\textbf{Object Attribute}}\\
\cmidrule(lr){2-4} \cmidrule(lr){5-7} \cmidrule(lr){8-12} 
\cmidrule(lr){13-16}
 &Placement &State Change &Composite Actions &Single Action & Two Actions& Three or More Actions & Surface & Container & Left–Right Relation & Ordinal Region & Relative Distance & Color & Size & Pattern & Category\\
\toprule

\rowcolor{gray!10}
\multirow{1}{*}{OpenVLA-OFT~(1.0$\times$)} & \textbf{20.3} & \textbf{0.0} & \textbf{0.0} & \textbf{25.3} & \textbf{3.8} & \textbf{0.0} & \textbf{15.9} & \textbf{19.9} & \textbf{22.1} & \textbf{11.5} & \textbf{0.0} & \textbf{22.8} & \textbf{4.1} & \textbf{18.5} & \textbf{5.7} \\
\multirow{1}{*}{OpenVLA-OFT~(1.1$\times$)} & \textbf{30.3} {\scriptsize (+10.0)} & \textbf{1.6} {\scriptsize (+1.6)} & \textbf{2.0} {\scriptsize (+2.0)} & \textbf{37.7} {\scriptsize (+12.4)} & \textbf{6.5} {\scriptsize (+2.7)} & \textbf{0.0} {\scriptsize (+0.0)} & \textbf{24.9} {\scriptsize (+9.0)} & \textbf{29.2} {\scriptsize (+9.3)} & \textbf{31.9} {\scriptsize (+9.8)} & \textbf{14.7} {\scriptsize (+3.2)} & \textbf{5.0} {\scriptsize (+5.0)} & \textbf{33.8} {\scriptsize (+11.0)} & \textbf{9.6} {\scriptsize (+5.5)} & \textbf{25.4} {\scriptsize (+6.9)} & \textbf{10.2} {\scriptsize (+4.5)} \\
\multirow{1}{*}{OpenVLA-OFT~(1.5$\times$)} & \textbf{39.6} {\scriptsize (+9.3)} & \textbf{6.8} {\scriptsize (+5.2)} & \textbf{7.3} {\scriptsize (+5.3)} & \textbf{49.0} {\scriptsize (+11.3)} & \textbf{10.2} {\scriptsize (+3.7)} & \textbf{0.4} {\scriptsize (+0.4)} & \textbf{32.5} {\scriptsize (+7.6)} & \textbf{38.4} {\scriptsize (+9.2)} & \textbf{40.5} {\scriptsize (+8.6)} & \textbf{23.6} {\scriptsize (+8.9)} & \textbf{12.0} {\scriptsize (+7.0)} & \textbf{42.6} {\scriptsize (+8.8)} & \textbf{12.7} {\scriptsize (+3.1)} & \textbf{35.7} {\scriptsize (+10.3)} & \textbf{20.3} {\scriptsize (+10.1)} \\
\multirow{1}{*}{OpenVLA-OFT~(\(\infty\))} & \textbf{43.0} {\scriptsize (+3.4)} & \textbf{13.0} {\scriptsize (+6.2)} & \textbf{9.0} {\scriptsize (+1.7)} & \textbf{53.8} {\scriptsize (+4.8)} & \textbf{11.7} {\scriptsize (+1.5)} & \textbf{0.4} {\scriptsize (+0.0)} & \textbf{36.8} {\scriptsize (+4.3)} & \textbf{41.7} {\scriptsize (+3.3)} & \textbf{43.6} {\scriptsize (+3.1)} & \textbf{28.2} {\scriptsize (+4.6)} & \textbf{16.0} {\scriptsize (+4.0)} & \textbf{45.9} {\scriptsize (+3.3)} & \textbf{14.7} {\scriptsize (+2.0)} & \textbf{38.5} {\scriptsize (+2.8)} & \textbf{26.9} {\scriptsize (+6.6)} \\
\midrule
\rowcolor{gray!10}
\multirow{1}{*}{$\pi_{0}$~(1.0$\times$)} & \textbf{19.3} & \textbf{9.5} & \textbf{5.0} & \textbf{24.8} & \textbf{4.2} & \textbf{0.0} & \textbf{22.6} & \textbf{15.5} & \textbf{16.2} & \textbf{21.1} & \textbf{2.0} & \textbf{19.7} & \textbf{15.7} & \textbf{15.2} & \textbf{14.0} \\
\multirow{1}{*}{$\pi_{0}$~(1.1$\times$)} & \textbf{29.9} {\scriptsize (+10.6)} & \textbf{15.4} {\scriptsize (+5.9)} & \textbf{10.7} {\scriptsize (+5.7)} & \textbf{37.5} {\scriptsize (+12.7)} & \textbf{9.8} {\scriptsize (+5.6)} & \textbf{0.0} {\scriptsize (+0.0)} & \textbf{34.1} {\scriptsize (+11.5)} & \textbf{24.9} {\scriptsize (+9.4)} & \textbf{25.6} {\scriptsize (+9.4)} & \textbf{30.7} {\scriptsize (+9.6)} & \textbf{7.0} {\scriptsize (+5.0)} & \textbf{30.2} {\scriptsize (+10.5)} & \textbf{26.9} {\scriptsize (+11.2)} & \textbf{22.0} {\scriptsize (+6.8)} & \textbf{23.1} {\scriptsize (+9.1)} \\
\multirow{1}{*}{$\pi_{0}$~(1.5$\times$)} & \textbf{41.0} {\scriptsize (+11.1)} & \textbf{24.9} {\scriptsize (+9.5)} & \textbf{18.0} {\scriptsize (+7.3)} & \textbf{51.2} {\scriptsize (+13.7)} & \textbf{15.4} {\scriptsize (+5.6)} & \textbf{0.0} {\scriptsize (+0.0)} & \textbf{45.1} {\scriptsize (+11.0)} & \textbf{35.7} {\scriptsize (+10.8)} & \textbf{36.0} {\scriptsize (+10.4)} & \textbf{42.5} {\scriptsize (+11.8)} & \textbf{9.0} {\scriptsize (+2.0)} & \textbf{41.2} {\scriptsize (+11.0)} & \textbf{43.1} {\scriptsize (+16.2)} & \textbf{26.5} {\scriptsize (+4.5)} & \textbf{33.6} {\scriptsize (+10.5)} \\
\multirow{1}{*}{$\pi_{0}$~(\(\infty\))} & \textbf{45.4} {\scriptsize (+4.4)} & \textbf{29.6} {\scriptsize (+4.7)} & \textbf{19.0} {\scriptsize (+1.0)} & \textbf{56.9} {\scriptsize (+5.7)} & \textbf{17.4} {\scriptsize (+2.0)} & \textbf{0.0} {\scriptsize (+0.0)} & \textbf{51.1} {\scriptsize (+6.0)} & \textbf{38.9} {\scriptsize (+3.2)} & \textbf{39.3} {\scriptsize (+3.3)} & \textbf{46.4} {\scriptsize (+3.9)} & \textbf{16.0} {\scriptsize (+7.0)} & \textbf{45.2} {\scriptsize (+4.0)} & \textbf{48.2} {\scriptsize (+5.1)} & \textbf{28.5} {\scriptsize (+2.0)} & \textbf{40.6} {\scriptsize (+7.0)} \\
\midrule
\rowcolor{gray!10}
\multirow{1}{*}{X-VLA~(1.0$\times$)} & \textbf{18.3} & \textbf{5.4} & \textbf{1.7} & \textbf{23.1} & \textbf{5.0} & \textbf{0.0} & \textbf{21.1} & \textbf{14.0} & \textbf{15.9} & \textbf{4.7} & \textbf{0.0} & \textbf{19.4} & \textbf{12.2} & \textbf{11.9} & \textbf{14.8} \\
\multirow{1}{*}{X-VLA~(1.1$\times$)} & \textbf{31.0} {\scriptsize (+12.7)} & \textbf{7.7} {\scriptsize (+2.3)} & \textbf{4.7} {\scriptsize (+3.0)} & \textbf{37.6} {\scriptsize (+14.5)} & \textbf{12.6} {\scriptsize (+7.6)} & \textbf{0.2} {\scriptsize (+0.2)} & \textbf{35.3} {\scriptsize (+14.2)} & \textbf{23.8} {\scriptsize (+9.8)} & \textbf{27.3} {\scriptsize (+11.4)} & \textbf{12.9} {\scriptsize (+8.2)} & \textbf{1.0} {\scriptsize (+1.0)} & \textbf{32.3} {\scriptsize (+12.9)} & \textbf{24.3} {\scriptsize (+12.1)} & \textbf{23.7} {\scriptsize (+11.8)} & \textbf{20.4} {\scriptsize (+5.6)} \\
\multirow{1}{*}{X-VLA~(1.5$\times$)} & \textbf{44.7} {\scriptsize (+13.7)} & \textbf{11.8} {\scriptsize (+4.1)} & \textbf{10.0} {\scriptsize (+5.3)} & \textbf{53.7} {\scriptsize (+16.1)} & \textbf{19.7} {\scriptsize (+7.1)} & \textbf{0.9} {\scriptsize (+0.7)} & \textbf{47.9} {\scriptsize (+12.6)} & \textbf{37.3} {\scriptsize (+13.5)} & \textbf{38.4} {\scriptsize (+11.1)} & \textbf{31.3} {\scriptsize (+18.4)} & \textbf{10.0} {\scriptsize (+9.0)} & \textbf{46.3} {\scriptsize (+14.0)} & \textbf{41.2} {\scriptsize (+16.9)} & \textbf{31.9} {\scriptsize (+8.2)} & \textbf{27.8} {\scriptsize (+7.4)} \\
\multirow{1}{*}{X-VLA~(\(\infty\))} & \textbf{48.8} {\scriptsize (+4.1)} & \textbf{14.6} {\scriptsize (+2.8)} & \textbf{12.0} {\scriptsize (+2.0)} & \textbf{58.8} {\scriptsize (+5.1)} & \textbf{21.8} {\scriptsize (+2.1)} & \textbf{0.9} {\scriptsize (+0.0)} & \textbf{51.8} {\scriptsize (+3.9)} & \textbf{41.4} {\scriptsize (+4.1)} & \textbf{42.3} {\scriptsize (+3.9)} & \textbf{33.6} {\scriptsize (+2.3)} & \textbf{18.0} {\scriptsize (+8.0)} & \textbf{50.7} {\scriptsize (+4.4)} & \textbf{43.9} {\scriptsize (+2.7)} & \textbf{34.1} {\scriptsize (+2.2)} & \textbf{31.4} {\scriptsize (+3.6)} \\
\midrule
\rowcolor{gray!10}
\multirow{1}{*}{GR00T1.5~(1.0$\times$)} & \textbf{33.2} & \textbf{22.1} & \textbf{15.0} & \textbf{41.1} & \textbf{14.8} & \textbf{0.4} & \textbf{41.7} & \textbf{25.6} & \textbf{25.8} & \textbf{31.5} & \textbf{20.0} & \textbf{33.4} & \textbf{33.7} & \textbf{18.3} & \textbf{30.6} \\
\multirow{1}{*}{GR00T1.5~(1.1$\times$)} & \textbf{43.6} {\scriptsize (+10.4)} & \textbf{29.3} {\scriptsize (+7.2)} & \textbf{21.7} {\scriptsize (+6.7)} & \textbf{53.4} {\scriptsize (+12.3)} & \textbf{21.0} {\scriptsize (+6.2)} & \textbf{0.9} {\scriptsize (+0.5)} & \textbf{53.5} {\scriptsize (+11.8)} & \textbf{34.5} {\scriptsize (+8.9)} & \textbf{34.6} {\scriptsize (+8.8)} & \textbf{45.5} {\scriptsize (+14.0)} & \textbf{28.0} {\scriptsize (+8.0)} & \textbf{43.3} {\scriptsize (+9.9)} & \textbf{45.9} {\scriptsize (+12.2)} & \textbf{23.3} {\scriptsize (+5.0)} & \textbf{42.1} {\scriptsize (+11.5)} \\
\multirow{1}{*}{GR00T1.5~(1.5$\times$)} & \textbf{52.0} {\scriptsize (+8.4)} & \textbf{42.6} {\scriptsize (+13.3)} & \textbf{30.7} {\scriptsize (+9.0)} & \textbf{64.0} {\scriptsize (+10.6)} & \textbf{26.1} {\scriptsize (+5.1)} & \textbf{1.3} {\scriptsize (+0.4)} & \textbf{63.1} {\scriptsize (+9.6)} & \textbf{42.6} {\scriptsize (+8.1)} & \textbf{41.7} {\scriptsize (+7.1)} & \textbf{56.9} {\scriptsize (+11.4)} & \textbf{34.0} {\scriptsize (+6.0)} & \textbf{51.3} {\scriptsize (+8.0)} & \textbf{56.1} {\scriptsize (+10.2)} & \textbf{28.5} {\scriptsize (+5.2)} & \textbf{54.9} {\scriptsize (+12.8)} \\
\multirow{1}{*}{GR00T1.5~(\(\infty\))} & \textbf{54.9} {\scriptsize (+2.9)} & \textbf{48.2} {\scriptsize (+5.6)} & \textbf{32.3} {\scriptsize (+1.6)} & \textbf{67.9} {\scriptsize (+3.9)} & \textbf{28.0} {\scriptsize (+1.9)} & \textbf{1.3} {\scriptsize (+0.0)} & \textbf{67.6} {\scriptsize (+4.5)} & \textbf{44.7} {\scriptsize (+2.1)} & \textbf{43.7} {\scriptsize (+2.0)} & \textbf{62.7} {\scriptsize (+5.8)} & \textbf{39.0} {\scriptsize (+5.0)} & \textbf{54.0} {\scriptsize (+2.7)} & \textbf{58.6} {\scriptsize (+2.5)} & \textbf{30.9} {\scriptsize (+2.4)} & \textbf{60.8} {\scriptsize (+5.9)} \\
\midrule
\rowcolor{gray!10}
\multirow{1}{*}{$\pi_{0.5}$~(1.0$\times$)} & \textbf{45.4} & \textbf{34.0} & \textbf{29.7} & \textbf{50.2} & \textbf{35.9} & \textbf{18.7} & \textbf{45.4} & \textbf{42.4} & \textbf{45.0} & \textbf{44.0} & \textbf{27.0} & \textbf{46.7} & \textbf{39.6} & \textbf{45.2} & \textbf{34.4} \\
\multirow{1}{*}{$\pi_{0.5}$~(1.1$\times$)} & \textbf{65.8} {\scriptsize (+20.4)} & \textbf{49.8} {\scriptsize (+15.8)} & \textbf{48.3} {\scriptsize (+18.6)} & \textbf{70.9} {\scriptsize (+20.7)} & \textbf{56.7} {\scriptsize (+20.8)} & \textbf{31.1} {\scriptsize (+12.4)} & \textbf{65.1} {\scriptsize (+19.7)} & \textbf{62.7} {\scriptsize (+20.3)} & \textbf{66.4} {\scriptsize (+21.4)} & \textbf{62.5} {\scriptsize (+18.5)} & \textbf{37.0} {\scriptsize (+10.0)} & \textbf{67.7} {\scriptsize (+21.0)} & \textbf{61.0} {\scriptsize (+21.4)} & \textbf{60.4} {\scriptsize (+15.2)} & \textbf{51.2} {\scriptsize (+16.8)} \\
\multirow{1}{*}{$\pi_{0.5}$~(1.5$\times$)} & \textbf{78.2} {\scriptsize (+12.4)} & \textbf{67.7} {\scriptsize (+17.9)} & \textbf{62.0} {\scriptsize (+13.7)} & \textbf{85.9} {\scriptsize (+15.0)} & \textbf{65.4} {\scriptsize (+8.7)} & \textbf{33.0} {\scriptsize (+1.9)} & \textbf{79.8} {\scriptsize (+14.7)} & \textbf{73.9} {\scriptsize (+11.2)} & \textbf{76.4} {\scriptsize (+10.0)} & \textbf{77.5} {\scriptsize (+15.0)} & \textbf{46.0} {\scriptsize (+9.0)} & \textbf{79.9} {\scriptsize (+12.2)} & \textbf{73.7} {\scriptsize (+12.7)} & \textbf{67.6} {\scriptsize (+7.2)} & \textbf{67.8} {\scriptsize (+16.6)} \\
\multirow{1}{*}{$\pi_{0.5}$~(\(\infty\))} & \textbf{81.2} {\scriptsize (+3.0)} & \textbf{73.0} {\scriptsize (+5.3)} & \textbf{63.7} {\scriptsize (+1.7)} & \textbf{89.9} {\scriptsize (+4.0)} & \textbf{67.1} {\scriptsize (+1.7)} & \textbf{33.0} {\scriptsize (+0.0)} & \textbf{83.8} {\scriptsize (+4.0)} & \textbf{76.7} {\scriptsize (+2.8)} & \textbf{78.5} {\scriptsize (+2.1)} & \textbf{82.4} {\scriptsize (+4.9)} & \textbf{51.0} {\scriptsize (+5.0)} & \textbf{82.7} {\scriptsize (+2.8)} & \textbf{75.5} {\scriptsize (+1.8)} & \textbf{68.5} {\scriptsize (+0.9)} & \textbf{74.8} {\scriptsize (+7.0)} \\

\bottomrule
\end{tabular}}

\vspace{-0.5cm}
\label{tab:app_multi_label_level1}
\end{table*}

\begin{table*}[!t]
\centering
\vspace{5pt}
\caption{
    \textnormal{\textbf{Results of Fine-grained Multi-label Evaluation and Different Time Limit Thresholds in \textcolor{red}{LEVEL 2}.}}
}
\setlength\tabcolsep{3pt}  
\renewcommand{\arraystretch}{1.2} 
\resizebox{1.0\linewidth}{!}{
\begin{tabular}{lccccccccccccccc}
\toprule
\multirow{2}{*}{\textbf{Model}}  & \multicolumn{3}{c}{\textbf{Interaction Type}} & \multicolumn{3}{c}{\textbf{Subtask Count}} & \multicolumn{5}{c}{\textbf{Spatial Relation}} & \multicolumn{4}{c}{\textbf{Object Attribute}}\\
\cmidrule(lr){2-4} \cmidrule(lr){5-7} \cmidrule(lr){8-12} \cmidrule(lr){13-16}
 &Placement &State Change &Composite Actions &Single Action & Two Actions& Three or More Actions & Surface & Container & Left–Right Relation & Ordinal Region & Relative Distance & Color & Size & Pattern & Category\\
\toprule

\rowcolor{gray!10}
\multirow{1}{*}{OpenVLA-OFT~(1.0$\times$)} & \textbf{12.5} & \textbf{0.5} & \textbf{0.0} & \textbf{15.7} & \textbf{2.1} & \textbf{0.2} & \textbf{10.8} & \textbf{11.7} & \textbf{12.5} & \textbf{7.6} & \textbf{1.0} & \textbf{14.3} & \textbf{1.4} & \textbf{9.6} & \textbf{3.8} \\
\multirow{1}{*}{OpenVLA-OFT~(1.1$\times$)} & \textbf{18.3} {\scriptsize (+5.8)} & \textbf{1.8} {\scriptsize (+1.3)} & \textbf{1.0} {\scriptsize (+1.0)} & \textbf{22.8} {\scriptsize (+7.1)} & \textbf{4.2} {\scriptsize (+2.1)} & \textbf{0.2} {\scriptsize (+0.0)} & \textbf{15.1} {\scriptsize (+4.3)} & \textbf{17.6} {\scriptsize (+5.9)} & \textbf{19.2} {\scriptsize (+6.7)} & \textbf{10.4} {\scriptsize (+2.8)} & \textbf{3.0} {\scriptsize (+2.0)} & \textbf{20.7} {\scriptsize (+6.4)} & \textbf{4.3} {\scriptsize (+2.9)} & \textbf{14.1} {\scriptsize (+4.5)} & \textbf{6.0} {\scriptsize (+2.2)} \\
\multirow{1}{*}{OpenVLA-OFT~(1.5$\times$)} & \textbf{26.1} {\scriptsize (+7.8)} & \textbf{5.4} {\scriptsize (+3.6)} & \textbf{3.7} {\scriptsize (+2.7)} & \textbf{32.6} {\scriptsize (+9.8)} & \textbf{6.7} {\scriptsize (+2.5)} & \textbf{0.4} {\scriptsize (+0.2)} & \textbf{22.1} {\scriptsize (+7.0)} & \textbf{24.9} {\scriptsize (+7.3)} & \textbf{26.3} {\scriptsize (+7.1)} & \textbf{15.1} {\scriptsize (+4.7)} & \textbf{8.0} {\scriptsize (+5.0)} & \textbf{29.5} {\scriptsize (+8.8)} & \textbf{5.9} {\scriptsize (+1.6)} & \textbf{20.4} {\scriptsize (+6.3)} & \textbf{10.6} {\scriptsize (+4.6)} \\
\multirow{1}{*}{OpenVLA-OFT~(\(\infty\))} & \textbf{29.2} {\scriptsize (+3.1)} & \textbf{8.6} {\scriptsize (+3.2)} & \textbf{4.3} {\scriptsize (+0.6)} & \textbf{36.7} {\scriptsize (+4.1)} & \textbf{7.5} {\scriptsize (+0.8)} & \textbf{0.4} {\scriptsize (+0.0)} & \textbf{24.9} {\scriptsize (+2.8)} & \textbf{28.1} {\scriptsize (+3.2)} & \textbf{29.4} {\scriptsize (+3.1)} & \textbf{20.0} {\scriptsize (+4.9)} & \textbf{10.0} {\scriptsize (+2.0)} & \textbf{32.4} {\scriptsize (+2.9)} & \textbf{9.6} {\scriptsize (+3.7)} & \textbf{22.4} {\scriptsize (+2.0)} & \textbf{14.3} {\scriptsize (+3.7)} \\
\midrule
\rowcolor{gray!10}
\multirow{1}{*}{$\pi_{0}$~(1.0$\times$)} & \textbf{13.9} & \textbf{6.1} & \textbf{1.3} & \textbf{18.3} & \textbf{2.0} & \textbf{0.0} & \textbf{15.8} & \textbf{11.2} & \textbf{12.2} & \textbf{13.1} & \textbf{2.0} & \textbf{15.3} & \textbf{8.6} & \textbf{10.7} & \textbf{7.2} \\
\multirow{1}{*}{$\pi_{0}$~(1.1$\times$)} & \textbf{22.2} {\scriptsize (+8.3)} & \textbf{9.6} {\scriptsize (+3.5)} & \textbf{5.0} {\scriptsize (+3.7)} & \textbf{28.5} {\scriptsize (+10.2)} & \textbf{4.7} {\scriptsize (+2.7)} & \textbf{0.0} {\scriptsize (+0.0)} & \textbf{24.3} {\scriptsize (+8.5)} & \textbf{18.2} {\scriptsize (+7.0)} & \textbf{19.3} {\scriptsize (+7.1)} & \textbf{20.4} {\scriptsize (+7.3)} & \textbf{3.0} {\scriptsize (+1.0)} & \textbf{24.3} {\scriptsize (+9.0)} & \textbf{15.9} {\scriptsize (+7.3)} & \textbf{15.0} {\scriptsize (+4.3)} & \textbf{11.0} {\scriptsize (+3.8)} \\
\multirow{1}{*}{$\pi_{0}$~(1.5$\times$)} & \textbf{32.3} {\scriptsize (+10.1)} & \textbf{16.1} {\scriptsize (+6.5)} & \textbf{9.0} {\scriptsize (+4.0)} & \textbf{41.0} {\scriptsize (+12.5)} & \textbf{9.3} {\scriptsize (+4.6)} & \textbf{0.0} {\scriptsize (+0.0)} & \textbf{33.6} {\scriptsize (+9.3)} & \textbf{28.1} {\scriptsize (+9.9)} & \textbf{28.5} {\scriptsize (+9.2)} & \textbf{32.4} {\scriptsize (+12.0)} & \textbf{4.0} {\scriptsize (+1.0)} & \textbf{35.0} {\scriptsize (+10.7)} & \textbf{27.8} {\scriptsize (+11.9)} & \textbf{20.2} {\scriptsize (+5.2)} & \textbf{17.3} {\scriptsize (+6.3)} \\
\multirow{1}{*}{$\pi_{0}$~(\(\infty\))} & \textbf{35.7} {\scriptsize (+3.4)} & \textbf{21.1} {\scriptsize (+5.0)} & \textbf{10.3} {\scriptsize (+1.3)} & \textbf{45.6} {\scriptsize (+4.6)} & \textbf{10.6} {\scriptsize (+1.3)} & \textbf{0.0} {\scriptsize (+0.0)} & \textbf{38.1} {\scriptsize (+4.5)} & \textbf{31.0} {\scriptsize (+2.9)} & \textbf{31.5} {\scriptsize (+3.0)} & \textbf{35.8} {\scriptsize (+3.4)} & \textbf{5.0} {\scriptsize (+1.0)} & \textbf{38.6} {\scriptsize (+3.6)} & \textbf{32.9} {\scriptsize (+5.1)} & \textbf{21.3} {\scriptsize (+1.1)} & \textbf{21.2} {\scriptsize (+3.9)} \\
\midrule
\rowcolor{gray!10}
\multirow{1}{*}{X-VLA~(1.0$\times$)} & \textbf{14.1} & \textbf{4.7} & \textbf{2.7} & \textbf{17.8} & \textbf{3.6} & \textbf{0.0} & \textbf{15.5} & \textbf{11.2} & \textbf{12.9} & \textbf{2.4} & \textbf{0.0} & \textbf{15.0} & \textbf{7.8} & \textbf{9.3} & \textbf{10.2} \\
\multirow{1}{*}{X-VLA~(1.1$\times$)} & \textbf{23.3} {\scriptsize (+9.2)} & \textbf{6.3} {\scriptsize (+1.6)} & \textbf{4.7} {\scriptsize (+2.0)} & \textbf{28.9} {\scriptsize (+11.1)} & \textbf{7.2} {\scriptsize (+3.6)} & \textbf{0.0} {\scriptsize (+0.0)} & \textbf{25.3} {\scriptsize (+9.8)} & \textbf{18.5} {\scriptsize (+7.3)} & \textbf{20.8} {\scriptsize (+7.9)} & \textbf{8.5} {\scriptsize (+6.1)} & \textbf{2.0} {\scriptsize (+2.0)} & \textbf{25.0} {\scriptsize (+10.0)} & \textbf{16.9} {\scriptsize (+9.1)} & \textbf{16.1} {\scriptsize (+6.8)} & \textbf{13.1} {\scriptsize (+2.9)} \\
\multirow{1}{*}{X-VLA~(1.5$\times$)} & \textbf{36.2} {\scriptsize (+12.9)} & \textbf{8.1} {\scriptsize (+1.8)} & \textbf{6.7} {\scriptsize (+2.0)} & \textbf{44.5} {\scriptsize (+15.6)} & \textbf{11.8} {\scriptsize (+4.6)} & \textbf{0.4} {\scriptsize (+0.4)} & \textbf{37.5} {\scriptsize (+12.2)} & \textbf{30.4} {\scriptsize (+11.9)} & \textbf{31.8} {\scriptsize (+11.0)} & \textbf{22.5} {\scriptsize (+14.0)} & \textbf{5.0} {\scriptsize (+3.0)} & \textbf{38.3} {\scriptsize (+13.3)} & \textbf{29.6} {\scriptsize (+12.7)} & \textbf{25.6} {\scriptsize (+9.5)} & \textbf{19.8} {\scriptsize (+6.7)} \\
\multirow{1}{*}{X-VLA~(\(\infty\))} & \textbf{40.6} {\scriptsize (+4.4)} & \textbf{11.9} {\scriptsize (+3.8)} & \textbf{8.7} {\scriptsize (+2.0)} & \textbf{50.1} {\scriptsize (+5.6)} & \textbf{13.2} {\scriptsize (+1.4)} & \textbf{0.4} {\scriptsize (+0.0)} & \textbf{41.8} {\scriptsize (+4.3)} & \textbf{34.6} {\scriptsize (+4.2)} & \textbf{35.7} {\scriptsize (+3.9)} & \textbf{27.3} {\scriptsize (+4.8)} & \textbf{8.0} {\scriptsize (+3.0)} & \textbf{42.8} {\scriptsize (+4.5)} & \textbf{33.3} {\scriptsize (+3.7)} & \textbf{27.8} {\scriptsize (+2.2)} & \textbf{24.1} {\scriptsize (+4.3)} \\
\midrule
\rowcolor{gray!10}
\multirow{1}{*}{GR00T1.5~(1.0$\times$)} & \textbf{25.1} & \textbf{11.2} & \textbf{6.3} & \textbf{31.4} & \textbf{8.7} & \textbf{0.0} & \textbf{31.3} & \textbf{18.2} & \textbf{18.8} & \textbf{21.6} & \textbf{11.0} & \textbf{25.8} & \textbf{24.3} & \textbf{15.0} & \textbf{18.4} \\
\multirow{1}{*}{GR00T1.5~(1.1$\times$)} & \textbf{33.4} {\scriptsize (+8.3)} & \textbf{14.7} {\scriptsize (+3.5)} & \textbf{8.7} {\scriptsize (+2.4)} & \textbf{41.5} {\scriptsize (+10.1)} & \textbf{12.5} {\scriptsize (+3.8)} & \textbf{0.0} {\scriptsize (+0.0)} & \textbf{40.6} {\scriptsize (+9.3)} & \textbf{25.1} {\scriptsize (+6.9)} & \textbf{25.7} {\scriptsize (+6.9)} & \textbf{28.0} {\scriptsize (+6.4)} & \textbf{18.0} {\scriptsize (+7.0)} & \textbf{34.4} {\scriptsize (+8.6)} & \textbf{33.1} {\scriptsize (+8.8)} & \textbf{19.4} {\scriptsize (+4.4)} & \textbf{24.0} {\scriptsize (+5.6)} \\
\multirow{1}{*}{GR00T1.5~(1.5$\times$)} & \textbf{41.4} {\scriptsize (+8.0)} & \textbf{25.6} {\scriptsize (+10.9)} & \textbf{13.3} {\scriptsize (+4.6)} & \textbf{51.9} {\scriptsize (+10.4)} & \textbf{16.5} {\scriptsize (+4.0)} & \textbf{0.0} {\scriptsize (+0.0)} & \textbf{49.0} {\scriptsize (+8.4)} & \textbf{33.1} {\scriptsize (+8.0)} & \textbf{32.9} {\scriptsize (+7.2)} & \textbf{40.4} {\scriptsize (+12.4)} & \textbf{21.0} {\scriptsize (+3.0)} & \textbf{42.4} {\scriptsize (+8.0)} & \textbf{42.0} {\scriptsize (+8.9)} & \textbf{24.1} {\scriptsize (+4.7)} & \textbf{34.7} {\scriptsize (+10.7)} \\
\multirow{1}{*}{GR00T1.5~(\(\infty\))} & \textbf{44.4} {\scriptsize (+3.0)} & \textbf{30.5} {\scriptsize (+4.9)} & \textbf{14.3} {\scriptsize (+1.0)} & \textbf{56.0} {\scriptsize (+4.1)} & \textbf{17.9} {\scriptsize (+1.4)} & \textbf{0.0} {\scriptsize (+0.0)} & \textbf{53.2} {\scriptsize (+4.2)} & \textbf{35.6} {\scriptsize (+2.5)} & \textbf{35.3} {\scriptsize (+2.4)} & \textbf{44.9} {\scriptsize (+4.5)} & \textbf{26.0} {\scriptsize (+5.0)} & \textbf{45.2} {\scriptsize (+2.8)} & \textbf{45.5} {\scriptsize (+3.5)} & \textbf{26.1} {\scriptsize (+2.0)} & \textbf{40.0} {\scriptsize (+5.3)} \\
\midrule
\rowcolor{gray!10}
\multirow{1}{*}{$\pi_{0.5}$~(1.0$\times$)} & \textbf{37.9} & \textbf{26.1} & \textbf{20.3} & \textbf{42.9} & \textbf{26.2} & \textbf{15.4} & \textbf{37.6} & \textbf{35.4} & \textbf{38.0} & \textbf{30.2} & \textbf{15.0} & \textbf{39.1} & \textbf{35.1} & \textbf{36.7} & \textbf{27.7} \\
\multirow{1}{*}{$\pi_{0.5}$~(1.1$\times$)} & \textbf{53.5} {\scriptsize (+15.6)} & \textbf{36.8} {\scriptsize (+10.7)} & \textbf{28.0} {\scriptsize (+7.7)} & \textbf{59.8} {\scriptsize (+16.9)} & \textbf{40.1} {\scriptsize (+13.9)} & \textbf{21.3} {\scriptsize (+5.9)} & \textbf{53.1} {\scriptsize (+15.5)} & \textbf{49.8} {\scriptsize (+14.4)} & \textbf{53.6} {\scriptsize (+15.6)} & \textbf{44.0} {\scriptsize (+13.8)} & \textbf{21.0} {\scriptsize (+6.0)} & \textbf{56.4} {\scriptsize (+17.3)} & \textbf{47.5} {\scriptsize (+12.4)} & \textbf{49.4} {\scriptsize (+12.7)} & \textbf{36.3} {\scriptsize (+8.6)} \\
\multirow{1}{*}{$\pi_{0.5}$~(1.5$\times$)} & \textbf{66.8} {\scriptsize (+13.3)} & \textbf{49.8} {\scriptsize (+13.0)} & \textbf{37.7} {\scriptsize (+9.7)} & \textbf{75.6} {\scriptsize (+15.8)} & \textbf{49.1} {\scriptsize (+9.0)} & \textbf{23.0} {\scriptsize (+1.7)} & \textbf{66.8} {\scriptsize (+13.7)} & \textbf{62.7} {\scriptsize (+12.9)} & \textbf{65.7} {\scriptsize (+12.1)} & \textbf{60.7} {\scriptsize (+16.7)} & \textbf{28.0} {\scriptsize (+7.0)} & \textbf{70.5} {\scriptsize (+14.1)} & \textbf{59.6} {\scriptsize (+12.1)} & \textbf{57.8} {\scriptsize (+8.4)} & \textbf{47.8} {\scriptsize (+11.5)} \\
\multirow{1}{*}{$\pi_{0.5}$~(\(\infty\))} & \textbf{69.8} {\scriptsize (+3.0)} & \textbf{55.1} {\scriptsize (+5.3)} & \textbf{39.0} {\scriptsize (+1.3)} & \textbf{79.7} {\scriptsize (+4.1)} & \textbf{50.7} {\scriptsize (+1.6)} & \textbf{23.0} {\scriptsize (+0.0)} & \textbf{70.8} {\scriptsize (+4.0)} & \textbf{65.5} {\scriptsize (+2.8)} & \textbf{68.2} {\scriptsize (+2.5)} & \textbf{64.7} {\scriptsize (+4.0)} & \textbf{34.0} {\scriptsize (+6.0)} & \textbf{73.1} {\scriptsize (+2.6)} & \textbf{62.5} {\scriptsize (+2.9)} & \textbf{60.0} {\scriptsize (+2.2)} & \textbf{54.2} {\scriptsize (+6.4)} \\

\bottomrule
\end{tabular}}

\vspace{-0.5cm}
\label{tab:app_multi_label_level2}
\end{table*}

\begin{table*}[!t]
\centering
\vspace{5pt}
\caption{
    \textnormal{\textbf{Results of Fine-grained Multi-label Evaluation and Different Time Limit Thresholds in \textcolor{red}{LEVEL 3}.}}
}
\setlength\tabcolsep{3pt}  
\renewcommand{\arraystretch}{1.2} 
\resizebox{1.0\linewidth}{!}{
\begin{tabular}{lccccccccccccccc}
\toprule
\multirow{2}{*}{\textbf{Model}}  & \multicolumn{3}{c}{\textbf{Interaction Type}} & \multicolumn{3}{c}{\textbf{Subtask Count}} & \multicolumn{5}{c}{\textbf{Spatial Relation}} & \multicolumn{4}{c}{\textbf{Object Attribute}}\\
\cmidrule(lr){2-4} \cmidrule(lr){5-7} \cmidrule(lr){8-12} \cmidrule(lr){13-16}
 &Placement &State Change &Composite Actions &Single Action & Two Actions& Three or More Actions & Surface & Container & Left–Right Relation & Ordinal Region & Relative Distance & Color & Size & Pattern & Category\\
\toprule

\rowcolor{gray!10}
\multirow{1}{*}{OpenVLA-OFT~(1.0$\times$)} & \textbf{6.9} & \textbf{0.0} & \textbf{0.0} & \textbf{8.8} & \textbf{0.8} & \textbf{0.0} & \textbf{4.4} & \textbf{7.6} & \textbf{7.6} & \textbf{8.4} & \textbf{1.0} & \textbf{7.8} & \textbf{2.2} & \textbf{6.3} & \textbf{1.4} \\
\multirow{1}{*}{OpenVLA-OFT~(1.1$\times$)} & \textbf{9.2} {\scriptsize (+2.3)} & \textbf{0.0} {\scriptsize (+0.0)} & \textbf{0.0} {\scriptsize (+0.0)} & \textbf{11.7} {\scriptsize (+2.9)} & \textbf{0.9} {\scriptsize (+0.1)} & \textbf{0.0} {\scriptsize (+0.0)} & \textbf{6.0} {\scriptsize (+1.6)} & \textbf{9.8} {\scriptsize (+2.2)} & \textbf{9.8} {\scriptsize (+2.2)} & \textbf{9.5} {\scriptsize (+1.1)} & \textbf{1.0} {\scriptsize (+0.0)} & \textbf{10.3} {\scriptsize (+2.5)} & \textbf{2.4} {\scriptsize (+0.2)} & \textbf{8.1} {\scriptsize (+1.8)} & \textbf{2.6} {\scriptsize (+1.2)} \\
\multirow{1}{*}{OpenVLA-OFT~(1.5$\times$)} & \textbf{13.5} {\scriptsize (+4.3)} & \textbf{0.7} {\scriptsize (+0.7)} & \textbf{0.0} {\scriptsize (+0.0)} & \textbf{17.3} {\scriptsize (+5.6)} & \textbf{1.4} {\scriptsize (+0.5)} & \textbf{0.0} {\scriptsize (+0.0)} & \textbf{9.6} {\scriptsize (+3.6)} & \textbf{13.9} {\scriptsize (+4.1)} & \textbf{13.2} {\scriptsize (+3.4)} & \textbf{14.7} {\scriptsize (+5.2)} & \textbf{3.0} {\scriptsize (+2.0)} & \textbf{15.0} {\scriptsize (+4.7)} & \textbf{2.5} {\scriptsize (+0.1)} & \textbf{10.9} {\scriptsize (+2.8)} & \textbf{5.6} {\scriptsize (+3.0)} \\
\multirow{1}{*}{OpenVLA-OFT~(\(\infty\))} & \textbf{16.0} {\scriptsize (+2.5)} & \textbf{1.2} {\scriptsize (+0.5)} & \textbf{0.0} {\scriptsize (+0.0)} & \textbf{20.5} {\scriptsize (+3.2)} & \textbf{1.6} {\scriptsize (+0.2)} & \textbf{0.0} {\scriptsize (+0.0)} & \textbf{11.9} {\scriptsize (+2.3)} & \textbf{16.2} {\scriptsize (+2.3)} & \textbf{15.1} {\scriptsize (+1.9)} & \textbf{18.0} {\scriptsize (+3.3)} & \textbf{4.0} {\scriptsize (+1.0)} & \textbf{17.9} {\scriptsize (+2.9)} & \textbf{4.1} {\scriptsize (+1.6)} & \textbf{11.1} {\scriptsize (+0.2)} & \textbf{7.0} {\scriptsize (+1.4)} \\
\midrule
\rowcolor{gray!10}
\multirow{1}{*}{$\pi_{0}$~(1.0$\times$)} & \textbf{7.6} & \textbf{4.6} & \textbf{1.0} & \textbf{10.2} & \textbf{0.5} & \textbf{0.0} & \textbf{7.8} & \textbf{6.9} & \textbf{6.5} & \textbf{12.5} & \textbf{0.0} & \textbf{7.9} & \textbf{1.8} & \textbf{8.0} & \textbf{6.4} \\
\multirow{1}{*}{$\pi_{0}$~(1.1$\times$)} & \textbf{10.9} {\scriptsize (+3.3)} & \textbf{6.3} {\scriptsize (+1.7)} & \textbf{1.3} {\scriptsize (+0.3)} & \textbf{14.7} {\scriptsize (+4.5)} & \textbf{0.9} {\scriptsize (+0.4)} & \textbf{0.0} {\scriptsize (+0.0)} & \textbf{11.8} {\scriptsize (+4.0)} & \textbf{9.5} {\scriptsize (+2.6)} & \textbf{9.1} {\scriptsize (+2.6)} & \textbf{15.8} {\scriptsize (+3.3)} & \textbf{0.0} {\scriptsize (+0.0)} & \textbf{11.7} {\scriptsize (+3.8)} & \textbf{2.2} {\scriptsize (+0.4)} & \textbf{9.4} {\scriptsize (+1.4)} & \textbf{9.4} {\scriptsize (+3.0)} \\
\multirow{1}{*}{$\pi_{0}$~(1.5$\times$)} & \textbf{16.9} {\scriptsize (+6.0)} & \textbf{9.5} {\scriptsize (+3.2)} & \textbf{3.7} {\scriptsize (+2.4)} & \textbf{22.1} {\scriptsize (+7.4)} & \textbf{3.2} {\scriptsize (+2.3)} & \textbf{0.0} {\scriptsize (+0.0)} & \textbf{17.5} {\scriptsize (+5.7)} & \textbf{15.0} {\scriptsize (+5.5)} & \textbf{13.9} {\scriptsize (+4.8)} & \textbf{23.8} {\scriptsize (+8.0)} & \textbf{0.0} {\scriptsize (+0.0)} & \textbf{18.3} {\scriptsize (+6.6)} & \textbf{3.5} {\scriptsize (+1.3)} & \textbf{12.4} {\scriptsize (+3.0)} & \textbf{13.8} {\scriptsize (+4.4)} \\
\multirow{1}{*}{$\pi_{0}$~(\(\infty\))} & \textbf{19.9} {\scriptsize (+3.0)} & \textbf{12.6} {\scriptsize (+3.1)} & \textbf{3.7} {\scriptsize (+0.0)} & \textbf{26.2} {\scriptsize (+4.1)} & \textbf{3.9} {\scriptsize (+0.7)} & \textbf{0.0} {\scriptsize (+0.0)} & \textbf{20.1} {\scriptsize (+2.6)} & \textbf{18.1} {\scriptsize (+3.1)} & \textbf{17.4} {\scriptsize (+3.5)} & \textbf{27.3} {\scriptsize (+3.5)} & \textbf{0.0} {\scriptsize (+0.0)} & \textbf{21.8} {\scriptsize (+3.5)} & \textbf{4.3} {\scriptsize (+0.8)} & \textbf{13.7} {\scriptsize (+1.3)} & \textbf{16.2} {\scriptsize (+2.4)} \\
\midrule
\rowcolor{gray!10}
\multirow{1}{*}{X-VLA~(1.0$\times$)} & \textbf{7.3} & \textbf{1.8} & \textbf{0.0} & \textbf{9.4} & \textbf{1.2} & \textbf{0.0} & \textbf{6.7} & \textbf{6.6} & \textbf{6.8} & \textbf{4.4} & \textbf{0.0} & \textbf{8.1} & \textbf{3.9} & \textbf{6.9} & \textbf{3.6} \\
\multirow{1}{*}{X-VLA~(1.1$\times$)} & \textbf{10.5} {\scriptsize (+3.2)} & \textbf{3.2} {\scriptsize (+1.4)} & \textbf{1.0} {\scriptsize (+1.0)} & \textbf{13.5} {\scriptsize (+4.1)} & \textbf{1.9} {\scriptsize (+0.7)} & \textbf{0.0} {\scriptsize (+0.0)} & \textbf{9.4} {\scriptsize (+2.7)} & \textbf{9.8} {\scriptsize (+3.2)} & \textbf{9.8} {\scriptsize (+3.0)} & \textbf{8.2} {\scriptsize (+3.8)} & \textbf{0.0} {\scriptsize (+0.0)} & \textbf{11.6} {\scriptsize (+3.5)} & \textbf{5.7} {\scriptsize (+1.8)} & \textbf{8.9} {\scriptsize (+2.0)} & \textbf{5.2} {\scriptsize (+1.6)} \\
\multirow{1}{*}{X-VLA~(1.5$\times$)} & \textbf{18.0} {\scriptsize (+7.5)} & \textbf{6.3} {\scriptsize (+3.1)} & \textbf{4.3} {\scriptsize (+3.3)} & \textbf{22.9} {\scriptsize (+9.4)} & \textbf{3.8} {\scriptsize (+1.9)} & \textbf{0.0} {\scriptsize (+0.0)} & \textbf{16.7} {\scriptsize (+7.3)} & \textbf{16.7} {\scriptsize (+6.9)} & \textbf{15.8} {\scriptsize (+6.0)} & \textbf{19.6} {\scriptsize (+11.4)} & \textbf{1.0} {\scriptsize (+1.0)} & \textbf{19.7} {\scriptsize (+8.1)} & \textbf{9.2} {\scriptsize (+3.5)} & \textbf{14.4} {\scriptsize (+5.5)} & \textbf{9.6} {\scriptsize (+4.4)} \\
\multirow{1}{*}{X-VLA~(\(\infty\))} & \textbf{22.1} {\scriptsize (+4.1)} & \textbf{9.3} {\scriptsize (+3.0)} & \textbf{6.3} {\scriptsize (+2.0)} & \textbf{28.2} {\scriptsize (+5.3)} & \textbf{4.8} {\scriptsize (+1.0)} & \textbf{0.0} {\scriptsize (+0.0)} & \textbf{20.8} {\scriptsize (+4.1)} & \textbf{20.6} {\scriptsize (+3.9)} & \textbf{19.3} {\scriptsize (+3.5)} & \textbf{26.5} {\scriptsize (+6.9)} & \textbf{5.0} {\scriptsize (+4.0)} & \textbf{24.1} {\scriptsize (+4.4)} & \textbf{12.9} {\scriptsize (+3.7)} & \textbf{15.9} {\scriptsize (+1.5)} & \textbf{13.1} {\scriptsize (+3.5)} \\
\midrule
\rowcolor{gray!10}
\multirow{1}{*}{GR00T1.5~(1.0$\times$)} & \textbf{13.4} & \textbf{8.9} & \textbf{3.3} & \textbf{17.7} & \textbf{2.6} & \textbf{0.0} & \textbf{16.1} & \textbf{10.2} & \textbf{10.7} & \textbf{19.6} & \textbf{9.0} & \textbf{14.2} & \textbf{6.7} & \textbf{8.7} & \textbf{12.7} \\
\multirow{1}{*}{GR00T1.5~(1.1$\times$)} & \textbf{18.7} {\scriptsize (+5.3)} & \textbf{11.4} {\scriptsize (+2.5)} & \textbf{4.3} {\scriptsize (+1.0)} & \textbf{24.2} {\scriptsize (+6.5)} & \textbf{4.6} {\scriptsize (+2.0)} & \textbf{0.0} {\scriptsize (+0.0)} & \textbf{22.5} {\scriptsize (+6.4)} & \textbf{13.9} {\scriptsize (+3.7)} & \textbf{14.4} {\scriptsize (+3.7)} & \textbf{25.6} {\scriptsize (+6.0)} & \textbf{11.0} {\scriptsize (+2.0)} & \textbf{19.5} {\scriptsize (+5.3)} & \textbf{10.4} {\scriptsize (+3.7)} & \textbf{12.0} {\scriptsize (+3.3)} & \textbf{17.6} {\scriptsize (+4.9)} \\
\multirow{1}{*}{GR00T1.5~(1.5$\times$)} & \textbf{25.4} {\scriptsize (+6.7)} & \textbf{20.4} {\scriptsize (+9.0)} & \textbf{7.3} {\scriptsize (+3.0)} & \textbf{33.3} {\scriptsize (+9.1)} & \textbf{6.6} {\scriptsize (+2.0)} & \textbf{0.0} {\scriptsize (+0.0)} & \textbf{31.3} {\scriptsize (+8.8)} & \textbf{19.5} {\scriptsize (+5.6)} & \textbf{18.6} {\scriptsize (+4.2)} & \textbf{36.4} {\scriptsize (+10.8)} & \textbf{15.0} {\scriptsize (+4.0)} & \textbf{26.6} {\scriptsize (+7.1)} & \textbf{15.9} {\scriptsize (+5.5)} & \textbf{13.3} {\scriptsize (+1.3)} & \textbf{27.0} {\scriptsize (+9.4)} \\
\multirow{1}{*}{GR00T1.5~(\(\infty\))} & \textbf{29.2} {\scriptsize (+3.8)} & \textbf{27.0} {\scriptsize (+6.6)} & \textbf{9.0} {\scriptsize (+1.7)} & \textbf{38.7} {\scriptsize (+5.4)} & \textbf{8.3} {\scriptsize (+1.7)} & \textbf{0.0} {\scriptsize (+0.0)} & \textbf{37.1} {\scriptsize (+5.8)} & \textbf{22.4} {\scriptsize (+2.9)} & \textbf{20.8} {\scriptsize (+2.2)} & \textbf{43.5} {\scriptsize (+7.1)} & \textbf{22.0} {\scriptsize (+7.0)} & \textbf{30.3} {\scriptsize (+3.7)} & \textbf{19.6} {\scriptsize (+3.7)} & \textbf{15.4} {\scriptsize (+2.1)} & \textbf{34.7} {\scriptsize (+7.7)} \\
\midrule
\rowcolor{gray!10}
\multirow{1}{*}{$\pi_{0.5}$~(1.0$\times$)} & \textbf{26.0} & \textbf{21.1} & \textbf{14.5} & \textbf{30.9} & \textbf{15.3} & \textbf{7.2} & \textbf{26.0} & \textbf{24.1} & \textbf{24.3} & \textbf{28.4} & \textbf{0.0} & \textbf{28.3} & \textbf{10.0} & \textbf{25.4} & \textbf{21.2} \\
\multirow{1}{*}{$\pi_{0.5}$~(1.1$\times$)} & \textbf{35.5} {\scriptsize (+9.5)} & \textbf{35.1} {\scriptsize (+14.0)} & \textbf{24.6} {\scriptsize (+10.1)} & \textbf{42.4} {\scriptsize (+11.5)} & \textbf{22.0} {\scriptsize (+6.7)} & \textbf{9.1} {\scriptsize (+1.9)} & \textbf{37.5} {\scriptsize (+11.5)} & \textbf{32.2} {\scriptsize (+8.1)} & \textbf{31.7} {\scriptsize (+7.4)} & \textbf{42.7} {\scriptsize (+14.3)} & \textbf{0.0} {\scriptsize (+0.0)} & \textbf{38.5} {\scriptsize (+10.2)} & \textbf{17.8} {\scriptsize (+7.8)} & \textbf{32.2} {\scriptsize (+6.8)} & \textbf{30.3} {\scriptsize (+9.1)} \\
\multirow{1}{*}{$\pi_{0.5}$~(1.5$\times$)} & \textbf{48.1} {\scriptsize (+12.6)} & \textbf{48.8} {\scriptsize (+13.7)} & \textbf{34.7} {\scriptsize (+10.1)} & \textbf{57.7} {\scriptsize (+15.3)} & \textbf{29.6} {\scriptsize (+7.6)} & \textbf{10.9} {\scriptsize (+1.8)} & \textbf{51.0} {\scriptsize (+13.5)} & \textbf{44.1} {\scriptsize (+11.9)} & \textbf{42.2} {\scriptsize (+10.5)} & \textbf{56.4} {\scriptsize (+13.7)} & \textbf{1.0} {\scriptsize (+1.0)} & \textbf{52.6} {\scriptsize (+14.1)} & \textbf{28.2} {\scriptsize (+10.4)} & \textbf{35.9} {\scriptsize (+3.7)} & \textbf{41.3} {\scriptsize (+11.0)} \\
\multirow{1}{*}{$\pi_{0.5}$~(\(\infty\))} & \textbf{52.6} {\scriptsize (+4.5)} & \textbf{54.5} {\scriptsize (+5.7)} & \textbf{36.2} {\scriptsize (+1.5)} & \textbf{63.7} {\scriptsize (+6.0)} & \textbf{31.6} {\scriptsize (+2.0)} & \textbf{10.9} {\scriptsize (+0.0)} & \textbf{56.1} {\scriptsize (+5.1)} & \textbf{48.5} {\scriptsize (+4.4)} & \textbf{46.4} {\scriptsize (+4.2)} & \textbf{63.5} {\scriptsize (+7.1)} & \textbf{1.0} {\scriptsize (+0.0)} & \textbf{56.9} {\scriptsize (+4.3)} & \textbf{34.1} {\scriptsize (+5.9)} & \textbf{39.3} {\scriptsize (+3.4)} & \textbf{47.5} {\scriptsize (+6.2)} \\

\bottomrule
\end{tabular}}

\vspace{-0.5cm}
\label{tab:app_multi_label_level3}
\end{table*}

\begin{table*}[!t]
\centering
\vspace{5pt}
\caption{
    \textnormal{\textbf{Results of Fine-grained Multi-label Evaluation and Different Time Limit Thresholds in \textcolor{red}{LEVEL 4}.}}
}
\setlength\tabcolsep{3pt}  
\renewcommand{\arraystretch}{1.2} 
\resizebox{1.0\linewidth}{!}{
\begin{tabular}{lccccccccccccccc}
\toprule
\multirow{2}{*}{\textbf{Model}}  & \multicolumn{3}{c}{\textbf{Interaction Type}} & \multicolumn{3}{c}{\textbf{Subtask Count}} & \multicolumn{5}{c}{\textbf{Spatial Relation}} & \multicolumn{4}{c}{\textbf{Object Attribute}}\\
\cmidrule(lr){2-4} \cmidrule(lr){5-7} \cmidrule(lr){8-12} \cmidrule(lr){13-16}
 &Placement &State Change &Composite Actions &Single Action & Two Actions& Three or More Actions & Surface & Container & Left–Right Relation & Ordinal Region & Relative Distance & Color & Size & Pattern & Category\\
\toprule

\rowcolor{gray!10}
\multirow{1}{*}{OpenVLA-OFT~(1.0$\times$)} & \textbf{4.7} & \textbf{0.4} & \textbf{0.0} & \textbf{5.6} & \textbf{0.2} & \textbf{0.0} & \textbf{3.7} & \textbf{4.8} & \textbf{5.0} & \textbf{6.3} & \textbf{0.0} & \textbf{5.6} & \textbf{2.0} & \textbf{4.4} & \textbf{1.2} \\
\multirow{1}{*}{OpenVLA-OFT~(1.1$\times$)} & \textbf{6.6} {\scriptsize (+1.9)} & \textbf{0.5} {\scriptsize (+0.1)} & \textbf{0.0} {\scriptsize (+0.0)} & \textbf{7.8} {\scriptsize (+2.2)} & \textbf{0.3} {\scriptsize (+0.1)} & \textbf{0.0} {\scriptsize (+0.0)} & \textbf{5.7} {\scriptsize (+2.0)} & \textbf{6.3} {\scriptsize (+1.5)} & \textbf{7.1} {\scriptsize (+2.1)} & \textbf{6.8} {\scriptsize (+0.5)} & \textbf{0.0} {\scriptsize (+0.0)} & \textbf{8.0} {\scriptsize (+2.4)} & \textbf{3.4} {\scriptsize (+1.4)} & \textbf{6.2} {\scriptsize (+1.8)} & \textbf{1.6} {\scriptsize (+0.4)} \\
\multirow{1}{*}{OpenVLA-OFT~(1.5$\times$)} & \textbf{9.7} {\scriptsize (+3.1)} & \textbf{0.7} {\scriptsize (+0.2)} & \textbf{0.0} {\scriptsize (+0.0)} & \textbf{11.5} {\scriptsize (+3.7)} & \textbf{0.5} {\scriptsize (+0.2)} & \textbf{0.0} {\scriptsize (+0.0)} & \textbf{8.9} {\scriptsize (+3.2)} & \textbf{9.0} {\scriptsize (+2.7)} & \textbf{9.6} {\scriptsize (+2.5)} & \textbf{10.0} {\scriptsize (+3.2)} & \textbf{0.0} {\scriptsize (+0.0)} & \textbf{11.8} {\scriptsize (+3.8)} & \textbf{6.4} {\scriptsize (+3.0)} & \textbf{9.6} {\scriptsize (+3.4)} & \textbf{2.9} {\scriptsize (+1.3)} \\
\multirow{1}{*}{OpenVLA-OFT~(\(\infty\))} & \textbf{11.9} {\scriptsize (+2.2)} & \textbf{2.8} {\scriptsize (+2.1)} & \textbf{0.3} {\scriptsize (+0.3)} & \textbf{14.4} {\scriptsize (+2.9)} & \textbf{0.7} {\scriptsize (+0.2)} & \textbf{0.0} {\scriptsize (+0.0)} & \textbf{11.2} {\scriptsize (+2.3)} & \textbf{11.1} {\scriptsize (+2.1)} & \textbf{11.2} {\scriptsize (+1.6)} & \textbf{16.3} {\scriptsize (+6.3)} & \textbf{0.0} {\scriptsize (+0.0)} & \textbf{14.6} {\scriptsize (+2.8)} & \textbf{8.1} {\scriptsize (+1.7)} & \textbf{12.8} {\scriptsize (+3.2)} & \textbf{4.2} {\scriptsize (+1.3)} \\
\midrule
\rowcolor{gray!10}
\multirow{1}{*}{$\pi_{0}$~(1.0$\times$)} & \textbf{5.1} & \textbf{5.6} & \textbf{1.3} & \textbf{6.4} & \textbf{0.7} & \textbf{0.0} & \textbf{6.4} & \textbf{4.1} & \textbf{4.6} & \textbf{8.5} & \textbf{0.0} & \textbf{6.3} & \textbf{2.4} & \textbf{5.6} & \textbf{2.8} \\
\multirow{1}{*}{$\pi_{0}$~(1.1$\times$)} & \textbf{7.5} {\scriptsize (+2.4)} & \textbf{6.8} {\scriptsize (+1.2)} & \textbf{1.7} {\scriptsize (+0.4)} & \textbf{9.3} {\scriptsize (+2.9)} & \textbf{0.8} {\scriptsize (+0.1)} & \textbf{0.0} {\scriptsize (+0.0)} & \textbf{9.1} {\scriptsize (+2.7)} & \textbf{6.0} {\scriptsize (+1.9)} & \textbf{6.8} {\scriptsize (+2.2)} & \textbf{12.3} {\scriptsize (+3.8)} & \textbf{0.0} {\scriptsize (+0.0)} & \textbf{9.1} {\scriptsize (+2.8)} & \textbf{3.5} {\scriptsize (+1.1)} & \textbf{8.9} {\scriptsize (+3.3)} & \textbf{3.7} {\scriptsize (+0.9)} \\
\multirow{1}{*}{$\pi_{0}$~(1.5$\times$)} & \textbf{11.7} {\scriptsize (+4.2)} & \textbf{9.8} {\scriptsize (+3.0)} & \textbf{2.0} {\scriptsize (+0.3)} & \textbf{14.4} {\scriptsize (+5.1)} & \textbf{1.5} {\scriptsize (+0.7)} & \textbf{0.0} {\scriptsize (+0.0)} & \textbf{14.1} {\scriptsize (+5.0)} & \textbf{9.4} {\scriptsize (+3.4)} & \textbf{10.0} {\scriptsize (+3.2)} & \textbf{19.0} {\scriptsize (+6.7)} & \textbf{0.0} {\scriptsize (+0.0)} & \textbf{14.4} {\scriptsize (+5.3)} & \textbf{6.6} {\scriptsize (+3.1)} & \textbf{14.6} {\scriptsize (+5.7)} & \textbf{5.8} {\scriptsize (+2.1)} \\
\multirow{1}{*}{$\pi_{0}$~(\(\infty\))} & \textbf{14.3} {\scriptsize (+2.6)} & \textbf{12.8} {\scriptsize (+3.0)} & \textbf{2.3} {\scriptsize (+0.3)} & \textbf{17.8} {\scriptsize (+3.4)} & \textbf{2.1} {\scriptsize (+0.6)} & \textbf{0.0} {\scriptsize (+0.0)} & \textbf{16.9} {\scriptsize (+2.8)} & \textbf{11.9} {\scriptsize (+2.5)} & \textbf{12.6} {\scriptsize (+2.6)} & \textbf{24.2} {\scriptsize (+5.2)} & \textbf{2.9} {\scriptsize (+2.9)} & \textbf{17.7} {\scriptsize (+3.3)} & \textbf{8.9} {\scriptsize (+2.3)} & \textbf{18.1} {\scriptsize (+3.5)} & \textbf{7.0} {\scriptsize (+1.2)} \\
\midrule
\rowcolor{gray!10}
\multirow{1}{*}{X-VLA~(1.0$\times$)} & \textbf{4.2} & \textbf{2.1} & \textbf{0.0} & \textbf{5.2} & \textbf{0.4} & \textbf{0.0} & \textbf{4.9} & \textbf{3.4} & \textbf{3.6} & \textbf{3.2} & \textbf{0.0} & \textbf{5.4} & \textbf{4.7} & \textbf{3.6} & \textbf{1.7} \\
\multirow{1}{*}{X-VLA~(1.1$\times$)} & \textbf{6.0} {\scriptsize (+1.8)} & \textbf{3.0} {\scriptsize (+0.9)} & \textbf{0.3} {\scriptsize (+0.3)} & \textbf{7.2} {\scriptsize (+2.0)} & \textbf{0.8} {\scriptsize (+0.4)} & \textbf{0.0} {\scriptsize (+0.0)} & \textbf{6.9} {\scriptsize (+2.0)} & \textbf{4.9} {\scriptsize (+1.5)} & \textbf{5.2} {\scriptsize (+1.6)} & \textbf{4.4} {\scriptsize (+1.2)} & \textbf{0.0} {\scriptsize (+0.0)} & \textbf{7.5} {\scriptsize (+2.1)} & \textbf{5.9} {\scriptsize (+1.2)} & \textbf{4.7} {\scriptsize (+1.1)} & \textbf{2.8} {\scriptsize (+1.1)} \\
\multirow{1}{*}{X-VLA~(1.5$\times$)} & \textbf{10.2} {\scriptsize (+4.2)} & \textbf{4.9} {\scriptsize (+1.9)} & \textbf{0.3} {\scriptsize (+0.0)} & \textbf{12.4} {\scriptsize (+5.2)} & \textbf{1.1} {\scriptsize (+0.3)} & \textbf{0.2} {\scriptsize (+0.2)} & \textbf{11.8} {\scriptsize (+4.9)} & \textbf{8.4} {\scriptsize (+3.5)} & \textbf{8.5} {\scriptsize (+3.3)} & \textbf{9.7} {\scriptsize (+5.3)} & \textbf{1.4} {\scriptsize (+1.4)} & \textbf{12.5} {\scriptsize (+5.0)} & \textbf{10.2} {\scriptsize (+4.3)} & \textbf{8.7} {\scriptsize (+4.0)} & \textbf{5.0} {\scriptsize (+2.2)} \\
\multirow{1}{*}{X-VLA~(\(\infty\))} & \textbf{12.9} {\scriptsize (+2.7)} & \textbf{7.5} {\scriptsize (+2.6)} & \textbf{0.7} {\scriptsize (+0.4)} & \textbf{15.8} {\scriptsize (+3.4)} & \textbf{1.5} {\scriptsize (+0.4)} & \textbf{0.2} {\scriptsize (+0.0)} & \textbf{14.3} {\scriptsize (+2.5)} & \textbf{11.2} {\scriptsize (+2.8)} & \textbf{11.0} {\scriptsize (+2.5)} & \textbf{13.0} {\scriptsize (+3.3)} & \textbf{1.4} {\scriptsize (+0.0)} & \textbf{15.8} {\scriptsize (+3.3)} & \textbf{14.2} {\scriptsize (+4.0)} & \textbf{11.4} {\scriptsize (+2.7)} & \textbf{6.0} {\scriptsize (+1.0)} \\
\midrule
\rowcolor{gray!10}
\multirow{1}{*}{GR00T1.5~(1.0$\times$)} & \textbf{10.0} & \textbf{6.8} & \textbf{1.7} & \textbf{12.3} & \textbf{1.2} & \textbf{0.0} & \textbf{13.1} & \textbf{7.2} & \textbf{7.6} & \textbf{14.4} & \textbf{0.0} & \textbf{12.3} & \textbf{11.0} & \textbf{8.5} & \textbf{3.7} \\
\multirow{1}{*}{GR00T1.5~(1.1$\times$)} & \textbf{13.1} {\scriptsize (+3.1)} & \textbf{9.5} {\scriptsize (+2.7)} & \textbf{2.0} {\scriptsize (+0.3)} & \textbf{16.0} {\scriptsize (+3.7)} & \textbf{2.1} {\scriptsize (+0.9)} & \textbf{0.0} {\scriptsize (+0.0)} & \textbf{17.6} {\scriptsize (+4.5)} & \textbf{9.2} {\scriptsize (+2.0)} & \textbf{9.7} {\scriptsize (+2.1)} & \textbf{18.9} {\scriptsize (+4.5)} & \textbf{0.0} {\scriptsize (+0.0)} & \textbf{16.2} {\scriptsize (+3.9)} & \textbf{13.4} {\scriptsize (+2.4)} & \textbf{12.3} {\scriptsize (+3.8)} & \textbf{4.8} {\scriptsize (+1.1)} \\
\multirow{1}{*}{GR00T1.5~(1.5$\times$)} & \textbf{18.1} {\scriptsize (+5.0)} & \textbf{16.7} {\scriptsize (+7.2)} & \textbf{3.0} {\scriptsize (+1.0)} & \textbf{22.4} {\scriptsize (+6.4)} & \textbf{3.2} {\scriptsize (+1.1)} & \textbf{0.0} {\scriptsize (+0.0)} & \textbf{25.1} {\scriptsize (+7.5)} & \textbf{12.8} {\scriptsize (+3.6)} & \textbf{12.9} {\scriptsize (+3.2)} & \textbf{27.2} {\scriptsize (+8.3)} & \textbf{1.4} {\scriptsize (+1.4)} & \textbf{22.1} {\scriptsize (+5.9)} & \textbf{20.3} {\scriptsize (+6.9)} & \textbf{16.2} {\scriptsize (+3.9)} & \textbf{8.8} {\scriptsize (+4.0)} \\
\multirow{1}{*}{GR00T1.5~(\(\infty\))} & \textbf{21.5} {\scriptsize (+3.4)} & \textbf{24.6} {\scriptsize (+7.9)} & \textbf{3.3} {\scriptsize (+0.3)} & \textbf{27.0} {\scriptsize (+4.6)} & \textbf{4.1} {\scriptsize (+0.9)} & \textbf{0.0} {\scriptsize (+0.0)} & \textbf{31.0} {\scriptsize (+5.9)} & \textbf{15.1} {\scriptsize (+2.3)} & \textbf{15.2} {\scriptsize (+2.3)} & \textbf{33.1} {\scriptsize (+5.9)} & \textbf{4.3} {\scriptsize (+2.9)} & \textbf{26.5} {\scriptsize (+4.4)} & \textbf{23.9} {\scriptsize (+3.6)} & \textbf{19.1} {\scriptsize (+2.9)} & \textbf{12.3} {\scriptsize (+3.5)} \\
\midrule
\rowcolor{gray!10}
\multirow{1}{*}{$\pi_{0.5}$~(1.0$\times$)} & \textbf{17.3} & \textbf{16.7} & \textbf{4.0} & \textbf{21.2} & \textbf{3.5} & \textbf{2.0} & \textbf{18.2} & \textbf{16.3} & \textbf{17.5} & \textbf{22.4} & \textbf{4.3} & \textbf{21.5} & \textbf{15.7} & \textbf{16.5} & \textbf{9.7} \\
\multirow{1}{*}{$\pi_{0.5}$~(1.1$\times$)} & \textbf{23.4} {\scriptsize (+6.1)} & \textbf{25.3} {\scriptsize (+8.6)} & \textbf{7.7} {\scriptsize (+3.7)} & \textbf{28.5} {\scriptsize (+7.3)} & \textbf{6.1} {\scriptsize (+2.6)} & \textbf{2.8} {\scriptsize (+0.8)} & \textbf{26.1} {\scriptsize (+7.9)} & \textbf{21.2} {\scriptsize (+4.9)} & \textbf{22.7} {\scriptsize (+5.2)} & \textbf{29.0} {\scriptsize (+6.6)} & \textbf{7.1} {\scriptsize (+2.8)} & \textbf{29.0} {\scriptsize (+7.5)} & \textbf{21.7} {\scriptsize (+6.0)} & \textbf{22.4} {\scriptsize (+5.9)} & \textbf{14.4} {\scriptsize (+4.7)} \\
\multirow{1}{*}{$\pi_{0.5}$~(1.5$\times$)} & \textbf{30.5} {\scriptsize (+7.1)} & \textbf{39.1} {\scriptsize (+13.8)} & \textbf{13.0} {\scriptsize (+5.3)} & \textbf{37.4} {\scriptsize (+8.9)} & \textbf{9.6} {\scriptsize (+3.5)} & \textbf{3.5} {\scriptsize (+0.7)} & \textbf{35.6} {\scriptsize (+9.5)} & \textbf{27.5} {\scriptsize (+6.3)} & \textbf{28.1} {\scriptsize (+5.4)} & \textbf{39.4} {\scriptsize (+10.4)} & \textbf{18.6} {\scriptsize (+11.5)} & \textbf{36.9} {\scriptsize (+7.9)} & \textbf{31.8} {\scriptsize (+10.1)} & \textbf{29.3} {\scriptsize (+6.9)} & \textbf{21.6} {\scriptsize (+7.2)} \\
\multirow{1}{*}{$\pi_{0.5}$~(\(\infty\))} & \textbf{38.0} {\scriptsize (+7.5)} & \textbf{43.2} {\scriptsize (+4.1)} & \textbf{13.3} {\scriptsize (+0.3)} & \textbf{46.5} {\scriptsize (+9.1)} & \textbf{11.0} {\scriptsize (+1.4)} & \textbf{3.7} {\scriptsize (+0.2)} & \textbf{42.7} {\scriptsize (+7.1)} & \textbf{35.1} {\scriptsize (+7.6)} & \textbf{34.6} {\scriptsize (+6.5)} & \textbf{49.0} {\scriptsize (+9.6)} & \textbf{21.4} {\scriptsize (+2.8)} & \textbf{45.5} {\scriptsize (+8.6)} & \textbf{38.0} {\scriptsize (+6.2)} & \textbf{36.7} {\scriptsize (+7.4)} & \textbf{25.7} {\scriptsize (+4.1)} \\

\bottomrule
\end{tabular}}

\vspace{-0.5cm}
\label{tab:app_multi_label_level4}
\end{table*}

\begin{table*}[!t]
\centering
\vspace{5pt}
\caption{
    \textnormal{\textbf{Results of Fine-grained Multi-label Evaluation and Different Time Limit Thresholds in \textcolor{red}{LEVEL 5}.}}
}
\setlength\tabcolsep{3pt}  
\renewcommand{\arraystretch}{1.2} 
\resizebox{1.0\linewidth}{!}{
\begin{tabular}{lccccccccccccccc}
\toprule
\multirow{2}{*}{\textbf{Model}}  & \multicolumn{3}{c}{\textbf{Interaction Type}} & \multicolumn{3}{c}{\textbf{Subtask Count}} & \multicolumn{5}{c}{\textbf{Spatial Relation}} & \multicolumn{4}{c}{\textbf{Object Attribute}}\\
\cmidrule(lr){2-4} \cmidrule(lr){5-7} \cmidrule(lr){8-12} \cmidrule(lr){13-16}
 &Placement &State Change &Composite Actions &Single Action & Two Actions& Three or More Actions & Surface & Container & Left–Right Relation & Ordinal Region & Relative Distance & Color & Size & Pattern & Category\\
\toprule

\rowcolor{gray!10}
\multirow{1}{*}{OpenVLA-OFT~(1.0$\times$)} & \textbf{3.0} & \textbf{0.5} & \textbf{0.0} & \textbf{3.6} & \textbf{0.2} & \textbf{0.0} & \textbf{2.5} & \textbf{3.1} & \textbf{3.1} & \textbf{5.2} & \textbf{0.0} & \textbf{3.6} & \textbf{1.2} & \textbf{3.1} & \textbf{0.8} \\
\multirow{1}{*}{OpenVLA-OFT~(1.1$\times$)} & \textbf{4.3} {\scriptsize (+1.3)} & \textbf{0.5} {\scriptsize (+0.0)} & \textbf{0.0} {\scriptsize (+0.0)} & \textbf{5.1} {\scriptsize (+1.5)} & \textbf{0.2} {\scriptsize (+0.0)} & \textbf{0.0} {\scriptsize (+0.0)} & \textbf{3.8} {\scriptsize (+1.3)} & \textbf{4.1} {\scriptsize (+1.0)} & \textbf{4.4} {\scriptsize (+1.3)} & \textbf{5.9} {\scriptsize (+0.7)} & \textbf{0.0} {\scriptsize (+0.0)} & \textbf{5.2} {\scriptsize (+1.6)} & \textbf{1.6} {\scriptsize (+0.4)} & \textbf{4.5} {\scriptsize (+1.4)} & \textbf{1.3} {\scriptsize (+0.5)} \\
\multirow{1}{*}{OpenVLA-OFT~(1.5$\times$)} & \textbf{6.4} {\scriptsize (+2.1)} & \textbf{0.7} {\scriptsize (+0.2)} & \textbf{0.0} {\scriptsize (+0.0)} & \textbf{7.6} {\scriptsize (+2.5)} & \textbf{0.3} {\scriptsize (+0.1)} & \textbf{0.0} {\scriptsize (+0.0)} & \textbf{6.0} {\scriptsize (+2.2)} & \textbf{5.8} {\scriptsize (+1.7)} & \textbf{6.1} {\scriptsize (+1.7)} & \textbf{7.5} {\scriptsize (+1.6)} & \textbf{0.0} {\scriptsize (+0.0)} & \textbf{7.8} {\scriptsize (+2.6)} & \textbf{3.4} {\scriptsize (+1.8)} & \textbf{7.1} {\scriptsize (+2.6)} & \textbf{1.9} {\scriptsize (+0.6)} \\
\multirow{1}{*}{OpenVLA-OFT~(\(\infty\))} & \textbf{8.1} {\scriptsize (+1.7)} & \textbf{1.2} {\scriptsize (+0.5)} & \textbf{0.0} {\scriptsize (+0.0)} & \textbf{9.8} {\scriptsize (+2.2)} & \textbf{0.4} {\scriptsize (+0.1)} & \textbf{0.0} {\scriptsize (+0.0)} & \textbf{7.6} {\scriptsize (+1.6)} & \textbf{7.6} {\scriptsize (+1.8)} & \textbf{7.4} {\scriptsize (+1.3)} & \textbf{12.5} {\scriptsize (+5.0)} & \textbf{0.0} {\scriptsize (+0.0)} & \textbf{10.1} {\scriptsize (+2.3)} & \textbf{4.8} {\scriptsize (+1.4)} & \textbf{9.6} {\scriptsize (+2.5)} & \textbf{2.5} {\scriptsize (+0.6)} \\
\midrule
\rowcolor{gray!10}
\multirow{1}{*}{$\pi_{0}$~(1.0$\times$)} & \textbf{3.5} & \textbf{3.5} & \textbf{0.0} & \textbf{4.4} & \textbf{0.4} & \textbf{0.0} & \textbf{4.9} & \textbf{2.4} & \textbf{2.7} & \textbf{7.7} & \textbf{0.0} & \textbf{4.6} & \textbf{1.9} & \textbf{4.3} & \textbf{1.4} \\
\multirow{1}{*}{$\pi_{0}$~(1.1$\times$)} & \textbf{5.0} {\scriptsize (+1.5)} & \textbf{4.6} {\scriptsize (+1.1)} & \textbf{0.0} {\scriptsize (+0.0)} & \textbf{6.3} {\scriptsize (+1.9)} & \textbf{0.4} {\scriptsize (+0.0)} & \textbf{0.0} {\scriptsize (+0.0)} & \textbf{6.8} {\scriptsize (+1.9)} & \textbf{3.6} {\scriptsize (+1.2)} & \textbf{4.1} {\scriptsize (+1.4)} & \textbf{10.0} {\scriptsize (+2.3)} & \textbf{0.0} {\scriptsize (+0.0)} & \textbf{6.2} {\scriptsize (+1.6)} & \textbf{2.6} {\scriptsize (+0.7)} & \textbf{6.2} {\scriptsize (+1.9)} & \textbf{2.2} {\scriptsize (+0.8)} \\
\multirow{1}{*}{$\pi_{0}$~(1.5$\times$)} & \textbf{8.6} {\scriptsize (+3.6)} & \textbf{8.2} {\scriptsize (+3.6)} & \textbf{0.3} {\scriptsize (+0.3)} & \textbf{10.9} {\scriptsize (+4.6)} & \textbf{0.6} {\scriptsize (+0.2)} & \textbf{0.0} {\scriptsize (+0.0)} & \textbf{12.3} {\scriptsize (+5.5)} & \textbf{5.9} {\scriptsize (+2.3)} & \textbf{6.5} {\scriptsize (+2.4)} & \textbf{16.8} {\scriptsize (+6.8)} & \textbf{0.0} {\scriptsize (+0.0)} & \textbf{10.7} {\scriptsize (+4.5)} & \textbf{5.9} {\scriptsize (+3.3)} & \textbf{11.2} {\scriptsize (+5.0)} & \textbf{4.1} {\scriptsize (+1.9)} \\
\multirow{1}{*}{$\pi_{0}$~(\(\infty\))} & \textbf{11.3} {\scriptsize (+2.7)} & \textbf{11.2} {\scriptsize (+3.0)} & \textbf{0.7} {\scriptsize (+0.4)} & \textbf{14.2} {\scriptsize (+3.3)} & \textbf{1.0} {\scriptsize (+0.4)} & \textbf{0.0} {\scriptsize (+0.0)} & \textbf{15.7} {\scriptsize (+3.4)} & \textbf{7.9} {\scriptsize (+2.0)} & \textbf{8.3} {\scriptsize (+1.8)} & \textbf{22.5} {\scriptsize (+5.7)} & \textbf{0.0} {\scriptsize (+0.0)} & \textbf{13.8} {\scriptsize (+3.1)} & \textbf{8.4} {\scriptsize (+2.5)} & \textbf{13.8} {\scriptsize (+2.6)} & \textbf{5.6} {\scriptsize (+1.5)} \\
\midrule
\rowcolor{gray!10}
\multirow{1}{*}{X-VLA~(1.0$\times$)} & \textbf{2.7} & \textbf{2.6} & \textbf{0.3} & \textbf{3.4} & \textbf{0.1} & \textbf{0.0} & \textbf{4.0} & \textbf{1.8} & \textbf{1.9} & \textbf{2.0} & \textbf{0.0} & \textbf{3.6} & \textbf{3.1} & \textbf{1.7} & \textbf{1.2} \\
\multirow{1}{*}{X-VLA~(1.1$\times$)} & \textbf{4.0} {\scriptsize (+1.3)} & \textbf{3.3} {\scriptsize (+0.7)} & \textbf{0.3} {\scriptsize (+0.0)} & \textbf{5.0} {\scriptsize (+1.6)} & \textbf{0.3} {\scriptsize (+0.2)} & \textbf{0.0} {\scriptsize (+0.0)} & \textbf{5.8} {\scriptsize (+1.8)} & \textbf{2.7} {\scriptsize (+0.9)} & \textbf{2.9} {\scriptsize (+1.0)} & \textbf{2.4} {\scriptsize (+0.4)} & \textbf{0.0} {\scriptsize (+0.0)} & \textbf{5.3} {\scriptsize (+1.7)} & \textbf{4.8} {\scriptsize (+1.7)} & \textbf{2.5} {\scriptsize (+0.8)} & \textbf{1.7} {\scriptsize (+0.5)} \\
\multirow{1}{*}{X-VLA~(1.5$\times$)} & \textbf{7.0} {\scriptsize (+3.0)} & \textbf{5.1} {\scriptsize (+1.8)} & \textbf{0.3} {\scriptsize (+0.0)} & \textbf{8.7} {\scriptsize (+3.7)} & \textbf{0.5} {\scriptsize (+0.2)} & \textbf{0.0} {\scriptsize (+0.0)} & \textbf{10.1} {\scriptsize (+4.3)} & \textbf{4.7} {\scriptsize (+2.0)} & \textbf{5.0} {\scriptsize (+2.1)} & \textbf{6.1} {\scriptsize (+3.7)} & \textbf{0.0} {\scriptsize (+0.0)} & \textbf{8.6} {\scriptsize (+3.3)} & \textbf{9.0} {\scriptsize (+4.2)} & \textbf{4.7} {\scriptsize (+2.2)} & \textbf{3.6} {\scriptsize (+1.9)} \\
\multirow{1}{*}{X-VLA~(\(\infty\))} & \textbf{10.0} {\scriptsize (+3.0)} & \textbf{7.2} {\scriptsize (+2.1)} & \textbf{0.7} {\scriptsize (+0.4)} & \textbf{12.4} {\scriptsize (+3.7)} & \textbf{0.7} {\scriptsize (+0.2)} & \textbf{0.0} {\scriptsize (+0.0)} & \textbf{13.4} {\scriptsize (+3.3)} & \textbf{7.3} {\scriptsize (+2.6)} & \textbf{7.5} {\scriptsize (+2.5)} & \textbf{10.8} {\scriptsize (+4.7)} & \textbf{0.0} {\scriptsize (+0.0)} & \textbf{12.3} {\scriptsize (+3.7)} & \textbf{12.6} {\scriptsize (+3.6)} & \textbf{7.8} {\scriptsize (+3.1)} & \textbf{4.8} {\scriptsize (+1.2)} \\
\midrule
\rowcolor{gray!10}
\multirow{1}{*}{GR00T1.5~(1.0$\times$)} & \textbf{7.1} & \textbf{6.0} & \textbf{1.7} & \textbf{8.8} & \textbf{1.0} & \textbf{0.0} & \textbf{10.0} & \textbf{4.7} & \textbf{5.2} & \textbf{7.5} & \textbf{0.0} & \textbf{8.8} & \textbf{6.8} & \textbf{5.8} & \textbf{3.7} \\
\multirow{1}{*}{GR00T1.5~(1.1$\times$)} & \textbf{9.6} {\scriptsize (+2.5)} & \textbf{7.2} {\scriptsize (+1.2)} & \textbf{2.3} {\scriptsize (+0.6)} & \textbf{11.7} {\scriptsize (+2.9)} & \textbf{2.1} {\scriptsize (+1.1)} & \textbf{0.0} {\scriptsize (+0.0)} & \textbf{13.3} {\scriptsize (+3.3)} & \textbf{6.4} {\scriptsize (+1.7)} & \textbf{7.2} {\scriptsize (+2.0)} & \textbf{10.0} {\scriptsize (+2.5)} & \textbf{0.0} {\scriptsize (+0.0)} & \textbf{11.9} {\scriptsize (+3.1)} & \textbf{8.7} {\scriptsize (+1.9)} & \textbf{8.8} {\scriptsize (+3.0)} & \textbf{4.2} {\scriptsize (+0.5)} \\
\multirow{1}{*}{GR00T1.5~(1.5$\times$)} & \textbf{13.9} {\scriptsize (+4.3)} & \textbf{14.6} {\scriptsize (+7.4)} & \textbf{2.7} {\scriptsize (+0.4)} & \textbf{17.2} {\scriptsize (+5.5)} & \textbf{2.9} {\scriptsize (+0.8)} & \textbf{0.0} {\scriptsize (+0.0)} & \textbf{19.9} {\scriptsize (+6.6)} & \textbf{9.3} {\scriptsize (+2.9)} & \textbf{9.6} {\scriptsize (+2.4)} & \textbf{18.6} {\scriptsize (+8.6)} & \textbf{2.9} {\scriptsize (+2.9)} & \textbf{17.0} {\scriptsize (+5.1)} & \textbf{14.4} {\scriptsize (+5.7)} & \textbf{12.6} {\scriptsize (+3.8)} & \textbf{7.4} {\scriptsize (+3.2)} \\
\multirow{1}{*}{GR00T1.5~(\(\infty\))} & \textbf{17.3} {\scriptsize (+3.4)} & \textbf{22.5} {\scriptsize (+7.9)} & \textbf{3.3} {\scriptsize (+0.6)} & \textbf{21.8} {\scriptsize (+4.6)} & \textbf{3.6} {\scriptsize (+0.7)} & \textbf{0.0} {\scriptsize (+0.0)} & \textbf{25.7} {\scriptsize (+5.8)} & \textbf{11.5} {\scriptsize (+2.2)} & \textbf{11.6} {\scriptsize (+2.0)} & \textbf{26.1} {\scriptsize (+7.5)} & \textbf{4.3} {\scriptsize (+1.4)} & \textbf{21.3} {\scriptsize (+4.3)} & \textbf{18.8} {\scriptsize (+4.4)} & \textbf{15.5} {\scriptsize (+2.9)} & \textbf{10.5} {\scriptsize (+3.1)} \\
\midrule
\rowcolor{gray!10}
\multirow{1}{*}{$\pi_{0.5}$~(1.0$\times$)} & \textbf{12.5} & \textbf{16.5} & \textbf{3.3} & \textbf{15.7} & \textbf{2.1} & \textbf{1.1} & \textbf{14.3} & \textbf{11.5} & \textbf{11.8} & \textbf{13.7} & \textbf{4.3} & \textbf{15.4} & \textbf{13.0} & \textbf{10.4} & \textbf{9.3} \\
\multirow{1}{*}{$\pi_{0.5}$~(1.1$\times$)} & \textbf{17.1} {\scriptsize (+4.6)} & \textbf{22.8} {\scriptsize (+6.3)} & \textbf{5.3} {\scriptsize (+2.0)} & \textbf{21.5} {\scriptsize (+5.8)} & \textbf{3.9} {\scriptsize (+1.8)} & \textbf{1.5} {\scriptsize (+0.4)} & \textbf{20.1} {\scriptsize (+5.8)} & \textbf{15.6} {\scriptsize (+4.1)} & \textbf{15.6} {\scriptsize (+3.8)} & \textbf{19.0} {\scriptsize (+5.3)} & \textbf{7.1} {\scriptsize (+2.8)} & \textbf{21.1} {\scriptsize (+5.7)} & \textbf{18.5} {\scriptsize (+5.5)} & \textbf{14.8} {\scriptsize (+4.4)} & \textbf{12.4} {\scriptsize (+3.1)} \\
\multirow{1}{*}{$\pi_{0.5}$~(1.5$\times$)} & \textbf{23.0} {\scriptsize (+5.9)} & \textbf{33.3} {\scriptsize (+10.5)} & \textbf{8.7} {\scriptsize (+3.4)} & \textbf{28.5} {\scriptsize (+7.0)} & \textbf{7.3} {\scriptsize (+3.4)} & \textbf{2.4} {\scriptsize (+0.9)} & \textbf{28.4} {\scriptsize (+8.3)} & \textbf{20.3} {\scriptsize (+4.7)} & \textbf{19.3} {\scriptsize (+3.7)} & \textbf{27.0} {\scriptsize (+8.0)} & \textbf{15.7} {\scriptsize (+8.6)} & \textbf{27.5} {\scriptsize (+6.4)} & \textbf{28.7} {\scriptsize (+10.2)} & \textbf{19.5} {\scriptsize (+4.7)} & \textbf{18.2} {\scriptsize (+5.8)} \\
\multirow{1}{*}{$\pi_{0.5}$~(\(\infty\))} & \textbf{28.9} {\scriptsize (+5.9)} & \textbf{36.8} {\scriptsize (+3.5)} & \textbf{9.0} {\scriptsize (+0.3)} & \textbf{35.7} {\scriptsize (+7.2)} & \textbf{8.4} {\scriptsize (+1.1)} & \textbf{2.4} {\scriptsize (+0.0)} & \textbf{34.2} {\scriptsize (+5.8)} & \textbf{26.2} {\scriptsize (+5.9)} & \textbf{23.6} {\scriptsize (+4.3)} & \textbf{33.9} {\scriptsize (+6.9)} & \textbf{18.6} {\scriptsize (+2.9)} & \textbf{34.4} {\scriptsize (+6.9)} & \textbf{35.8} {\scriptsize (+7.1)} & \textbf{24.8} {\scriptsize (+5.3)} & \textbf{21.3} {\scriptsize (+3.1)} \\

\bottomrule
\end{tabular}}

\vspace{-0.5cm}
\label{tab:app_multi_label_level5}
\end{table*}

\begin{figure*}[htbp]
    \centering
    \includegraphics[width=\textwidth]{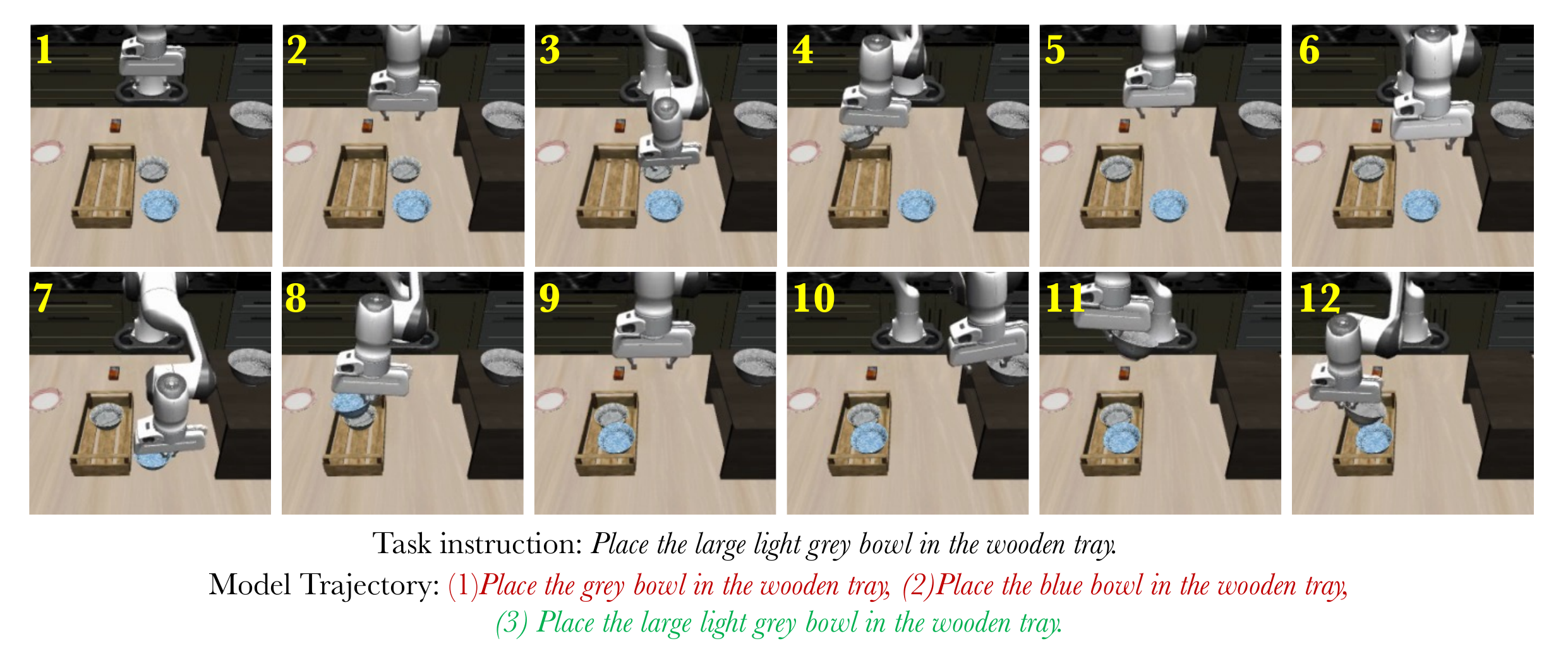} 
    \vspace{-15pt}
    \caption{\textbf{Visualization of the Model’s Enumeration Behavior under Increased Time Limits.}}
    \label{fig:time}
\end{figure*}

\subsection{Further Analysis of the Impact of Time Limits}
\label{sec:time_limits_analysis}

As shown in Tables \ref{tab:app_multi_label_level1}, \ref{tab:app_multi_label_level2}, \ref{tab:app_multi_label_level3}, \ref{tab:app_multi_label_level4}, and \ref{tab:app_multi_label_level5}, a consistent trend is observed across LEVEL 1 to LEVEL 5: success rates correlate positively with extended time limits. Specifically, with 1.0$\times$ representing the human reference time, the 1.1$\times$ threshold (a 10\% extension) yields the most significant performance gain. The improvement is attributed to the temporal buffer, which compensates for inevitable execution noise and environmental dynamics. Although models learn optimal trajectories from human demonstrations, they cannot perfectly replicate human precision. The additional time provides a critical margin for error correction and recovery from deviations. Consequently, 1.1$\times$ is established as the baseline time limit for our evaluation by default.

Between 1.1$\times$ to 1.5$\times$, success rates show continued but diminishing improvements. This gain is largely attributable to the increased window for trial-and-error interactions, allowing the model to exhaustively interact with objects rather than robust long-horizon planning (as shown in Figure \ref{fig:time}). Data from Table \ref{tab:app_multi_label_level1} underscores this limitation: while the $\pi_{0.5}$ model's success rate in the “Three or More Actions” category surges by 12.4\% from 1.0$\times$ to 1.1$\times$, it only increases by a marginal 1.9\% from 1.1$\times$ to 1.5$\times$. Beyond 1.5$\times$, performance plateaus. This saturation indicates that time is no longer the limiting factor; instead, the model is constrained by intrinsic bottlenecks in perception and reasoning. Consequently, further extending the time limit fails to yield significant gains on complex tasks.

\subsection{Model Details}
\label{sec:model_details}
In this section, we detail the experimental setup, covering model architectures, training data, and hyperparameter configurations. We provide an in-depth analysis of each model's structural design and the specific components that drive its performance. Furthermore, we describe the fine-tuning protocols employed to adapt these models to our evaluation tasks. This comprehensive background serves to contextualize the experimental results, elucidating the strengths and limitations of each architecture while establishing a baseline for future optimizations.

\begin{table}[!ht] 
    \caption{
        \textbf{Model Checkpoint Addresses.}
    }
    \centering
    \small
    \renewcommand\tabcolsep{6pt} 
    \renewcommand\arraystretch{1.0} 
    \resizebox{1.0\linewidth}{!}{ 
        \begin{tabular}{c|c}
            \toprule
            \rowcolor{gray!25} 
            \textbf{Model}
            & \textbf{Checkpoint Address} \\
            \midrule
            OpenVLA-OFT & \href{https://huggingface.co/openvla/openvla-7b}{https://huggingface.co/openvla/openvla-7b}\\
            $\pi_{0}$ &  \href{https://storage.googleapis.com/openpi-assets/checkpoints/pi0_base}{https://storage.googleapis.com/openpi-assets/checkpoints/pi0\_base}\\
            X-VLA & \href{https://huggingface.co/2toINF/X-VLA-Pt}{https://huggingface.co/2toINF/X-VLA-Pt}\\
            GR00T1.5 & \href{https://huggingface.co/nvidia/GR00T-N1.5-3B}{https://huggingface.co/nvidia/GR00T-N1.5-3B} \\
            $\pi_{0.5}$ & \href{https://storage.googleapis.com/openpi-assets/checkpoints/pi05_base}{https://storage.googleapis.com/openpi-assets/checkpoints/pi05\_base} \\
            \bottomrule
        \end{tabular}
    }
    \vspace{-6pt} 
    \label{tab:model_checkpoint}
\end{table}
As summarized in Table \ref{tab:model_checkpoint}, we evaluate a diverse set of VLA models. We initialize these models using their official pre-trained checkpoints and fine-tune them on the LIBERO-X dataset. This approach allows us to rigorously assess how different architectures adapt to complex manipulation scenarios within our proposed multi-level evaluation framework.

\begin{itemize}
    \item {OpenVLA-OFT~\cite{kim2025fine}}
\end{itemize}

\textbf{Model Structure:} OpenVLA-OFT~\cite{kim2025fine} is an optimized variant of OpenVLA~\cite{kim2024openvla}, engineered for real-time, high-frequency robotic control. It integrates parallel decoding and action chunking, enabling the generation of action sequences in a single inference pass. This design significantly enhances inference efficiency and reduces the need for frequent re-planning in long-horizon tasks. Furthermore, the model adopts continuous action representations trained with an L1 regression objective to accelerate convergence. To improve language grounding, FiLM (Feature-wise Linear Modulation)~\cite{perez2018film} is incorporated, ensuring precise adherence to linguistic instructions during execution.

\textbf{Training Data and Configuration:} On data preparation, we utilize \href{https://github.com/Tavish9/any4lerobot/}{Any4LeRobot} convert datasets from LeRobot format to RLDS/TFDS format. This conversion preserves the original temporal structure of episodes, mapping observations, actions, and instructions to standard RLDS fields for compatibility with downstream learning frameworks.

Fine-tuning was conducted using \texttt{torchrun} in a distributed setup (8 processes per node), initialized from the OpenVLA-7B checkpoint. The training configuration specified an L1 training objective, a batch size of 8, a learning rate of 5e-4, and a total of 150,000 steps. We employed LoRA (rank 32)~\cite{hu2022lora} for parameter-efficient adaptation. Image augmentation was used, while diffusion modules and FiLM were disabled in this specific configuration.

\begin{itemize}
    \item {$\pi_0$~\cite{black2410pi0} and $\pi_{0.5}$~\cite{intelligence2025pi05}}
\end{itemize}

\textbf{$\pi_0$ Model Structure:} The $\pi_0$ model is a VLA framework designed for general-purpose robot control, capable of handling complex and dexterous tasks. By combining a VLM with flow matching, $\pi_0$ generates continuous actions. It is pre-trained on diverse robot data (single/dual-arm, mobile) to ensure broad generalization across various hardware and complex multi-stage tasks.

\textbf{$\pi_{0.5}$ Model Structure:} An evolution of $\pi_0$ focused on broader semantic generalization. It incorporates internet-scale data and employs a two-stage training process. $\pi_{0.5}$ integrates high-level subtask prediction with low-level control, significantly improving performance in novel environments and reducing the need for task-specific retraining.

\textbf{Training Data and Configuration:} Visual inputs include resized 224$\times$224 wrist and third-person images. Both models are fine-tuned on LIBERO-X using pre-trained weights. 

\begin{itemize}
\item $\pi_0$: Fine-tuned via LoRA~\cite{hu2022lora} on 1 A100 GPU (Batch Size: 32, Action Chunk: 50).
\item $\pi_{0.5}$: Full-parameter fine-tuning on 4 A100 GPUs (Batch Size: 256, Action Chunk: 10).
\end{itemize}
Both models undergo 30,000 training iterations with standard normalization.

\begin{itemize}
    \item {X-VLA~\cite{zheng2025xvla}}
\end{itemize}

\textbf{Model Structure:} The X-VLA architecture is a scalable, cross-embodiment VLA framework designed for robust adaptation across diverse robotic platforms. It incorporates learnable soft prompts to condition the model on specific hardware configurations and task distributions. The architecture fuses visual observations, language instructions, and proprioceptive states via a Transformer encoder, while employing a flow-matching policy to refine continuous action outputs for dexterous execution.

\textbf{Training Data and Configuration:} Data preparation utilizes the LeRobot pipeline. Raw data is converted into the standard LeRobot format (parquet + multi-view video), and for each dataset, a \texttt{meta.json} file is generated, containing the dataset name, episode list, and instruction information. The corresponding dataset name is then registered in X-VLA. At training, the built-in processor dynamically tokenizes images and trajectories into the unified format required by the model.

Fine-tuning is conducted using the \texttt{accelerate} library with bf16 mixed precision to optimize computational efficiency. We initialize from a pre-trained checkpoint and train for 100,000 iterations with a batch size of 16. The optimization schedule includes a learning rate of 1e-4 and a coefficient of 0.1. To ensure stable adaptation, we employ a two-phase strategy: 1,000 steps of backbone freezing, followed by a 2,000-step learning rate warm-up.

\begin{itemize}
    \item {GR00T1.5~\cite{bjorck2025gr00t}}
\end{itemize}

\textbf{Model Structure:} GR00T N1.5 is a generalist humanoid foundation model developed by NVIDIA. Its architecture comprises three modular components: a frozen VLM for semantic embedding extraction, an Adapter for cross-modal alignment, and a Diffusion Transformer (DiT)~\cite{peebles2023scalable} for policy generation. The VLM encodes visual and linguistic inputs, which the Adapter projects into the DiT's latent space to synthesize actions via attention mechanisms. A critical innovation in N1.5 is FLARE (Future Latent Representation Alignment)~\cite{zhengflare}, a mechanism that aligns predicted future states with ground truth, significantly enhancing generalization in few-shot scenarios.

\textbf{Training Data and Configuration:} GR00T1.5 is fine-tuned from the GR00T-N1.5-3B checkpoint using min/max normalization for input scaling. Training is performed on a single GPU with a batch size of 32 over 60,000 steps. To optimize memory efficiency and stability, we employ gradient accumulation (4 steps).

\subsection{Scene Design Details}
\label{sec:scene_design}
To facilitate the design of more complex and diverse scenarios, we enhance the original LIBERO framework by optimizing core features and expanding the 3D asset library.

\textbf{(1) Addition of Novel Predicates}

The original LIBERO~\cite{liu2023libero} determines task success or failure by evaluating object relationships defined by predicates. To support a wider range of tasks, we additionally introduce new predicates to describe object relations, including \texttt{ExactIn}, \texttt{UprightOn}, and \texttt{SideOn}.

\begin{figure}[htbp]
    \centering
    \vspace{-5pt}
    \includegraphics[width=0.95\columnwidth]{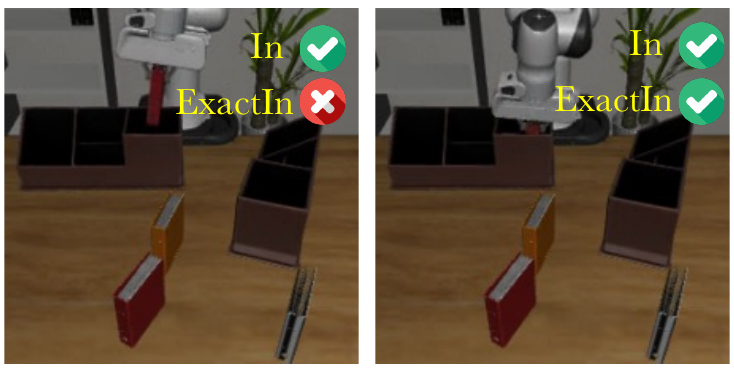} 
    \vspace{-5pt}
    \caption{\textbf{Comparison between \texttt{In} and \texttt{ExactIn}.}}
    \label{fig:in_and_exactin}
\end{figure}

As illustrated in Figure \ref{fig:in_and_exactin}, the legacy \texttt{In} logic suffers from imprecise success detection, often flagging a task as complete the moment an object breached the container's boundary, regardless of stability or depth. This results in frequent false positives where tasks were marked successful before the object was truly contained.

To rectify this, we introduce the \texttt{ExactIn} predicate. Unlike its predecessor, \texttt{ExactIn} enforces strict containment criteria by calculating the vertical (\texttt{z}-axis) displacement between the object and the container base. The predicate returns \texttt{true} only when this distance falls below a predefined threshold, confirming that the object is fully inserted and resting at the bottom. This mechanism effectively eliminates premature success signals, ensuring robust task evaluation.

\begin{figure}[htbp]
    \centering
    \vspace{-5pt}
    \includegraphics[width=0.95\columnwidth]{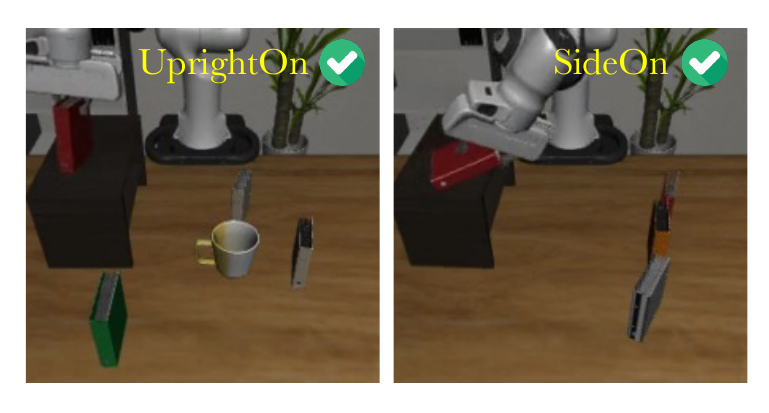} 
    \vspace{-5pt}
    \caption{\textbf{Comparison between \texttt{UprightOn} and \texttt{SideOn}.}}
    \label{fig:uprighton_and_sideon}
\end{figure}

In object placement tasks, accurately determining the posture of objects such as books is essential for successful execution. Unlike objects such as bowls or cups, books can be placed either upright or side-on, providing greater flexibility in placement. To improve posture recognition, we introduce two new predicates, as illustrated in Figure \ref{fig:uprighton_and_sideon}.

For upright placement, the object’s up-vector should be closely aligned with the surface normal. The object is classified as upright when the angle between the up-vector and the surface normal is less than approximately 37°.

For side-on placement, the object’s up-vector should be nearly perpendicular to the surface normal. The object is classified as side-on when the angle between the up-vector and the surface normal is within 17° of perpendicular.

This enhancement enables more precise posture recognition in object placement tasks, thereby improving task accuracy and system performance in complex environments. It supports reliable execution, particularly in scenarios that require clear differentiation between object postures.

\textbf{(2) Addition of Novel Objects} 

To increase the diversity of scenes and tasks, we expand the training and testing datasets. The new additions focus on enhancing color, texture, and size diversity, ensuring a broader range of objects for training. Additionally, we make full use of the functional zones of existing objects, such as different shelves in a bookshelf and the various drawers in a cabinet. For the testing dataset, we include objects that are not seen in the training set to assess the model's generalization ability.

\begin{figure}[htbp]
    \centering
    \includegraphics[width=1.0\columnwidth]{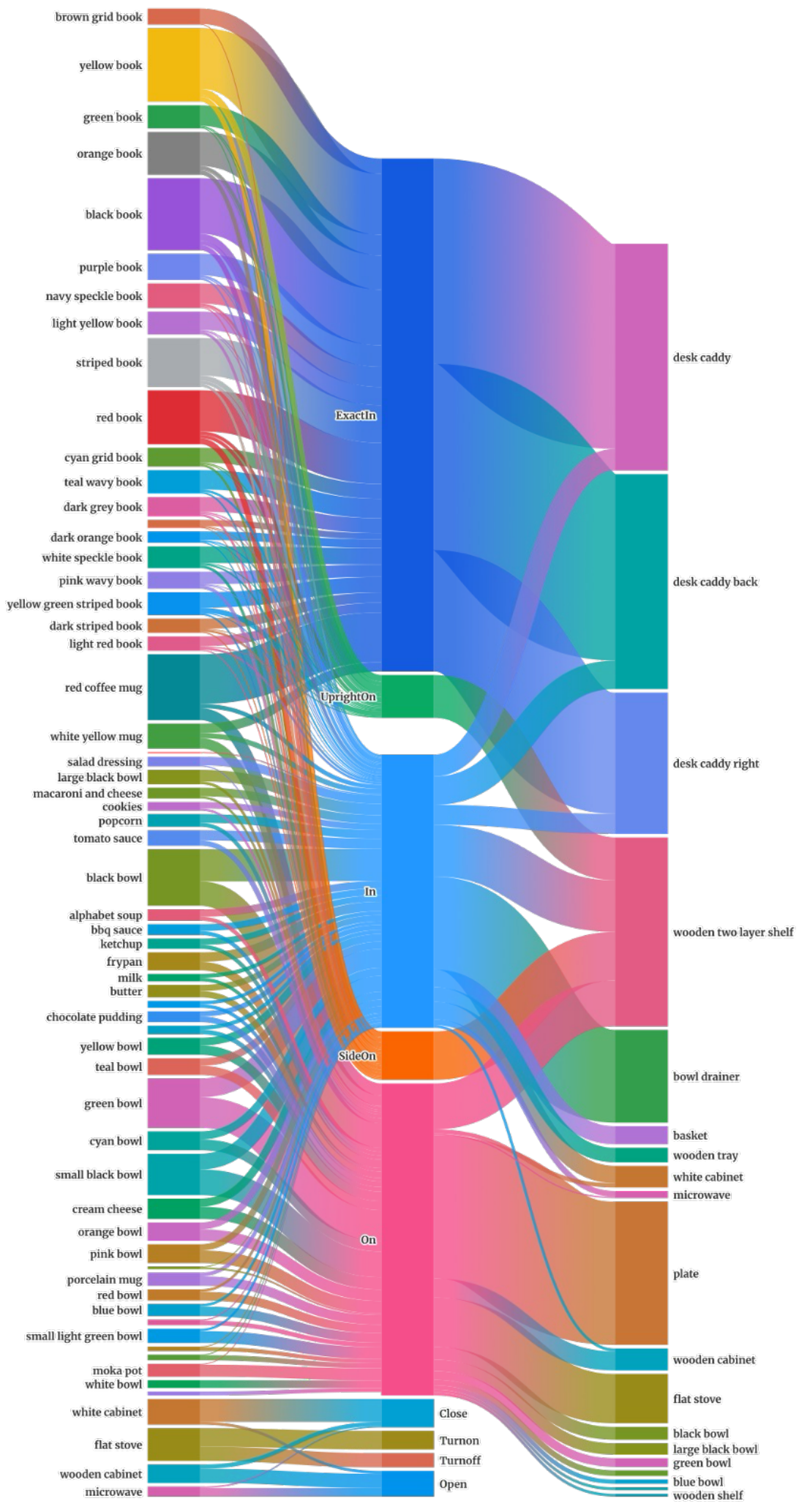} 
    \caption{\textbf{Sankey Diagram of the Flow between Predicate, Object, and Container.}}
    \label{fig:sankey_png}
\end{figure}

Table \ref{tab:train_object_frequency} presents the frequency of all Predicate, Object, and Container across various scenes and tasks in the training dataset. Table \ref{tab:test_object_frequency} provides the corresponding counts for all levels (from level 1 to level 5) in the test dataset. Figure \ref{fig:sankey_png} offers a detailed visualization of the data flow relationships between Predicate, Object, and Container.
\begin{table}[!ht]
    \caption{
        \textbf{Predicate, Object and Container Frequency in the Training Set across Scenes and Tasks.}
    }
    \centering
    \small
    \renewcommand\tabcolsep{6pt} 
    \renewcommand\arraystretch{1.2} 
    \resizebox{1.0\linewidth}{!}{ 
        \begin{tabular}{c|c|c|c|c|c}
            \toprule
            \rowcolor{gray!25} 
            \textbf{Name} & \textbf{Scene Count} & \textbf{Task Count} & \textbf{Name} & \textbf{Scene Count} & \textbf{Task Count} \\
            \midrule
            \rowcolor{gray!15} 
            \multicolumn{6}{c}{\textbf{Predicate}} \\
            ExactIn & 40 & 240 & On & 53 & 185 \\
            In & 41 & 116 & UprightOn & 10 & 27 \\
            SideOn & 9 & 25 & Close & 6 & 17 \\
            Open & 9 & 18 & Turnon & 8 & 13 \\
            Turnoff & 5 & 10 & \\
            \midrule
            \rowcolor{gray!15} 
            \multicolumn{6}{c}{\textbf{Object}} \\
            yellow book & 34 & 84 & black book & 29 & 79 \\
            red coffee mug & 37 & 77 & red book & 22 & 60 \\
            green bowl & 21 & 56 & striped book & 24 & 54 \\
            black bowl & 16 & 50 & orange book & 20 & 50 \\
            small black bowl & 9 & 29 & white yellow mug & 23 & 29 \\
            green book & 14 & 27 & flat stove & 13 & 23 \\
            frypan & 10 & 21 & yellow bowl & 9 & 20 \\
            white cabinet & 5 & 16 & moka pot & 10 & 15 \\
            blue bowl & 5 & 14 & porcelain mug & 4 & 14 \\
            red bowl & 7 & 13 & wooden cabinet & 7 & 12 \\
            large black bowl & 4 & 10 & white bowl & 4 & 9 \\
            wine bottle & 3 & 8 & cream cheese & 3 & 7 \\
            microwave & 3 & 7 & ramekin & 1 & 3 \\
            \midrule
            \rowcolor{gray!15} 
            \multicolumn{6}{c}{\textbf{Container}} \\
            desk caddy & 20 & 100 & back desk caddy & 20 & 95 \\
            wooden two layer shelf & 21 & 94 & right desk caddy & 10 & 60 \\
            plate & 26 & 80 & bowl drainer & 12 & 51 \\
            flat stove & 16 & 33 & white cabinet & 5 & 14 \\
            wooden cabinet & 4 & 13 & basket & 5 & 11 \\
            black bowl & 4 & 9 & wooden tray & 1 & 6 \\
            large black bowl & 3 & 8 & green bowl & 4 & 6 \\
            microwave & 3 & 5 & wine rack & 2 & 4 \\
            blue bowl & 3 & 3 & small black bowl & 2 & 2 \\
            wooden shelf & 1 & 2 & \\
            \bottomrule
        \end{tabular}
    }
    \vspace{-6pt} 
    \label{tab:train_object_frequency}
\end{table}
\begin{table}[!ht]
    \caption{
        \textbf{Predicate, Object and Container Frequency in the Test Set across Scenes and Tasks.}
    }
    \centering
    \small
    \renewcommand\tabcolsep{6pt} 
    \renewcommand\arraystretch{1.2} 
    \resizebox{1.0\linewidth}{!}{ 
        \begin{tabular}{c|c|c|c|c|c}
            \toprule
            \rowcolor{gray!25} 
            \textbf{Name} & \textbf{Scene Count} & \textbf{Task Count} & \textbf{Name} & \textbf{Scene Count} & \textbf{Task Count} \\
            \midrule
            \rowcolor{gray!15} 
            \multicolumn{6}{c}{\textbf{Predicate}} \\
            ExactIn & 211 & 530 & On & 238 & 428 \\
            In & 226 & 324 & SideOn & 33 & 65 \\
            UprightOn & 27 & 54 & Close & 23 & 34 \\
            Open & 27 & 36 & Turnon & 21 & 26 \\
            Turnoff & 15 & 20 & \\
            \midrule
            \rowcolor{gray!15} 
            \multicolumn{6}{c}{\textbf{Object}} \\
            yellow book & 34 & 84 & black book & 29 & 79 \\
            red coffee mug & 37 & 77 & black bowl & 40 & 74 \\
            red book & 22 & 60 & green bowl & 22 & 57 \\
            striped book & 24 & 54 & orange book & 20 & 50 \\
            small black bowl & 38 & 58 & flat stove & 36 & 46 \\
            purple book & 43 & 43 & white cabinet & 21 & 32 \\
            white yellow mug & 23 & 29 & navy speckle book & 40 & 40 \\
            teal wavy book & 39 & 39 & green book & 14 & 27 \\
            light yellow book & 40 & 40 & yellow green striped book & 39 & 39 \\
            white speckle book & 38 & 38 & cream cheese & 28 & 32 \\
            dark grey book & 34 & 34 & cyan bowl & 33 & 33 \\
            cyan grid book & 32 & 32 & pink bowl & 32 & 32 \\
            wooden cabinet & 19 & 24 & orange bowl & 32 & 32 \\
            frypan & 10 & 21 & pink wavy book & 30 & 30 \\
            yellow bowl & 9 & 20 & teal bowl & 29 & 29 \\
            brown grid book & 27 & 27 & tomato sauce & 26 & 26 \\
            dark striped book & 25 & 25 & large black bowl & 14 & 20 \\
            light red book & 25 & 25 & small light green bowl & 25 & 25 \\
            porcelain mug & 6 & 16 & moka pot & 10 & 15 \\
            butter & 22 & 22 & popcorn & 22 & 22 \\
            blue bowl & 5 & 14 & alphabet soup & 20 & 20 \\
            dark orange book & 20 & 20 & red bowl & 7 & 13 \\
            bbq sauce & 19 & 19 & chocolate pudding & 19 & 19 \\
            macaroni and cheese & 19 & 19 & ketchup & 18 & 18 \\
            microwave & 10 & 14 & salad dressing & 16 & 16 \\
            cookies & 15 & 15 & orange juice & 15 & 15 \\
            light green book & 14 & 14 & white bowl & 4 & 9 \\
            milk & 13 & 13 & wine bottle & 3 & 8 \\
            large light red bowl & 10 & 10 & small dark yellow bowl & 10 & 10 \\
            large dark blue bowl & 8 & 8 & large light grey bowl & 8 & 8 \\
            large off white bowl & 7 & 7 & ramekin & 1 & 3 \\
            \midrule
            \rowcolor{gray!15} 
            \multicolumn{6}{c}{\textbf{Container}} \\
            desk caddy & 120 & 250 & desk caddy back & 115 & 229 \\
            wooden two layer shelf & 115 & 229 & plate & 106 & 206 \\
            desk caddy right & 70 & 150 & bowl drainer & 63 & 102 \\
            flat stove & 49 & 70 & wooden cabinet & 17 & 32 \\
            white cabinet & 19 & 31 & basket & 16 & 26 \\
            wooden tray & 7 & 15 & black bowl & 13 & 18 \\
            large black bowl & 11 & 16 & green bowl & 10 & 12 \\
            microwave & 8 & 10 & wine rack & 6 & 8 \\
            blue bowl & 6 & 6 & small black bowl & 4 & 4 \\
            wooden shelf & 3 & 4 & \\
            \bottomrule
        \end{tabular}
    }
    \vspace{-6pt} 
    \label{tab:test_object_frequency}
\end{table}

\subsection{Visual Attribute Variation Cases Analysis}
\label{sec:visual_attribute_variation}

\begin{figure*}[htbp]
    \centering
    \includegraphics[width=\textwidth]{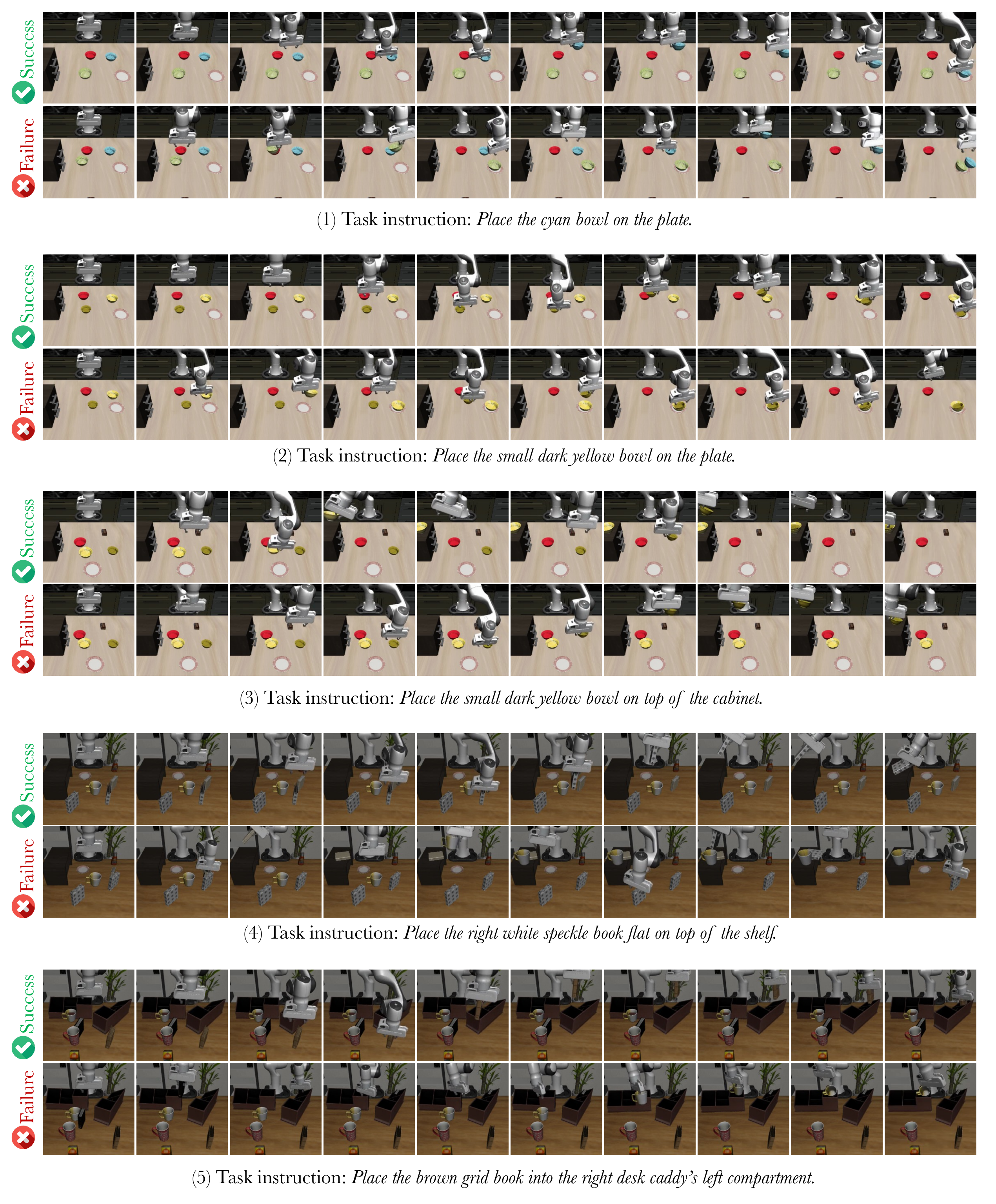} 
    \caption{\textbf{Visualization of Test Samples with Confounding Objects (CO) under Visual Attribute Variation (Level 4).}}
    \label{fig:example_level4_CO}
\end{figure*}

\begin{figure*}[htbp]
    \centering
    \includegraphics[width=\textwidth]{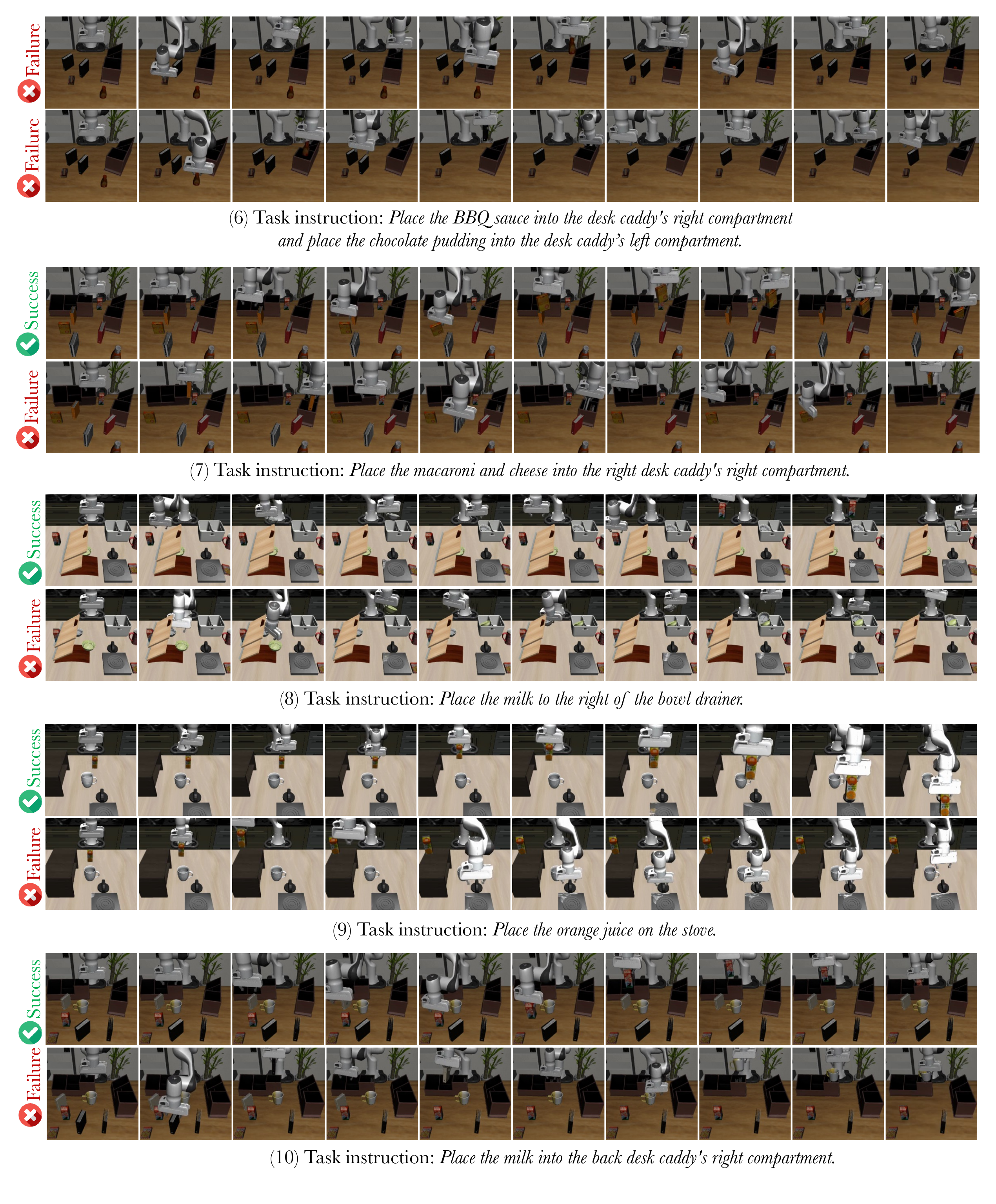} 
    \caption{\textbf{Visualization of Test Samples with Unseen Objects (UO) under Visual Attribute Variation (Level 4).}}
    \label{fig:example_level4_UO}
\end{figure*}

In this section, we analyze test cases under \textbf{Visual Attribute Variation (Level 4)}, focusing on \textbf{Confounding Objects (CO)} and \textbf{Unseen Objects (UO)}. These cases assess the model’s performance when exposed to variations in object attributes such as color, size, and texture, as well as when encountering objects that were not present in the training data. As shown in Figures \ref{fig:example_level4_CO} and \ref{fig:example_level4_UO}, the results presented here are based on tests conducted with the $\pi_{0.5}$ model.

\textbf{Incorrect Target Selection and Environmental Disruption:} The failure cases in Figure \ref{fig:example_level4_CO} (1), (2), and (3) show that when tasked with grabbing confounding objects, the model sometimes initially selects objects that resemble the target rather than the correct target itself. Although the model may later attempt to grasp the intended object, the initial incorrect selection disrupts the environment. These disruptions can alter the scene or occupy the target position, both of which contribute to task failure. This behavior suggests difficulty in distinguishing visually similar objects and maintaining task consistency in complex environments.

\textbf{Instruction Neglect and Target Deviation:} The failure cases in Figure \ref{fig:example_level4_CO} (4) and (5) and Figure \ref{fig:example_level4_UO} (7), (8), and (10) indicate that, despite receiving clear instructions, the model sometimes grasps objects unrelated to the task. It repeatedly attempts to interact with these irrelevant objects and fails to complete the task within the allotted time. This behavior reflects limitations in interpreting task instructions and in aligning them with the correct target object. The model also fails to effectively filter out confounding objects based on task requirements and instead focuses on irrelevant items, demonstrating weaknesses in target recognition and decision-making. These failure patterns highlight the need to improve instruction following and target selection accuracy.

\textbf{Multi-Step Trajectory Switching and Target Grabbing Difficulties:} As shown in failure case (6) in Figure \ref{fig:example_level4_UO}, the model may fail to complete the full task despite successfully executing the initial step. One contributing factor is difficulty in trajectory switching, which prevents smooth transitions between successive steps and disrupts task execution. In addition, the target object may become harder to grasp in later stages or may be misidentified due to environmental interference. These factors cause the model to perform adequately in early sub-tasks but ultimately fail to complete the multi-step task. This behavior indicates that, while the model can handle localized actions, it struggles with global task continuity. Improving overall planning and execution in multi-step scenarios, particularly ensuring effective transitions between steps, is therefore essential for enhancing performance.

\subsection{Semantic-Equivalent Reformulation Cases}
\label{sec:semantic_reformulation}
As shown in Table \ref{tab:example_level5}, \textbf{Semantic-Equivalent Reformulation (Level 5)} evaluates robustness to variations in task descriptions. The applied modification strategies include \textbf{synonym replacement (L5-1)}, which substitutes phrases using a synonym mapping table while maintaining the original meaning; \textbf{word compression (L5-2)}, which removes redundant elements such as auxiliary verbs and articles to retain only the core semantics; \textbf{word order adjustment (L5-3)}, which changes sentence structure without affecting meaning; \textbf{voice conversion (L5-4)}, which transforms imperative sentences into the passive voice; and \textbf{redundant descriptions (L5-5)}, which add supplementary phrases at the beginning or end of instructions to enhance clarity or emphasis.
\begin{table}[!ht]
    \vspace{-5pt} 
    \centering
    \caption{
        \textbf{Example of Semantic-Equivalent Reformulation (Level 5).}
    }
    \setlength\tabcolsep{4pt}
    \renewcommand\arraystretch{1.2} 
    \resizebox{1.0\linewidth}{!}{
        \begin{tabular}{c|c|c l}
        \hline 
        \rowcolor{gray!20} 
        \textbf{level5} & \textbf{Converted} & \textbf{Example} \\
        \hline
        \multirow{2}{*}{\textbf{5-1}}  
            & \multirow{1}{*}{Original} & \cellcolor{gray!5}{\darkblue{\textit{Place the pink bowl on the plate.}}} \\  
            \hhline{~|-|~} 
            & \multirow{1}{*}{Converted} & \cellcolor{gray!5}{\darkblue{\textit{Lay down the pink soup bowl over the serving plate.}}} \\ 
        
        \hline
        \multirow{2}{*}{\textbf{5-2}}  
            & \multirow{1}{*}{Original} & \cellcolor{gray!5}{\darkblue{\textit{Place the left dark grey book into the left desk caddy's back compartment.}}} \\  
            \hhline{~|-|~} 
            & \multirow{1}{*}{Converted} & \cellcolor{gray!5}{\darkblue{\textit{left dark grey book into left caddy back}}} \\ 
        \cline{2-2} 
        \hline
        \multirow{2}{*}{\textbf{5-3}}  
            & \multirow{1}{*}{Original} & \cellcolor{gray!5}{\darkblue{\textit{Place the small dark yellow bowl on top of the wooden cabinet.}}} \\  
            \hhline{~|-|~} 
            & \multirow{1}{*}{Converted} & \cellcolor{gray!5}{\darkblue{\textit{On top of the wooden cabinet, place the small dark yellow bowl.}}} \\ 
        \cline{2-2} 
        \hline
        \multirow{2}{*}{\textbf{5-4}}  
            & \multirow{1}{*}{Original} & \cellcolor{gray!5}{\darkblue{\textit{Place the navy speckle book in the top region of the shelf.}}} \\  
            \hhline{~|-|~} 
            & \multirow{1}{*}{Converted} & \cellcolor{gray!5}{\darkblue{\textit{The navy speckle book should be placed in the top region of the shelf}}} \\ 
        \cline{2-2} 
        \hline
        \multirow{3}{*}{\textbf{5-5}}  
            & \multirow{1}{*}{Original} & \cellcolor{gray!5}{\darkblue{\textit{Place the small light green bowl on the plate.}}} \\  
            \hhline{~|-|~} 
            & \multirow{2}{*}{Converted} & \cellcolor{gray!5}{\darkblue{\textit{In this scene, the goal is to place the small light green bowl on the plate, }}} \\ 
            & & \cellcolor{gray!5}{\darkblue{\textit{no additional actions are required. }}} \\ 
        \cline{2-2} 
        \hline
        \end{tabular}
    }
    \vspace{-6pt} 

    \label{tab:example_level5}
\end{table}


\end{document}